\documentclass[10pt]{article}

\usepackage{caption}
\usepackage{subcaption}
\usepackage{threeparttable}
\usepackage{times}
\usepackage{hyperref}
\def\mycitecolor{green!50!black}
\def\mylinkcolor{red!60!black}
\hypersetup{ colorlinks=true, linkcolor=\mylinkcolor, citecolor=\mycitecolor }
\usepackage[utf8]{inputenc}
\usepackage[T1]{fontenc}
\usepackage{graphicx}
\usepackage{booktabs}
\usepackage{amsmath}
\usepackage{amssymb} %mathbb packages
\usepackage{siunitx}
\usepackage{algorithm, algorithmicx, algpseudocode}
\usepackage{multirow}
\usepackage[sorting=nyt,style=numeric, natbib=true,doi=false,isbn=false,url=false,eprint=false, maxnames=999, maxcitenames=2]{biblatex} 
\DeclareNameAlias{default}{last-first}
\usepackage{placeins}
\usepackage{arxiv}
\usepackage{authblk}% for author affiliations
\usepackage[dvipsnames]{xcolor}

\usepackage{comment}
\usepackage{snaptodo}

\snaptodoset{block rise=1em}
\snaptodoset{margin block/.style={font=\tiny}}

\newcommand{\indep}{\perp \!\!\! \perp}
\DeclareMathOperator*{\argmin}{argmin} \def\mycitecolor{green!50!black}
\usepackage{mathtools}
\newcommand\myeq{\stackrel{\mathclap{\text{def}}}{=}}

\newtheorem{definition}{Definition}

\newtheorem{proposition*}{Proposition}
\newtheorem{proposition}{Proposition}
\newtheorem{lemma}{Lemma}

\newtheorem{assumption}{Assumption}
\newtheorem{proof}{Proof}

\title{How to select predictive models for causal inference?}
\author[1,2,*]{Matthieu Doutreligne}
\author[1]{Gaël Varoquaux}

\affil[1]{Inria, Soda, Saclay, France}
\affil[2]{Mission Data, Haute Autorité de Santé, Saint-Denis, France}
\affil[*]{Corresponding author: matthieu.doutreligne@inria.fr}

\begin{document}

%\corres{*\email{matthieu.doutreligne@inria.fr}}

\maketitle

\begin{abstract}
    As predictive models --\emph{eg} from machine learning-- give likely
    outcomes, they may be used to reason on the effect of an intervention,
    a causal-inference task. The increasing complexity of
    health data has opened the door to a plethora of models, but also the
    Pandora box of model selection: which of these models yield the most
    valid causal estimates?
    Here we highlight that classic machine-learning model selection
    does not select the best outcome models for causal inference.
    Indeed, causal model selection should control both outcome errors
    for each individual, treated or not
    treated, whereas only one outcome is observed.
    Theoretically, simple risks used in machine learning do not control
    causal effects when treated and non-treated population differ too much.
    More elaborate risks build proxies of the causal error using
    ``nuisance'' re-weighting to compute it on the observed data. But
    does computing these nuisance adds noise to model selection? Drawing
    from an extensive empirical study, we outline a good causal
    model-selection procedure: using the so-called
    $R\text{-risk}$; using flexible estimators to
    compute the nuisance models on the train set; and splitting out 10\% of
    the data to compute risks.
\end{abstract}
\clearpage

\section{Introduction}\label{sec:intro}

\subsection{Extending prediction to prescription requires causal model selection}

% Framing:
% - Increasing rich observational, ``real-life'' data.
%   - routinely collected health care
%   - convenient to build prognostic model (eg using claims or
%   administrative data
%   \cite{desai2020comparison} ) \cite{yurkovich2015systematic},
%   electronic health records \cite{fontana2019can}, encompassing notes and questionnaires
%   \cite{horng2017creating,simon2018predicting}
% - Motivates building increasingly complex predictive model, such as
%   prognostic models
% - Plethora of predictive models
% - From prediction to prescription There is often a desire to reason on intervention => causal inference is necessary

Increasingly rich data drives new predictive models. In health, new risks
or prognostic models leverage routinely-collected data, sometimes with
machine learning \cite{mooney2018bigdata}: predicting morbidity-related
outcomes from administrative data \cite{yurkovich2015systematic}, heart
failure from claims \cite{desai2020comparison}, sepsis from clinical
records \cite{horng2017creating}, suicide attempts from patient records
and questionnaires \cite{simon2018predicting}... Data may be difficult to
control and model, but claims of accurate prediction can be established
on left-out data
\cite{altman2009prognosis,poldrack2020establishment,varoquaux2022evaluating}.
Given a model predicting of an outcome of interest, it is tempting to use
it to guide decisions: will an individual benefit or not from an
intervention such as surgery \cite{fontana2019can}? This is a
causal-inference task and principled approaches can build
on contrasting the
prediction of the outcome with and without the treatment
\cite{snowden_implementation_2011,blakely2020reflection}.

% - Causal inference is difficult
% - Beyond the dream of assessing effectiveness and safety in real-word
%   practice \cite{black1996we} include off-label drug usage 
%   \cite{radley2006off} (which may lead to drug repurposing)
% - Plug in that outcome modeling is complementary to RCTs: observational
%   data gives weaker evidence than trials, but on both data outcome
%   modeling opens the door to individual.
% - Epidemiology has focused on propensity-score methods
% - Principle causal inference with outcome models is possible
% - It brings the promises of \emph{individualized treatment effects} --a
%   goal related to capturing heterogeneity and thus CATE--
% - It also performs well on causal-inference competition, although they
%   focused on the ATE

% The plethora of approaches approaches give different causal-inference
% estimates => need for guideline on causal model selection.

Outcome modeling is an integral part of the causal modeling toolkit,
under the names of
G-computation, G-formula \cite{robins_role_1986}, Q-model
\cite{snowden_implementation_2011}, or conditional mean
regression \cite{wendling_comparing_2018}. On observational data, causal
inference of treatment effects is brittle to un-accounted for
confounding, but it can assess
effectiveness and safety in real-word practice
\cite{black1996we,hernan_methods_2021} or off-label drug usage
\cite{radley2006off} for potential repurposing
\cite{hurle2013computational,dudley2011exploiting}. Epidemiology has
historically focused on methods that model treatment assignment
\cite{austin_moving_2015,grose_use_2020}, based on the propensity score
\cite{rosenbaum_central_1983} but recent empirical results
\cite{wendling_comparing_2018,dorie_automated_2019} show benefits of
outcome modeling to estimate average treatment effect (ATE). A major
benefit of using outcome modeling for causal inference is that these
methods naturally go beyond average effects,
estimating individualized or conditional average treatment effects
(CATE), important for precision medicine.
For this purpose, such methods are also invaluable on randomized trials
\cite{su2018random,lamont2018identification,hoogland2021tutorial}.

\begin{figure}[b!]
    \begin{minipage}{.4\linewidth}
        \caption{\textbf{Different outcome models lead to different
                estimation error on the Average Treatment Effects},
            with six different outcome
            models on 77 classic simulations where the true causal effect is
            known \cite{dorie_automated_2019}. The models are random forests, ridge regression with and without
            interaction with the treatment,
            (hyper-parameters detailed in Appendix
            \ref{apd:toy_example:acic_2016_ate_variability}). The different configurations are
            plotted as a function of increasing difference between treated and
            untreated population --detailed in
            \autoref{subsec:measuring_overlap}.  \\[1ex]
            There is no systematic best performer: choosing
            the best model among a family of candidate estimators is
            important.
        }\label{fig:acic_2016_ate_heterogeneity}%
    \end{minipage}\hfill%
    \begin{minipage}{.6\linewidth}
        \centering
        \includegraphics[width=0.95\linewidth]{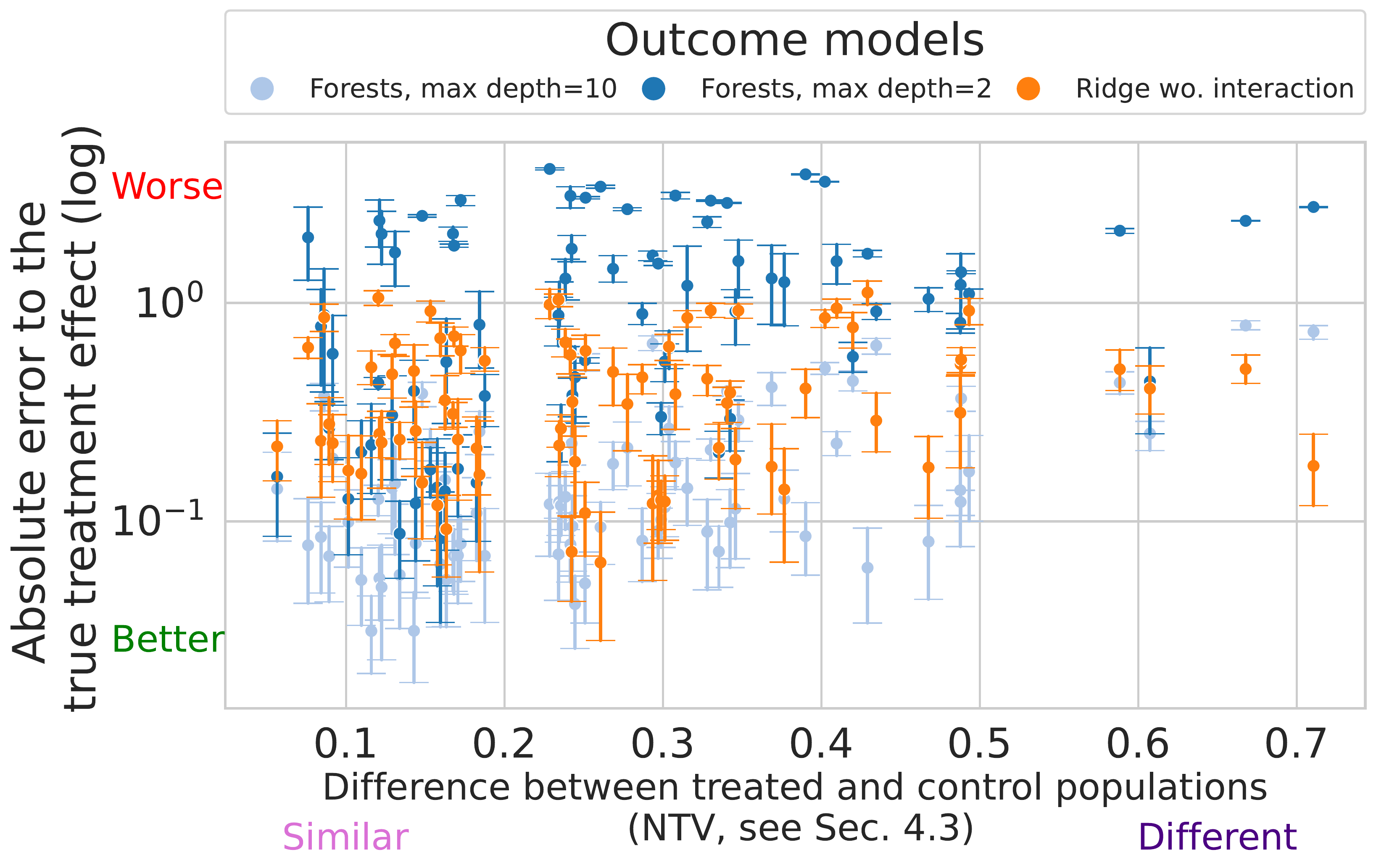}%
    \end{minipage}%
\end{figure}

%\idea{There is a variety of models for g-estimation}
Recent developments have seen the multiplication of predictive modeling
methods. Leaving aside the, overwhelming, machine-learning literature,
even methods specifically designed for causal
inference are numerous: Bayesian Additive Regression Trees
\cite{hill_bayesian_2011}, Targeted Maximum Likelihood Estimation
\cite{laan_targeted_2011,schuler_targeted_2017}, causal boosting
\cite{powers_methods_2018}, causal multivariate adaptive regression
splines (MARS) \cite{powers_methods_2018}, random forests
\cite{wager_estimation_2018, athey_generalized_2019},
Meta-learners \cite{kunzel_metalearners_2019}, R-learners
\cite{nie_quasioracle_2017}, Doubly robust estimation
\cite{chernozhukov_double_2018}...
%\idea{The variety of estimators calls for model selection procedures}
The wide variety of methods leaves the applied researcher with the difficult choice
of selecting between different estimators based on the data at hand.
Indeed, estimates may vary markedly when using different
models. For instance, Figure \ref{fig:acic_2016_ate_heterogeneity} shows
large variations obtained across four different outcome estimators on
2016 semi-synthetic datasets \cite{dorie_automated_2019}. Flexible models
such as random forests are doing well in most settings except
for setups where treated and untreated populations differ markedly in
which case a simple linear model (ridge) is to be preferred.
However a different choice of hyper-parameters
(max depth= 2) for random forests yield the poorest performances.
A simple rule of thumb such as preferring more flexible models does not work in
general; model selection is necessary.

%\idea{It is crucial to have tools for model selection.}

Standard practices to select models in predictive settings rely on
cross-validation on the error on the outcome
\cite{poldrack2020establishment,varoquaux2022evaluating}. However, as we
will see, these model-selection practices may not pick the best models
for causal inference, as they can be misled by inhomogeneities due to
treatment allocation.
Given complex, potentially noisy, data, which model is to be most trusted to
yield valid causal estimates? Because there is no single learner that performs
best on all data sets, there is a pressing need for clear guidelines to select
outcome models for causal inference.

\paragraph{Objectives and structure of the paper}

In this paper, we study \textit{model selection procedures} with a focus
on practical settings: \textit{finite samples} settings and without
\textit{well-specification} assumption. One question is whether
model-selection procedures, that rely on data split, can estimate
reliably enough the complex risks, theoretically motivated for causal
inference. Indeed, these risks force of departure from standard model-selection
procedures as they come with more quantities to estimate, which may
bring additional variance, leading to worse model selection.

We first give a simple illustration of the problem of causal model
selection and quickly review the important prior art. Then, in Section
\ref{sec:framework}, we set causal model selection in the
\emph{potential outcome} framework and detail the causal risks and
model-selection procedure. Section \ref{sec:theory} gives our theoretical
result. In section \ref{sec:empirical_study}, we run a thorough empirical
study, with many different settings covered. Finally, we comment our
findings in Section \ref{sec:discussion}.
Results outline how to best select outcome models for causal
inference with an adapted
cross-validation that estimate the so-called $R\text{-risk}$.
The $R\text{-risk}$
modulates observed prediction error to compensate for systematic
differences between treated and non-treated individuals. It
relies on the two \emph{nuisance} models,
themselves estimated from data and thus imperfect; yet these
imperfections do not undermine the benefit of the $R\text{-risk}$.

% XXX: merge the two paragraphs, above and below

\subsection{Illustration: the best predictor may not estimate best causal
    effects}%

\begin{figure}[b!]
    \begin{minipage}{.4\linewidth}
        \caption[The best predictor may not estimate best causal
            effects]{\textbf{Illustration}: a) a random-forest estimator
            with high performance for standard prediction (high $R^2$) but that
            yields poor causal estimates (large error between true effect $\tau$ and
            estimated $\hat{\tau}$), b) a linear estimator with smaller
            prediction performance leading to better causal estimation. \\[1ex]
            Selecting the estimator with the smallest error to the individual
            treatment effect $\mathbb{E}[(\tau(x) - \hat{\tau}(x))^2]$
            --the $\tau\text{-risk}$, def.\,\ref{def:tau_risk} -- would lead to
            the best causal estimates; however computing this error is not
            feasible: computing it requires access to unknown quantities:
            $\tau(x)$. \\[1ex]
            While the random forest fits the data better than the linear model, it
            gives worse causal inference because its error is very inhomogeneous between
            the treated and untreated. The $R^2$ score does not capture this
            inhomogeneity.
        }\label{fig:toy_example} \end{minipage}\hfill%
    \begin{minipage}{.577\linewidth}
        %\centering
        {\sffamily\small a) Random forest, good average prediction but bad
            causal inference}

        \hfill%
        \includegraphics[width=1\linewidth]{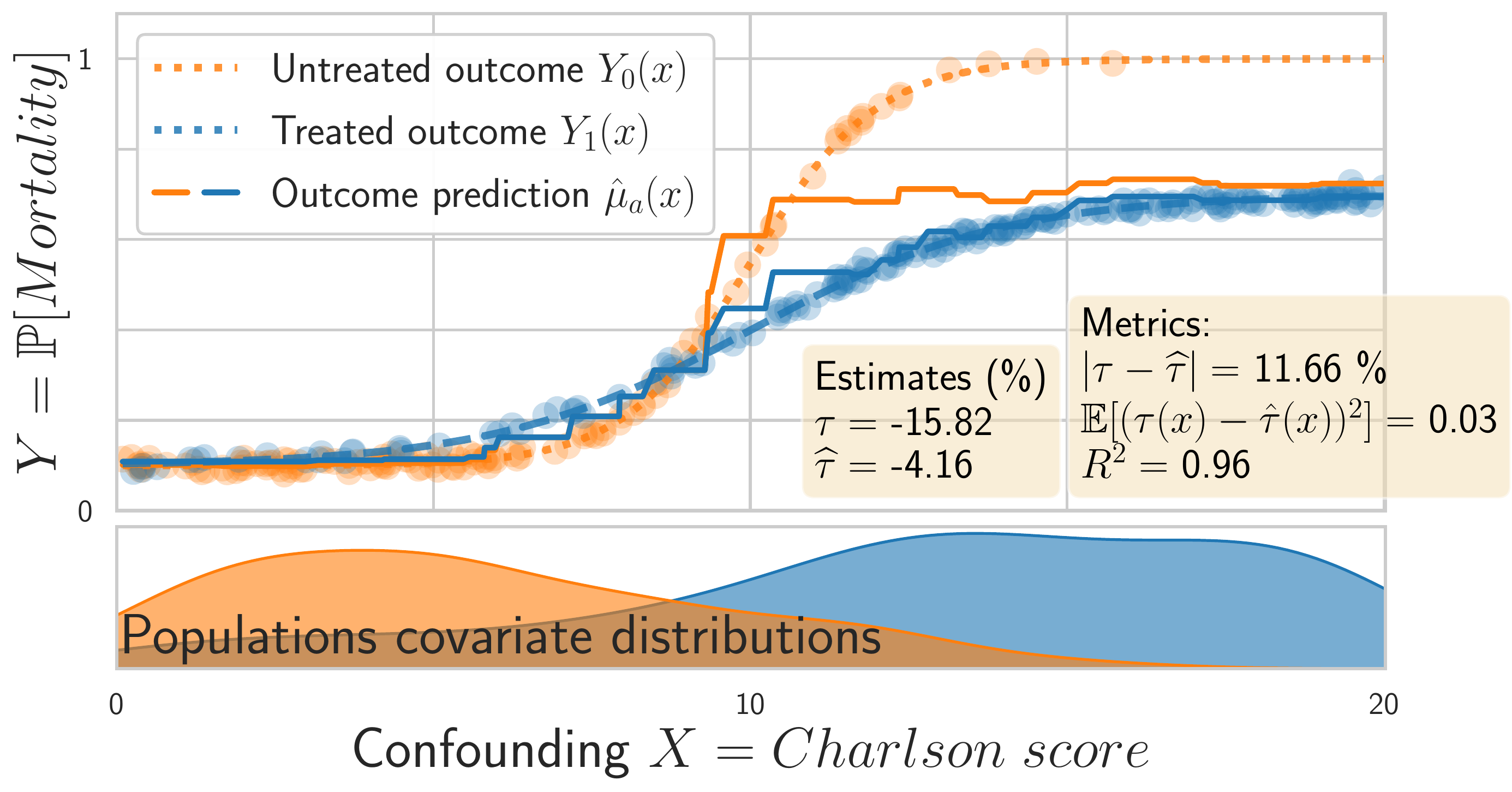}%

        {\sffamily\small b) Linear model, worse average prediction but better causal inference}

        \hfill%
        \includegraphics[width=1\linewidth]{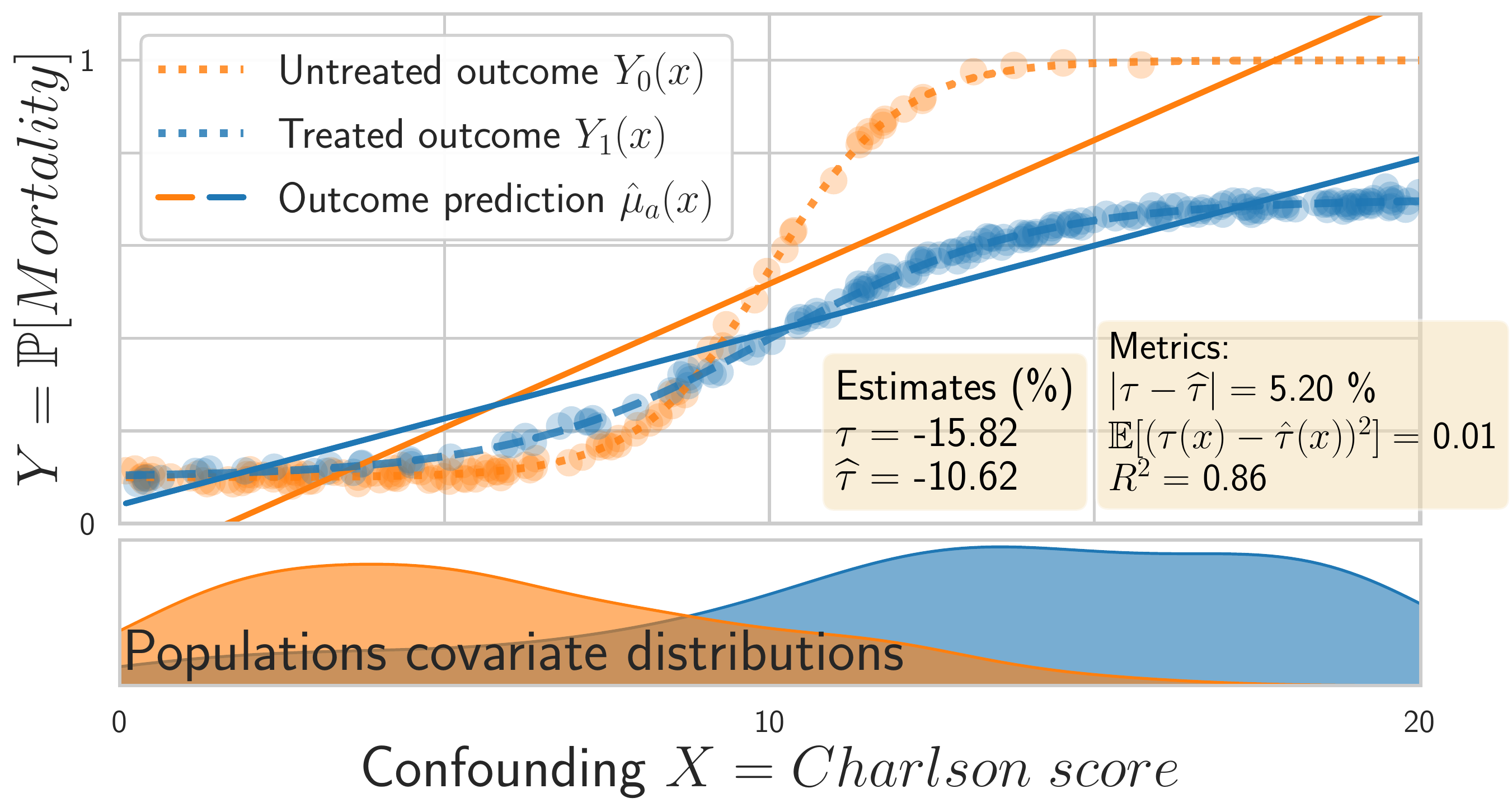}%
    \end{minipage}%
\end{figure}

Using a predictor to reason on causal effects relies on contrasting the
prediction of the outcome for a given individual with and without the
treatment --as detailed in \autoref{sec:neyman_rubin}.
Given various predictors of the outcome, we are interested in
selecting those that estimate best the treatment effect.
Standard predictive modeling or machine-learning practice selects the
best predictor, \emph{ie} the one that minimizes the expected error.
However, the best predictor may not be the best model to reason about
causal effects of an intervention, as we illustrate below.

Figure \ref{fig:toy_example} gives a toy example: the
outcome $Y \in [0, 1]$, the
probability of death, a binary treatment $A \in \{0, 1\}$ and a covariate
$X \in \mathbb R$ which summarizes the patient health status (eg. the
Charlson co-morbidity index \cite{charlson_new_1987}). We simulate a
situation for which the treatment is beneficial (decreases
mortality) for patients with high Charlson scores (bad health
status). On the contrary, the treatment has little effect for patients in good
condition (small Charlson scores).

Figure \ref{fig:toy_example}a shows a random forest predictor with a
counter-intuitive behavior: it predicts well on average the outcome (as
measured by a regression $R^2$ score) but perform poorly to estimate
causal quantities: the average treatment effect $\tau$ (as visible via
the error $|\tau - \hat{\tau}|$) or the individual treatment
effect (the error $\mathbb{E}[(\tau(x) - \hat{\tau}(x))^2]$).
On the contrary, Figure \ref{fig:toy_example}b shows a linear model with
smaller $R^2$ score but better causal inference.%

Intuitively, the problem is that causal estimation requires controlling
an error on both treated and non-treated outcome for the same individual:
the observed outcome, and the non-observed \emph{counterfactual} one.
The linear model is misspecified --the outcome functions are not
linear--, leading to poor $R^2$; but it interpolates better to regions
where there are few untreated individuals --high Charlson score-- and
thus gives better causal estimates. Conversely, the random forest puts
weaker assumptions on the data, thus has higher $R^2$ score but is biased
by the treated population in the region with poor overlap region, leading
to bad causal estimates.

This toy example illustrates that the classic minimum Mean Square Error
criterion is not suited to choosing a model among a family of candidate
estimators for causal inference.

\subsection{Prior work: model selection for outcome modeling (g-computation)}%

A natural way to select a predictive model for causal inference would be
an error measure between a causal quantity such as the conditional
average treatment effects (CATE) and models' estimate. But such error is
not a ``feasible'' risk: it cannot be computed solely from observed data
and requires oracle knowledge.

% XXX: I think that I should make the following paragraph shorter

\paragraph{Simulation studies of causal model selection}

Using eight simulations setups from \cite{powers_methods_2018}, where
the oracle CATE is known, \citet{schuler_comparison_2018} compare four
causal risks, concluding that for CATE estimation the best
model-selection risk is the so-called $R\text{-risk}$
\cite{nie_quasioracle_2017} --def.\,\ref{def:r_risk}, below. Their
empirical results are clear for randomized treatment allocation but less
convincing for observational settings where both simple Mean Squared
Error --MSE, $\mu\text{-risk}(f)$ def.\,\ref{def:mu_risk}-- and
reweighted MSE --$\mu\text{-risk}_{IPW}$ def.\,\ref{def:mu_ipw_risk}--
appear to perform better than $R\text{-risk}$ on half of the simulations.
Another work \cite{alaa_validating_2019} studied empirically both MSE and
reweighted MSE risks on the semi-synthetic ACIC 2016 datasets
\cite{dorie_automated_2019}, but did not include the $R\text{-risk}$ and
looked only at the agreement of the best selected model with the true
CATE risk --$\tau\text{-risk}(f)$ def.\, \ref{def:tau_risk}--, not on the
full ranking of methods compared to the true CATE. We complete these
prior empirical work by studying a wider variety of data generative
processes and varying the influence of overlap, an important parameter of
the data generation process which makes a given causal metric appropriate
\cite{damour_overlap_2020}. We also study how to best adapt
cross-validation procedures to causal metrics which themselves come with
models to estimate.

\paragraph{Theoretical studies of causal model selection}

Several theoretical works have proposed causal model selection procedures
that are \emph{consistent}: select the best model in a family given
asymptotically large data. These work typically rely on introducing a
CATE estimator in the testing procedure. For instance matching
\citep{rolling_model_2014}, an IPW estimate
\citep{gutierrez_causal_2016}, a doubly robust estimator
\citep{saito_counterfactual_2020}, or debiasing the error with influence
functions \cite{alaa_validating_2019}. However, for theoretical
guarantees to hold, the test-set correction needs to converge to the
oracle: it needs to be flexible enough --or well-posed-- and asymptotic
data. From a practical perspective, knowing that such requirements are
met implies having a good CATE estimate, which amounts to having solved
the original problem of causal model selection. We study how causal
model-selection procedures behave outside of these settings.

\paragraph{Statistical guarantees on causal estimation procedures}

Much work in causal inference has focused on building procedures that
guarantee asymptotically consistent estimators. Targeted Machine Learning
Estimation (TMLE) \cite{laan_targeted_2011,schuler_targeted_2017} and
Double Machine Learning \cite{chernozhukov_double_2018} both provide
estimators for Average Treatment Effect combining flexible treatment and
outcome models. Here also, theories require asymptotic regimes and
models to be \textit{well-specified}.

By contrast, \citet{johansson_generalization_2021} studies causal
estimation without assuming that estimators are well specified. They derive an
upper bound on the oracle error to the CATE ($\tau\text{-risk}$) that involves
the error on the outcome and the similarity of the distributions between the
features of treated and control patients. However, they focus on using this
upper bound for estimation, and do not give insights on model selection. In
addition, for hyperparameter selection, they rely on a plugin estimate of the $\tau\text{-risk}$ built with
counterfactual nearest neighbors, which has been shown ineffective
\cite{schuler_comparison_2018}.

\section{Formal setting of causal inference and model selection}\label{sec:framework}

\subsection{The Neyman-Rubin Potential Outcomes framework}%
\label{sec:neyman_rubin}%

\paragraph{Settings}

The Neyman-Rubin Potential Outcomes framework
\cite{naimi2023defining,imbens_causal_2015} enables statistical
reasoning on causal treatment effects: Given an outcome $Y \in \mathbb R$ (eg.
mortality risk or hospitalization length), function of a binary treatment $A \in
    \mathcal{A} = \{0, 1\}$ (eg.~a medical act, a drug administration), and baseline
covariates $X \in \mathcal{X} \subset \mathbb{R}^d$, we observe the factual
distribution,
$O = (Y(A), X, A) \sim \mathcal D = \mathbb P(y, x, a)$. However, we want to model the existence of potential observations
(unobserved ie. counterfactual) that correspond to a different treatment. Thus we want
quantities on the counterfactual distribution
$O^{*} = (Y(1), Y(0), X, A) \sim \mathcal D^{*} = \mathbb P(y(1), y(0), x, a)$.

Popular quantities of interest (estimands) are:
at the population level, the
Average Treatment Effect
\begin{equation*}
    \text{(ATE)}\qquad
    \tau \myeq \; \mathbb{E}_{Y(1),Y(0) \sim \mathcal D^*}[Y(1) - Y(0)];
\end{equation*}
to model heterogeneity, the Conditional Average Treatment Effect
\begin{equation*}
    \text{(CATE)}\qquad
    \tau (x) \myeq \; \mathbb{E}_{Y(1),Y(0) \sim \mathcal{D}^\star}[Y(1) - Y(0) | X=x].
\end{equation*}

\paragraph{Causal assumptions}

Some assumptions are necessary to assure identifiability of the causal estimands
in observational settings \cite{rubin_causal_2005}. We assume the usual
strong ignorability assumptions, composed of \emph{1)}
\emph{unconfoundedness} $\{Y(0),
    Y(1) \} \indep A | X$, \emph{~2)} \emph{strong overlap} ie. every patient has a
strictly positive probability to receive each treatment, \emph{3)}
\emph{consistency}, and \emph{4)} \emph{generalization} (detailed in Appendix
\ref{apd:causal_assumptions}). In this work, we investigate the
role of the overlap \cite{damour_overlap_2020}, which is testable with
data.

\paragraph{Estimating treatment effects with outcome models}\label{subsec:estimators}

Should we know the two expected outcomes for a given $X$,
we could compute the
difference between them, which gives the causal effect of the treatment.
These two expected outcomes can be computed from the observed data:
the consistency \ref{assumption:consistency} and unconfoundedness
\ref{assumption:ignorability} assumptions imply the equality of two different
expectations:
\begin{equation}\label{eq:mu_identification}
    \mathbb E_{Y(a) \sim \mathcal{D^{\star}}} [Y(a)|X=x] = \mathbb E_{Y \sim \mathcal{D}} [Y|X=x, A=a]
\end{equation}
On the left, the expectation is taken on the counterfactual unobserved
distribution. On the right, the expectation is taken on the factual observed
distribution conditionally on the treatment. This equality is referred as the
g-formula identification \cite{robins_new_1986}. For the rest of the
paper, the expectations will always be taken on the factual observed
distribution $\mathcal{D}$, and we will omit
to explicitly specify the distribution. This identification leads to outcome based estimators (ie.
g-computation estimators\cite{snowden_implementation_2011}), targeting the
ATE $\tau$ with outcome modeling:
\begin{equation}
    \tau = \mathbb E_{Y \sim \mathcal{D^{\star}}}[Y(1) - Y(0)|X=x] = \mathbb E_{Y \sim \mathcal{D}}[Y|A=1] - \mathbb E_{Y \sim \mathcal{D}}[Y| A=0]
    \label{eq:tau_population}
\end{equation}
This equation has two central quantities: the conditional expectancy
function of the outcome associated to specific covariates and
treatment or not is, often called the \emph{response function}:
\begin{equation}
    \text{(Response function)}
    \qquad
    \mu_{a}(x) \myeq \; \mathbb E_{Y \sim \mathcal{D}} [Y|X=x, A=a].
    \label{def:mu_a}
\end{equation}

Given a sample of data and the oracle response functions $\mu_0, \mu_1$, the
finite sum version of \autoref{eq:tau_population} leads to an
estimator of the ATE written:
\begin{equation}
    \hat \tau = \frac{1}{n} \biggl(\sum_{i=1}^n \mu_{1}(x_i) - \mu_{0}(x_i) \biggr)
    \label{eq:ate_estimate}
\end{equation}
This estimator is an oracle \textbf{finite sum estimator} by opposition to the
population expression of $\tau$, $\mathbb{E}[\mu_{1}(x_i) - \mu_{0}(x_i)]
$,
which involves an expectation taken on the full
distribution $\mathcal D$, which is observable but requires infinite data. For
each estimator $\ell$ taking an expectation over $\mathcal D$, we use the symbol
$\hat \ell$ to note its finite sum version.

Similarly to the ATE, for the CATE, at the individual level:
\begin{equation}
    \tau(x) = \mu_{1}(x) - \mu_{0}(x)
    \label{eq:cate_estimate}
\end{equation}

\paragraph{Robinson decomposition}
Another decomposition of the outcome model plays in important role,
the R-decomposition
\cite{robinson_rootnconsistent_1988}:
introducing two quantities, the conditional mean outcome
and the probability to be treated (known as propensity score \cite{rosenbaum_central_1983}):
\begin{align}
    \text{(Conditional mean outcome)} \qquad  m(x) \myeq \; & \mathbb E_{Y \sim \mathcal{D}} [Y|X=x].
    \label{def:m}
    \\
    \text{(Propensity score)}\qquad
    e(x) \myeq \;                                           & \mathbb P[A=1|X=x],
    \label{def:propensity_score}
\end{align}
the outcome can be written
\begin{equation}\label{eq:r_decomposition}
    \text{(R-decomposition)}\qquad y(a) = m(x) + \big( a - e(x) \big) \tau(x) + \varepsilon(x; a) \quad\text{with}\quad \mathbb E[\varepsilon(X; A)|X, A] = 0
\end{equation}
$m$ and $e$ are often called
\emph{nuisances} \cite{chernozhukov_double_2018}; they are unknown in
general.

%As noted by \cite{johansson_generalization_2021}, the machine learning
%community often referred to the CATE by ITE, the Individual Treatment Effect.
%From a purely causal point of view, the ITE is uniquely defined for each
%individual and might not be accessible: $ITE(x_i) = Y_i(1) -  Y_i(0)$. On the
%contrary, the CATE can always be derived by taking conditional expectancies. It
%is the expected effect of the treatment in the region of the covariate space
%around X. %Too cultivate

\subsection{Model-selection risks, oracle and feasible}\label{sec:problem:model_selection}

\paragraph{Causal model selection}\label{sec:problem:causal_selection}

We formalize model selection for causal estimation. Thanks to the g-formula
identification (\autoref{eq:mu_identification}), a given outcome model $f: \mathcal X
    \times \mathcal A \rightarrow \mathcal{Y}$ --learned from data or built from
domain knowledge-- induces feasible estimates of the ATE and CATE (eqs
\ref{eq:ate_estimate} and \ref{eq:cate_estimate}), $\hat \tau_{f}$ and $\hat \tau_{f}(x)$.
Let $\mathcal F=\{f: \mathcal X \times \mathcal A \rightarrow \mathcal{Y}\}$ be
a family of such estimators. Our goal is to select the best candidate in this
family for the observed dataset $O$ using a risk of interest
$\ell$:
\begin{equation}
    f^*_{\ell} = \argmin_{f \in \mathcal{F}} \ell(f, O)
    \label{eq:causal_model_selection}
\end{equation}

We now detail possible risks $\ell$, risks useful for causal
model selection, and how to compute them.

\paragraph{The $\tau\text{-risk}$: an oracle error risk}\label{paragraph:oracle_metrics}
As we would like to target the CATE, the following
evaluation risk is natural:
\begin{definition}[$\tau\text{-risk}(f)$]\label{def:tau_risk} also called PEHE
    \cite{schulam_reliable_2017, hill_bayesian_2011}:
    \begin{equation*}\label{eq:tau_risk}
        \tau\text{-risk}(f) = \mathbb E_{X\sim p(X)}[(\tau(X) - \hat \tau_f(X))^2]
    \end{equation*}
    its finite-sum version over the observed data:
    \begin{equation*}
        \widehat{\tau\text{-risk}}(f) = \sum_{x \in O} \big(\tau(x) - \hat \tau_f(x)\big)^2
    \end{equation*}
\end{definition}

However these risks are not feasible because the oracles $\tau(x)$ are
not accessible, with the observed data $(Y, X, A) \sim \mathcal D$.

\paragraph{Feasible error risks}\label{paragraph:feasible_metrics}
\emph{Feasible} risks are based on the prediction error of the outcome model
and \emph{observable} quantities.

\begin{definition}[Factual $\mu\text{-risk}$]\label{def:mu_risk}
    \cite{shalit_estimating_2017} This is the usual Mean Squared Error on
    the target y. It is what is typically meant by ``generalization error'' in
    supervised learning and estimated with cross-validation:
    \begin{equation*}\label{eq:mu_risk}
        \mu\text{-risk}(f)=\mathbb{E}_{(Y, X, A) \sim \mathcal D}\left[(Y-f(X ; A))^2 \right]
    \end{equation*}
\end{definition}

The following risks use the nuisances $e$
--propensity score, def \ref{def:propensity_score}-- and $m$ --conditional mean
outcome, def \ref{def:m}. We give the definitions as \textit{semi-oracles},
function of the true unknown nuisances, but later instantiate them with estimated
nuisances, noted $\big(\check e, \check m \big)$. Semi-oracles risks are
superscripted with the $^{\star}$ symbol.

\begin{definition}[$\mu\text{-risk}_{IPW}^{\star}$]\label{def:mu_ipw_risk}
    \cite{vanderlaan_unified_2003} Let the inverse propensity weighting
    function $w(x, a) = \frac{a}{e(x)} + \frac{1 - a}{1 - e(x)}$, we define the
    semi-oracle Inverse Propensity Weighting risk,
    \begin{equation*}\label{eq:mu_ipw_risk}
        \mu\text{-risk}_{IPW}^{\star}(f) = \mathbb{E}_{(Y, X, A) \sim \mathcal D}\left[ \Big( \frac{A}{e(X)} + \frac{1-A}{1-e(X)} \Big) (Y-f(X ; A))^2 \right]
    \end{equation*}
\end{definition}

\smallskip

\begin{definition}[$\tau\text{-risk}^{\star}_{IPW}$]\label{def:tau_ipw_risk}
    \cite{wager_estimation_2018}
    The CATE $\tau(x)$ can be estimated propensity score, with a
    regression against inverse propensity weighted outcomes \cite{
        athey2016recursive,gutierrez_causal_2016,wager_estimation_2018}. From this objective, we can derive
    the $\tau\text{-risk}_{IPW}$.
    \begin{equation*}
        \tau\text{-risk}^{\star}_{IPW}(f) =\mathbb{E}_{(Y, X, A) \sim
            \mathcal D} \left[ \Big(Y \frac{A - e(X)}{e(X)
                (1-e(X))}-\tau_f\left(X\right)\Big)^2 \right]  =\mathbb{E}_{(Y, X, A)
            \sim \mathcal D} \left[ \left(Y \left( \frac{A}{e(X)} -
            \frac{1-A}{1-e(X)}\right)-\tau_f\left(X\right)\right)^2 \right]
    \end{equation*}
\end{definition}

\begin{definition}[$U\text{-risk}^{\star}$]\label{def:u_risk}
    \cite{kunzel_metalearners_2019,nie_quasioracle_2017} Based on
    the Robinson decomposition --eq. \ref{eq:r_decomposition}
    \cite{robinson_rootnconsistent_1988}, the U-learner
    uses the $A-e(X)$ term
    in the denominator. The derived risk is:
    \begin{equation*}
        U\text{-risk}^{\star}(f) =\mathbb{E}_{(Y, X, A) \sim \mathcal D}
        \left[
            \left( \frac{Y-m\left(X\right)}{A-e\left(X\right)} -
            \tau_f\left(X\right)\right)^{2} \right]
    \end{equation*}
    Note that extreme propensity weights in the
    denominator term might inflate errors in the numerator due to imperfect
    estimation of the mean outcome $m$
    the numerator errors, leading to a highly biased metric.
\end{definition}

\begin{definition}[$R\text{-risk}^{\star}$]\label{def:r_risk}
    \cite{nie_quasioracle_2017,schuler_comparison_2018}
    The $R\text{-risk}$ also uses two nuisance $m$ and $e$:
    \begin{equation*}
        R\text{-risk}^{\star}(f) =\mathbb{E}_{(Y, X, A) \sim \mathcal D} \big[
            \big(\left(Y-m\left(X\right)\right) -\left(A-e\left(X\right)\right) \tau_f\left(X\right)\big)^{2} \big]
    \end{equation*}
\end{definition}

It is also motivated by the Robinson decomposition --eq. \ref{eq:r_decomposition}
\cite{robinson_rootnconsistent_1988}. It performs well in various simulations
where using lasso, boosting or kernel ridge regressions for both the nuisances
$(\check e, \check m)$ and the target $\hat \tau(x)$
\cite{nie_quasioracle_2017}.

These risks are summarized in Table \ref{tab:evaluation_metrics}.

\begin{table}[!h]
    \makebox[\linewidth]{
        \begin{threeparttable}[b]
            \caption{Review of causal risks}
            \centering
            \label{tab:evaluation_metrics}
            \begin{tabular}{llr}
                \toprule
                Risk                                                                                           & Equation
                                                                                                               & Reference                                                                                                                                           \\
                \midrule
                $mse(\tau(X), \tau_f(X))=\tau\text{-risk}$                                                     & $\mathbb E_{X\sim
                            p(X)}[(\tau(X) - \hat \tau_f(X))^2] $
                                                                                                               & Eq. \ref{eq:tau_risk} \cite{hill_bayesian_2011}                                                                                                     \\
                $mse(Y, f(X)) = \mu\text{-risk}$                                                               & $\mathbb{E}_{(Y, X, A)
                        \sim \mathcal D}\left[(Y-f(X ; A))^2 \right]$
                                                                                                               & Def. \ref{def:mu_risk} \cite{schuler_comparison_2018}                                                                                               \\
                $\mu\text{-risk}_{IPW}^*$                                                                      & $\mathbb{E}_{(Y, X, A)
                        \sim \mathcal D}\left[ \Big( \frac{A}{e(X)} + \frac{1-A}{1-e(X)} \Big)
                (Y-f(X ; A))^2 \right]$                                                                        & Def.
                \ref{def:mu_ipw_risk} \cite{vanderlaan_unified_2003}                                                                                                                                                                                                 \\
                $\tau\text{-risk}^{\star}_{IPW}$                                                               & $\mathbb{E}_{(Y, X, A) \sim \mathcal D} \left[ \Big(Y \left( \frac{A}{e(X)} - \frac{1-A}{1-e(X)}\right)-\hat \tau_f\left(X\right)\Big)^2 \right]$ &
                Def. \ref{def:tau_ipw_risk} \cite{wager_estimation_2018}
                \\
                $U\text{-risk}^*$                                                                              & $\mathbb{E}_{(Y, X, A) \sim \mathcal D}  \big[
                \big( \frac{Y-m\left(X\right)}{A-e\left(X\right)} -  \hat \tau_f\left(X\right)\big)^{2} \big]$ &
                Def. \ref{def:u_risk} \cite{nie_quasioracle_2017}
                \\
                $R\text{-risk}^*$ \tnote{1}                                                                    & $\mathbb{E}_{(Y, X, A)
                        \sim \mathcal D} \big[\big(\left(Y-m\left(X\right)\right)
                -\left(A-e\left(X\right)\right) \hat \tau_f\left(X\right)\big)^{2} \big]$                      &
                Def. \ref{def:r_risk} \cite{nie_quasioracle_2017}
                \\
                \bottomrule
            \end{tabular}
            \begin{tablenotes}
                \item [1] Called $\tau \text{-risk}_R$ in
                \citet{schuler_comparison_2018}.
            \end{tablenotes}
        \end{threeparttable}
    }
\end{table}

\subsection{Estimation and model selection procedure}\label{problem:estimation_procedure}

\begin{figure}[h!]
    \centering
    \includegraphics[width=0.7\linewidth]{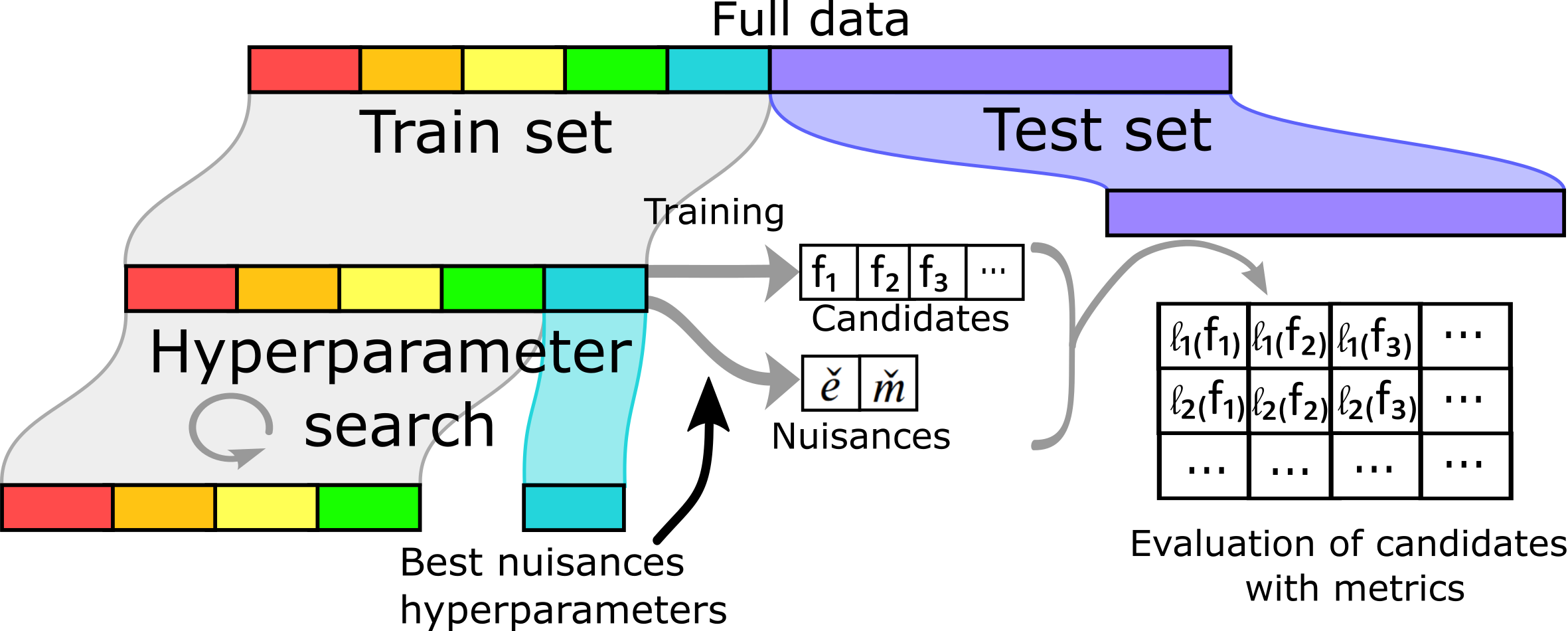}
    \caption{Estimation procedure for causal model
        selection.}\label{problem:estimation_procedure:figure}
\end{figure}

Causal model selection (as in
\autoref{eq:causal_model_selection}) may involve estimating various quantities
from the observed data: the outcome model $f$, its induced risk as
introduce in the previous section, and possibly nuisances required by the
risk.
Given a dataset with $N$ samples, we split out a train and a test sets
$(\mathcal{T}, \mathcal{S})$. We
fit each candidate estimator $f \in \mathcal{F}$ on $\mathcal{T}$. We also fit
the nuisance models $(\check e, \check m)$ on the train set
$\mathcal{T}$, setting hyperparameters by a nested
cross-validation before fitting the nuisance estimators with these parameters
on the full train set. Causal quantities are then computed by applying the fitted  candidates
estimators $f \in \mathcal{F}$ on the test set $\mathcal{S}$. Finally, we
compute the model-selection metrics for
each candidate model on the test set. This procedure is described in Algorithm
\ref{problem:estimation_procedure:algo} and Figure
\ref{problem:estimation_procedure:figure}.

As extreme inverse propensity weights induce high variance, clipping can be
useful for numerical stability
\cite{swaminathan_counterfactual_2015, ionides_truncated_2008}.

\begin{algorithm}[!htbp]
    \caption{Evaluation of selection procedures for one
        simulation}\label{problem:estimation_procedure:algo} {%
        Given a train and a test sets $(\mathcal{T}, \mathcal{S}) \sim \mathcal{D}$,
        a family of candidate estimators $\{f \in \mathcal F\}$, a set of causal
        metrics $\ell \in \mathcal L$:
        \begin{enumerate}
            \item Prefit: Learn estimators for unknown nuisance quantities $(\check e,\,\check m)$ on the training set $\mathcal{T}$
            \item Fit: $\forall f \in \mathcal{F}$ learn $\hat f(\cdot, a)$ on
                  $\mathcal T$
            \item Model selection: $\forall{x} \in \mathcal{S}$ predict $\big(\hat f(x, 1), \hat f(x, 0)\big)$ and evaluate each candidate estimator with
                  each causal metric $\mathcal M(\hat f, \mathcal{S})$. For each causal
                  metric $\ell  \in \mathcal{L}$ and each candidate estimator $f \in
                      \mathcal{F}$, store the metric value: $\ell(f, \mathcal S)$ -- possibly
                  function of $\check e$ and $\check m$
                  %\item Metric evaluation: return the oracle evaluation metrics
                  %evaluated on $\mathcal{S}$: $\big(\widehat{\tau\text{-risk}}(\hat
                  %f^*_{\ell}); \widehat{\ell}_{ATE}(\hat f^*_{\ell}) \big)$
        \end{enumerate}

    }
\end{algorithm}

\section{Theory: Links between feasible and oracle risks}\label{sec:theory}

We now relate two feasible risks, $\mu \text{-risk}_{IPW}$ and the
$R\text{-risk}$ to the oracle $\tau\text{-risk}$. Both results make
explicit the role of overlap for the performances of causal risks.

These bounds depend on a specific form of residual that we now define: for each potential outcome, $a \in  \{0; 1\}$, the variance conditionally on $x$
is\cite{shalit_estimating_2017}:
\begin{equation*}\label{eq:residuals}
    \sigma_{y}^{2}(x ; a) \overset{\text{def}}{=}
    \int_{y}\left(y-\mu_{a}(x)\right)^{2} p(y \mid x=x ; A=a) \, d y
\end{equation*}
Integrating over the population, we get the Bayes squared error:
$\sigma^2_{B}(a) = \int_{\mathcal X} \sigma_y^2(x;a) p(x)dx$
and its propensity weighted version:
$\tilde{\sigma}^2_{B}(a) = \int_{\mathcal X}\sigma_y^2(x;a)\,  p(x;
    a)\,dx$. In case of a purely deterministic link between the
covariates, the treatment, and the outcome, these residual terms are null.

\subsection{Upper bound of $\tau\text{-risk}$ with
    $\mu\text{-risk}_{IPW}$}\label{theory:mu_risk_ipw_bound}

\begin{proposition}[Upper bound with $\mu \text{-risk}_{IPW}$
    ]\label{theory:prop:mu_risk_ipw_bound}
    \cite{johansson_generalization_2021} Given an outcome model $f$, let a
    weighting function $w(x; a) = \frac{a}{e(x)} + \frac{1-a}{1-e(x)}$ as the
    Inverse Propensity Weight. Then, under overlap (assumption
    \ref{assumption:overlap}), we have:
    \begin{equation*}
        \tau\text{-risk}(f) \leq \; 2 \, \mu\text{-risk}_{IPW}(w, f)
        - 2 \, \big(\sigma^2_{B}(1) +  \sigma^2_{B}(0)\big)
    \end{equation*}
\end{proposition}
This result has been derived in previous work
\cite{johansson_generalization_2021}. It links $\mu\text{-risk}_{IPW}$ to
the squared residuals of each population thanks to a reweighted mean-variance
decomposition. For completeness, we provide the proof in .

The upper-bound comes from the triangular inequality applied to the residuals of
both populations. Interestingly, the two quantities are equal when the
absolute residuals on treated and untreated populations are equal on the
whole covariate space, \emph{ie} for all
$x \in \mathcal X, |\mu_1(x) - f(x, 1)| = |\mu_0(x) - f(x, 0)|$.
The main source of difference between the oracle $\tau \text{-risk}$ and the
reweighted mean squared error, $\mu\text{-risk}_{IPW}$, comes from heterogeneous
residuals between populations. These quantities are difficult to characterize as
they are linked both to the estimator and to the data distribution.
This bound indicates that minimizing the $\mu\text{-risk}_{IPW}$ helps to
minimize the $\tau\text{-risk}$, which leads to
interesting optimization procedures \cite{johansson_generalization_2021}. However, there is no
guarantee that this bound is tight, which makes it less useful for model
selection.

Assuming strict overlap (probability of all individuals being treated or not
bounded away from 0 and 1 by $\eta$, appendix \ref{apd:causal_assumptions}), the
above bound simplifies into a looser one involving the usual mean squared error:
$\tau\text{-risk}(f)\leq \frac{2}{\eta}\, \mu\text{-risk}(f) -  2 \, \big(\sigma^2_{B}(1) +  \sigma^2_{B}(0)\big)$. For weak overlap (propensity scores not bounded far from 0
or 1), this bound is very loose (as shown in Figure \ref{fig:toy_example})
and is not appropriate to discriminate between models with close performances.

%\md{Lower bound: I think that there is no lower bound of the tau-risk with the
%mu-risk. That means that we can have a mu risk of 0 wheras the taurisk is
%non-null. Todo for myself: develop an example with overfitted 1NN on the
%observed data.}

\subsection{Reformulation of the $R\text{-risk}$ as reweighted
    $\tau\text{-risk}$}\label{theory:r_risk_rewrite}

We now derive a novel rewriting of the $R\text{-risk}$, making explicit its link
with the oracle $\tau \text{-risk}$.

\begin{proposition}[$R \text{-risk}$ as reweighted
        $\tau\text{-risk}$]\label{theory:prop:r_risk_rewrite} Given an outcome model
    $f$, its $R\text{-risk}$ appears as weighted version of its $\tau\text{-risk}$
    (Proof in Appendix \ref{apd:proofs:r_risk_rewrite}):
    \begin{equation}
        R\text{-risk}^*(f) = \int_{x} e(x)\big(1-e(x)\big)\big(\tau(x)-\tau_ {f}(x)\big)^{2} p(x) d x \;+\; \tilde{\sigma}_B^2(1) \;+\; \tilde{\sigma}_B^{2}(0)
    \end{equation}
\end{proposition}

The $R \text{-risk}$ targets the oracle at the cost of an overlap re-weighting
and the addition of the reweighted Bayes residuals, which are independent of
$f$. In good overlap regions the weights $e(x) \big(1-e(x) \big)$ are close to
$\frac{1}{4}$, hence the $R \text{-risk}$ is close to the desired gold-standard
$\tau \text{-risk}$. On the contrary, for units with extreme overlap violation,
these weights goes down to zero with the propensity score.

% \begin{remark}[Trivial upper bound] Because $e(x)\leq1$ and $(1-e(x))\leq1$,
%   we see immediately that \begin{equation} R \text{-risk}^* \leqslant
%   \tau{\text{-risk}}+\tilde{\sigma_{B}}(1)^{2}+\tilde{\sigma_{B}}^{2}(0)
%   \end{equation} \end{remark}

% \begin{remark}[Lower bound] Due to strict overlap assumption, \begin{equation}
%   R \text{-risk}^* \geqslant \eta (1 - \eta)
%   \tau{\text{-risk}}-\tilde{\sigma_{B}}(1)^{2}-\tilde{\sigma_{B}}^{2}(0)
%   \end{equation} \end{remark}

% \begin{remark}[Cases of loose upper bounds] Giving a strict overlap assumption
%   $\eta$, we can exhibit for all $C \; st. \; \eta^{-1}>C>1$, simple
%   simulations where: \begin{equation} R \text{-risk}^* \leqslant C \; \eta \;
%   \tau \text{-risk}(f) \quad \end{equation} \end{remark}

\subsection{Interesting special cases}

\paragraph{Randomization special case}\label{remark:rct} If the treatment is
randomized as in RCTs, $p(A=1 \mid X=x) = p(A=1)=p_A$, thus
$\mu\text{-risk}_{IPW}$ takes a simpler form:
\begin{equation*}
    \mu\text{-risk}_{IPW} = \mathbb{E}_{(Y, X, A) \sim \mathcal D}\left[ \Big( \frac{A}{p_A} + \frac{1-A}{1-p_A} \Big) (Y-f(X ; A))^2 \right]
\end{equation*}
However, even if we have randomization, we still can have large differences
between $\tau\text{-risk}$ and $\mu\text{-risk}_{IPW}$ coming from heterogeneous
errors between populations as noted in Section \ref{theory:mu_risk_ipw_bound}
and shown experimentally in simulations \cite{schuler_comparison_2018}.

Concerning the $R\text{-risk}$, replacing $e(x)$ by its randomized value $p_A$
in Proposition \ref{theory:prop:r_risk_rewrite} yields the oracle
$\tau\text{-risk}$ up to multiplicative and additive constants:
\begin{equation}
    R\text{-risk} = p_A \, (1-p_A) \, \tau\text{-risk} \;+\; (1 - p_A) \,\sigma_B^2(0) \;+\; p_A \sigma_B^2(1)
\end{equation}
Therefore, optimizing estimators for CATE with $R\text{-risk}^*$ in the
randomized setting is optimal if we target the $\tau\text{-risk}$. This explains
the strong performances of $R\text{-risk}$ in randomized setups
\cite{schuler_comparison_2018} and is a strong argument in favor of this
risk for heterogeneity estimation in RCTs.

\paragraph{Oracle Bayes predictor}\label{remark:bayes_oracle} Consider the case
where we have access to the oracle Bayes predictor for the outcome ie.~$f(x,
    a)=\mu(x, a)$, then all risks are equivalent up to the residual variance:
\begin{equation}
    \tau\text{-risk}(\mu) = \mathbb E_{X\sim p(X)}[(\tau(X) - \tau_{\mu}(X))^2] = 0
\end{equation}
\begin{equation}
    \mu\text{-risk}(\mu) = \mathbb E_{(Y, X, A) \sim p(Y;X;A)}[\big( Y - \mu_A(X)\big)^2] = \int_{\mathcal X, \mathcal A}
    \,\varepsilon(x,a)^2 p(a \mid x) \,p(x) \,dx\,da  \leq \sigma_B^{2}(0) + \sigma_B^{2}(1)
\end{equation}
\hspace*{0.5em}
\begin{equation}
    \mu\text{-risk}_{IPW}(\mu) = \sigma_B^{2}(0) + \sigma_B^{2}(1)  \quad \text{follows from Lemma \ref{apd:proofs:mu_risk_ipw_link_mu}}
    %\notag
\end{equation}
\hspace*{0.5em}
\begin{equation}
    R\text{-risk}(\mu) = \tilde{\sigma}_B^{2}(0) + \tilde{\sigma}_B^{2}(1)
    \leq \sigma_B^{2}(0) + \sigma_B^{2}(1) \quad \text{follows directly from Proposition \ref{theory:prop:r_risk_rewrite}}%\notag
\end{equation}
Thus, differences between causal risks only matter in finite sample regimes.
Universally consistent learners converge to the Bayes risk in asymptotic
regimes, making all model selection risks equivalent. However, in practice
choices must be made in non-asymptotic regimes.

\section{Empirical Study}\label{sec:empirical_study}

%We now explore empirically the behavior of these different causal selection
%metrics.
%\subsection{Causal metrics under evaluation}

We evaluate the following causal metrics, oracle and feasible
versions, presented in Table
\ref{tab:evaluation_metrics}:
$\widehat{\mu\text{-risk}}_{IPW}^*$,
$\widehat{R\text{-risk}}^*$,
$\widehat{U\text{-risk}}^*$,
$\widehat{\tau\text{-risk}_{IPW}}^*$,
$\widehat{\mu\text{-risk}}$,
$\widehat{\mu\text{-risk}}_{IPW}$,
$\widehat{R\text{-risk}}$,
$\widehat{U\text{-risk}}$,
$\widehat{\tau\text{-risk}_{IPW}}$.
We benchmark the metrics in a variety of settings:
many different simulated data generation
processes and three semi-simulated datasets \footnote{Scripts for the simulations and the selection procedure are available at
    \url{https://github.com/strayMat/caussim}.
}.

\subsection{Caussim: Extensive simulation settings}\label{subsec:simulations}

\paragraph{Data Generation Process}

We use simulated data, on which the ground-truth causal effect is known. Going
further than prior empirical studies of causal model selection
\cite{schuler_comparison_2018,alaa_validating_2019}, we use multiple
generative processes, to reach more general conclusions (as discussed
in Appendix \ref{apd:results:fig:seed_effect}).

We generate the response functions using random bases.
Basis extension methods are common in biostatistics, \emph{eg} functional
regression with splines
\cite{howe_splines_2011, perperoglou_review_2019}. By allowing the
function to vary at specific knots, they give flexible --non-linear--
models of the studied mechanisms. Taking inspiration from splines, we use
random approximation of Radial Basis Function (RBF) kernels
\cite{rahimi_random_2008} to generate the response surfaces. RBF use the
same process as polynomial splines but replace polynomial by Gaussian
kernels. Unlike polynomials, Gaussian kernels have
decreasing influences in the input space.
This avoids unrealistic divergences of the population response surfaces
at the ends of the feature space.

The number of basis functions --\emph{ie. knots}--, controls the complexity of
the ground-truth response surfaces and treatment. We first use this process to
draw the non-treated response surface $\mu_0$ and the causal-effect $\tau$. We
then draw the observations from a mixture two Gaussians, for the treated and non
treated. We vary the separation between the two Gaussians to control the amount
of overlap between treated and control populations, as it an important parameter
for causal inference (related to $\eta$ which appears in section
\ref{theory:mu_risk_ipw_bound}). Finally, we generate the observed outcomes
adding Gaussian noise. We generated such datasets 1000 times, with
uniformly random overlap parameters $\theta \in \left[ 0, 2.5 \right]$. Appendix
\ref{apd:experiments:generation} gives more details on the data generation.

\captionsetup[sub]{font=large,labelfont={bf,sf}}%

\begin{figure}[!t]
    \begin{minipage}{0.48\linewidth}
        \centering
        %\makebox[(\textwidth-9mm)/2]{\textbf{a. Caussim}}
        \includegraphics[width=0.98\linewidth,clip,trim={0 15.3cm 0 0}]{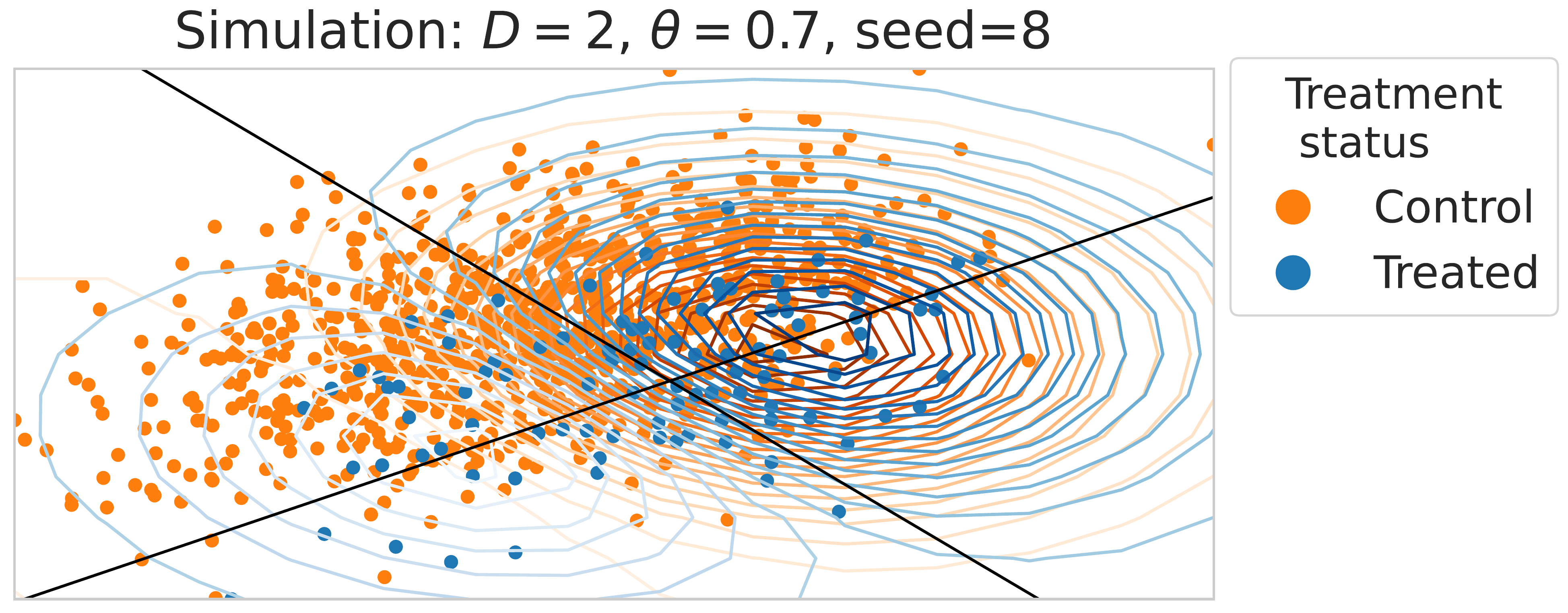}
    \end{minipage}
    \hfill
    \begin{minipage}{0.48\linewidth}
        \hfill%
        \includegraphics[width=0.98\linewidth,clip,trim={0 0 0 17cm}]{images/caussim_example_rs_gaussian=8_rs_rotation=8_ntv=0.37_D=2_overlap=0.7_p_A=0.1.pdf}
    \end{minipage}
    \caption{Example of the simulation setup in the input space with two
        knots --\emph{ie.}basis functions. The left panel gives
        views of the observations in feature space, while the right panel displays the
        two response surfaces on a 1D cut along the black lines drawn on
        the left panel.}
    \label{fig:simulation_examples}
\end{figure}

\paragraph{Family of candidate estimators}

We test model selection on a family of candidate estimators that
approximate imperfectly the data-generating process. To build such an
estimator in two steps, we first use a RBF expansion
similar as the one used for the data-generation generation process. Concretely,
we choose two random knots and apply a transformation of the raw data features
with the same Gaussian kernel used for the data-generation mechanism. This step
is referred as the featurization. Then, we fit a linear regression on
these
transformed features. We consider two ways of combining these steps for outcome
mode; using common nomenclature \cite{kunzel_metalearners_2019}, we refer
to these regression structures as different meta-learners which differ on how
they model, jointly or not, the treated and the non treated:
\begin{itemize}
    \item SLearner: A single learner for both population, taking the treatment as
          a supplementary covariate.

    \item SftLearner: A single set of basis functions is sampled at random for both
          populations, leading to a given feature space used to model both the treat and
          the non treated, then two
          separate different regressors are fitted on this representation.
    \item TLearner: Two completely different learners for each population, hence
          separate featurization and separate regressors.
\end{itemize}

We do not include more elaborated meta-learners such as R-learner
\cite{nie_quasioracle_2017} or X-learner
\cite{kunzel_metalearners_2019}. Our goal is not to have the best possible
learner but to have a variety of sub-optimal learners in order to compare the
different causal metrics. For the same reason, we did not include more powerful
outcome models such as random forests or boosting trees.

For the regression step, we fit a Ridge regression on the transformed features
with 6 different choices of the regularization parameter $\lambda \in [10^{-3},
        10^{-2}, 10^{-1}, 1, 10^{1}, 10^{2}]$, coupled with a TLearner or a SftLearner.
We sample 10 different random basis for the learning procedure and the
featurization yielding a family $\mathcal F$ of 120 candidate estimators.

\subsection{Semi-simulated datasets}

\paragraph{Datasets}\label{semi_simulated:datasets}

We also use semi-simulated datasets, where a known synthetic causal effect is
added to real --non synthetic-- covariate. We study datasets used in previous work
to evaluate causal inference:
\begin{description}
    \item[ACIC 2016] \cite{dorie_automated_2019}: The dataset is based on the
        Collaborative Perinatal Project \cite{niswander_women_1972}, a RCT
        conducted on a cohort of pregnant women to identify causes of infants’
        developmental disorders. The initial intervention was a child’s birth
        weight $(A = 1 \text{ if weight} < 2.5 kg)$, and outcome was the child’s
        IQ after a given follow-up period. The study contained $N=4\,802$ data
        points with $D=55$ features (5 binary, 27 count data, and 23
        continuous). They simulated 77 different setups with varying parameters
        for treatment and response generation models, treatment assignment
        probabilities, overlap, and interactions between treatment and
        covariates \footnote{Original R code available at
            \href{https://github.com/vdorie/aciccomp/tree/master/2016}{https://github.com/vdorie/aciccomp/tree/master/2016}
            to generate 77 simulations settings.}. We used 10 different seeds for
        every setup, totalizing 770 dataset instances.

    \item[ACIC 2018] \cite{shimoni_benchmarking_2018}: The raw covariates
        data comes from the Linked Births and Infant Deaths Database (LBIDD)
        \cite{macdorman_infant_1998} with $D=177$ covariates. Treatment and
        outcome models have been simulated with complex models to reflect
        different scenarii. The data do not provide the true propensity
        scores, so we evaluate only feasible metrics, which do not require this
        nuisance parameter. We used all 432 datasets\footnote{Using the scaling part of the data, from
            \href{https://github.com/IBM-HRL-MLHLS/IBM-Causal-Inference-Benchmarking-Framework}{github.com/IBM-HRL-MLHLS/IBM-Causal-Inference-Benchmarking-Framework}} of size $N=5\,000$.

    \item[Twins] \cite{louizos_causal_2017}: It is an augmentation of the
        real data on twin births and mortality rates in the USA from 1989-1991
        \cite{almond_costs_2005}. There are $N=11\,984$ samples (pairs of twins),
        and $D=50$ covariates\footnote{We obtained the dataset from
            \href{https://github.com/AMLab-Amsterdam/CEVAE/tree/master/datasets/TWINS}{https://github.com/AMLab-Amsterdam/CEVAE/tree/master/datasets/TWINS}}, The outcome is the mortality and the treatment is the
        weight of the heavier twin at birth. This is a "true" counterfactual dataset
        --as remarked in \cite{curth_really_2021}-- in the sense that we have
        both potential outcomes with each twin. They simulate the treatment with a
        sigmoid model based on GESTAT10 (number of gestation weeks before birth) and x
        the 45 other covariates:
        \begin{equation}
            \mathbf{t}_{i} \mid \mathbf{x}_{i}, \mathbf{z}_{i} \sim
            \operatorname{Bern}\left(\sigma\left(w_{o}^{\top}
            \mathbf{x}+w_{h}(\mathbf{z} / 10-0.1)\right)\right) \quad \text{with} \;
            w_{o} \sim \mathcal{N}(0,0.1 \cdot I),\; w_{h} \sim \mathcal{N}(5,0.1)
        \end{equation}
        We built upon this equation, adding a non-constant slope in the
        treatment sigmoid, allowing us to control the amount of overlap between
        treated and control populations.
        We sampled uniformly 1\,000 different overlap parameters between 0 and
        2.5, totalizing 1\,000 dataset instances. Unlike the previous datasets,
        only the overlap varies for these instances. The response surfaces are
        fixed by the original twin outcomes.
\end{description}

\paragraph{Family of candidate
    estimators}\label{semi_simulated:candidate_estimators}

For these three datasets, the family of candidate estimators are gradient
boosting trees for both the response surfaces and the treatment
\footnote{Scikit-learn regressor,
    \href{https://scikit-learn.org/stable/modules/generated/sklearn.ensemble.HistGradientBoostingRegressor.html}{HistGradientBoostingRegressor},
    and classifier,
    \href{https://scikit-learn.org/stable/modules/generated/sklearn.ensemble.HistGradientBoostingClassifier.html}{HistGradientBoostingClassifier}.}
with S-learner, learning rate
in $\{0.01, 0.1, 1\}$, and maximum number of leaf nodes in $\{25, 27, 30, 32,
    35, 40\}$ resulting in a family of size 18.

\paragraph{Nuisance estimators}

Drawing inspiration from the TMLE literature that uses combination of flexible
machine learning methods \cite{schuler_targeted_2017}, we use as models
for the nuisances $\check e$ (respectively $\check m$) a form of meta-learner: a
stacked estimator of ridge and boosting classifiers (respectively
regressions). We select hyper-parameters with randomized search on a validation set
$\mathcal{V}$ and keep them fix for model selection (detailed of the hyper
parameters in Appendix \ref{apd:experiments:nuisances_hp}). As extreme inverse
propensity weights induce high variance, we use clipping
\cite{swaminathan_counterfactual_2015, ionides_truncated_2008} to bound
$min(\check e, 1-\check e)$ away from 0 with a fixed $\eta=10^{-10}$, ensuring
strict overlap for numerical stability.

%\paragraph{Model selection procedure}\label{semi_simulated:selection_procedure}

%We used the same training procedure procedure as in
%\ref{paragraph:selection_procedure} (Details in supp.~mat.
%\ref{apd:selection_procedure_acic}).

\subsection{Measuring overlap between treated and non treated}\label{subsec:measuring_overlap}

%\idea{Overlap is important but not easily measurable}
Good overlap, or ``positivity'' between treated and control population is
crucial for causal inference as it is required by the positivity assumption
\ref{assumption:overlap} for causal identification. It is typically assessed by
qualitative methods using population histograms (as in Figure
\ref{fig:toy_example}) or side-by-side box plots, or quantitative approaches
such as Standardized Mean Difference
\cite{austin_introduction_2011,austin_moving_2015}. While these methods are
useful to decide if positivity holds, they do not summarize a dataset's overlap
in a single measure. Rather, we compute the divergence
between the population covariate distributions $\mathbb P(X|A=0)$ and $\mathbb
    P(X|A=1)$ to characterize the behavior of causal risk
\cite{damour_overlap_2020,johansson_generalization_2021}. We introduce the
Normalized Total Variation (NTV), a divergence based on the sole propensity
score. Details are given in Appendix \ref{apd:motivation_ntv}.

\begin{figure}[!b]
    \centering
    \includegraphics[width=\textwidth]{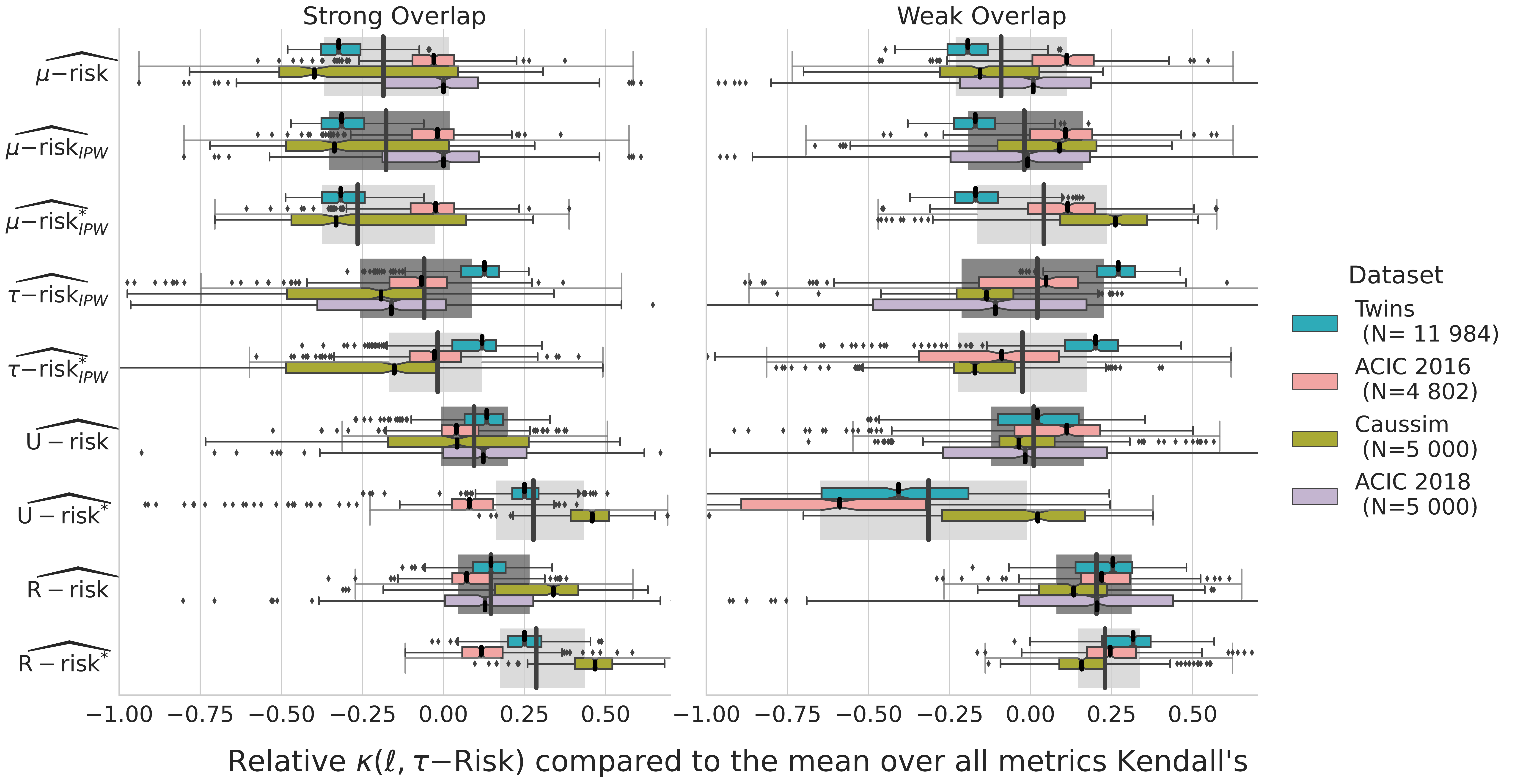}
    \hfill
    \caption{\textbf{The $R$-risk is the best metric}: Relative Kendall's $\tau$ agreement with $\tau\text{-risk}$, measured
        as the difference between each metric Kendall's $\tau$ and the mean kendall's
        $\tau$ over all metric:  $\kappa(\ell,\tau\mathrm{{-risk}}) - mean_{\ell} \big(\kappa(\ell,\tau\mathrm{{-rRisk}}) \big)$. Strong and Weak overlap correspond
        to the first and
        last tertiles of the overlap distribution measured with Normalized Total
        Varation eq. \ref{eq:ntv}.
        Appendix \ref{apd:experiments:additional_results} presents the same results
        measured with absolute Kendall's in Figure
        \ref{apd:fig:all_datasets_tau_risk_ranking_agreement} and with
        $\tau\mathrm{-risk}$ gains in Figure \ref{apd:all_datasets_normalized_bias_tau_risk_to_best_method}. Table \ref{apd:table:relative_kendalls_all_datasets} displays the median
        and IQR for the relative Kendall's results.}\label{fig:relative_kendalls_all_datasets}
\end{figure}

\subsection{Empirical results: factors driving good model selection across
    datasets}\label{empirical_study:results}

\paragraph{The $R\text{-risk}$ is the best metric}

Each metric ranks differently the candidate models. Figure
\ref{fig:relative_kendalls_all_datasets} shows the agreement between the
ideal ranking of methods given the oracle $\tau\text{-risk}$ and
the different causal metrics under evaluation. We measure this agreement
with a relative\footnote{To remove the variance
    across datasets (some datasets lead to easier model selection than
    others), we report values for one metric relative to the mean of all
    metrics for a given dataset instance: $\text{Relative} \, \kappa(\ell,\tau\mathrm{{-risk}})=
        \kappa(\ell,\tau\mathrm{{-risk}}) -
        mean_{\ell}\big(\kappa(\ell,\tau\mathrm{{-risk}})\big)$} Kendall tau $\kappa$ (eq.
\ref{eq:kendall_tau}) \cite{kendall_new_1938}.
Given the importance of overlap in
how well metrics approximate the oracle $\tau\text{-risk}$
(\autoref{apd:proofs:mu_risk_ipw_bound}), we separate
strong and weak overlap --defined as first and
last tertile of the Normalized Total Variation, eq. \ref{eq:ntv}.

Among all metrics the classical mean squared error (ie. factual
$\mu\text{-risk}$) is worse and reweighting it with propensity score
($\mu\text{-risk}_{IPW}$) does not bring much improvements. The
$R\text{-risk}$, which includes a model of mean outcome and propensity
scores, leads to the best performances. Interestingly, the
$U\text{-risk}$, which uses the same nuisances, has good performances for
strong overlap but deteriorates in weak overlap, probably due variance
inflation when dividing by extreme propensity scores.

Beyond the rankings, the differences in terms of absolute
ability to select the best model are large: The model selected by the R-risk
achieves a $\tau\text{-risk}$ only 1\% higher
than that of the best
possible candidate for strong overlap on Caussim, but selecting with
the $\mu\text{-risk}$ or $\mu\text{-risk}_{IPW}$ --as per machine-learning
practice-- leads to 10\% excess risk and using $\tau\text{-risk}_{IPW}$
--as in some causal-inference methods \citep{athey2016recursive,gutierrez_causal_2016}--leads to 100\% excess risk
(Figure
\ref{apd:all_datasets_normalized_bias_tau_risk_to_best_method}). Across
other datasets, the $R\text{-risk}$ consistently decreases the
risk compared to the $\mu\text{-risk}$:
0.1\% compared to 1\% on ACIC2016,  1\% compared to 20\% on ACIC 2018,
and 0.05\% compared to 1\% on Twins.

\paragraph{Model selection is harder in settings of low population
    overlap}

\begin{figure}[!h]
    \centering
    \includegraphics[width=0.9\textwidth]{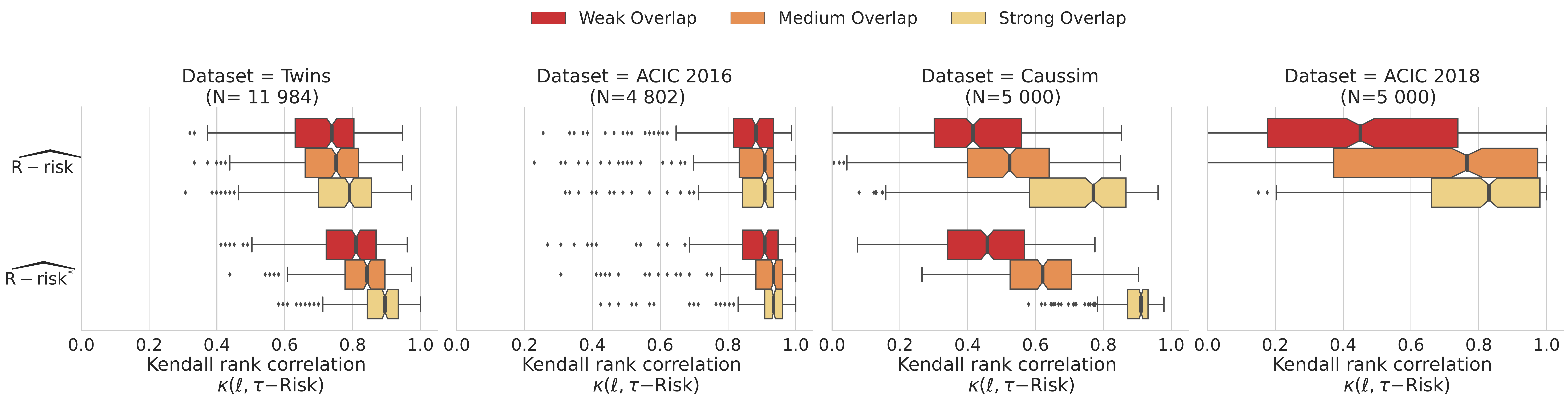}
    \hfill
    \caption{\textbf{Model selection is harder in settings of low population
            overlap}:
        Kendall's $\tau$ agreement with $\tau\text{-risk}$. Strong, medium and Weak overlap
        are the tertiles of the overlap distribution measured with NTV eq. \ref{eq:ntv}. Appendix \ref{apd:experiments:additional_results} presents results for all
        metrics in Figure \ref{apd:fig:all_datasets_overlap_effect} and in absolute
        Kendall's and continuous overlap values in Figure
            {\ref{apd:fig:all_datasets_tau_risk_ranking_agreement}}.}\label{fig:all_datasets_overlap_effect_r_risk}
\end{figure}

Model selection for
causal inference becomes more and more difficult with increasingly different
treated and control populations
(Figure \ref{fig:all_datasets_overlap_effect_r_risk}). The
absolute Kendall's coefficient correlation with $\tau\text{-risk}$
drops from values around 0.9 (excellent agreement with oracle selection) to 0.6 on
both Caussim and ACIC 2018
(Appendix \ref{apd:experiments:additional_results},
Figure {\ref{apd:fig:all_datasets_tau_risk_ranking_agreement}}).

\paragraph{Nuisances can be estimated on the same data as outcome models}

\begin{figure}[!h]
    %\centering
    \begin{subfigure}[b]{0.48\textwidth}
        %\centering
        %\caption{\textbf{Two sets procedure}}
        \includegraphics[width=2\textwidth]{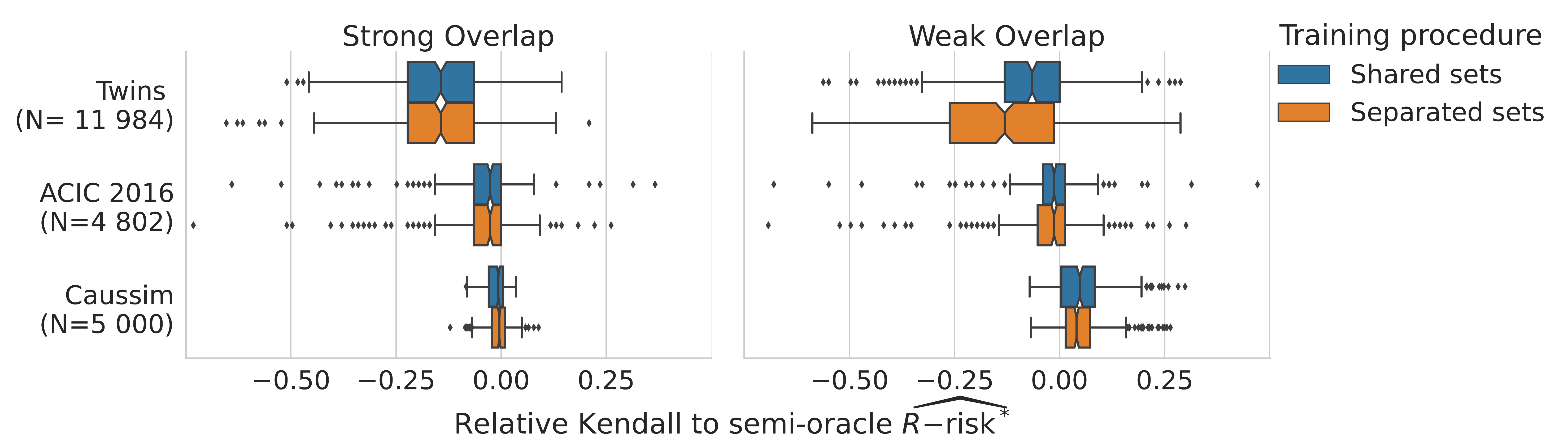}
    \end{subfigure}
    \hfill
    \caption{\textbf{Nuisances can be estimated on the same data as outcome
            models}: Results for the R-risk are similar between the
        \textcolor{MidnightBlue}{shared
            nuisances/candidate set} and
        the \textcolor{RedOrange}{separated nuisances set} procedures. Figure
        \ref{apd:fig:procedures_comparison_all_metrics} details results for all metrics.}\label
    {fig:procedures_comparison}
\end{figure}

Using the train set $\mathcal{T}$ both to fit the candidate estimator and the
nuisance estimates is a form of double dipping which can lead errors in
nuisances to be correlated to that of outcome models
\cite{nie_quasioracle_2017}. In theory, these correlations can bias model
selection and, strictly speaking, there are theoretical arguments
to split out a third separated data set --a ``nuisance set''-- to fit the
nuisance models. The drawback is that it depletes the data available for
model estimation and selection. However, Figure \ref{fig:procedures_comparison} shows that
there is no substantial differences between a procedure with a separated
nuisance set and the simpler shared nuisance-candidate set procedure
(results for all metrics in Figure
\ref{apd:fig:procedures_comparison_all_metrics}).

\paragraph{Stacked models are good overall estimators of nuisances}

\begin{figure}[!h]
    \begin{subfigure}[b]{0.48\textwidth}
        \centering
        \includegraphics[width=2\textwidth]{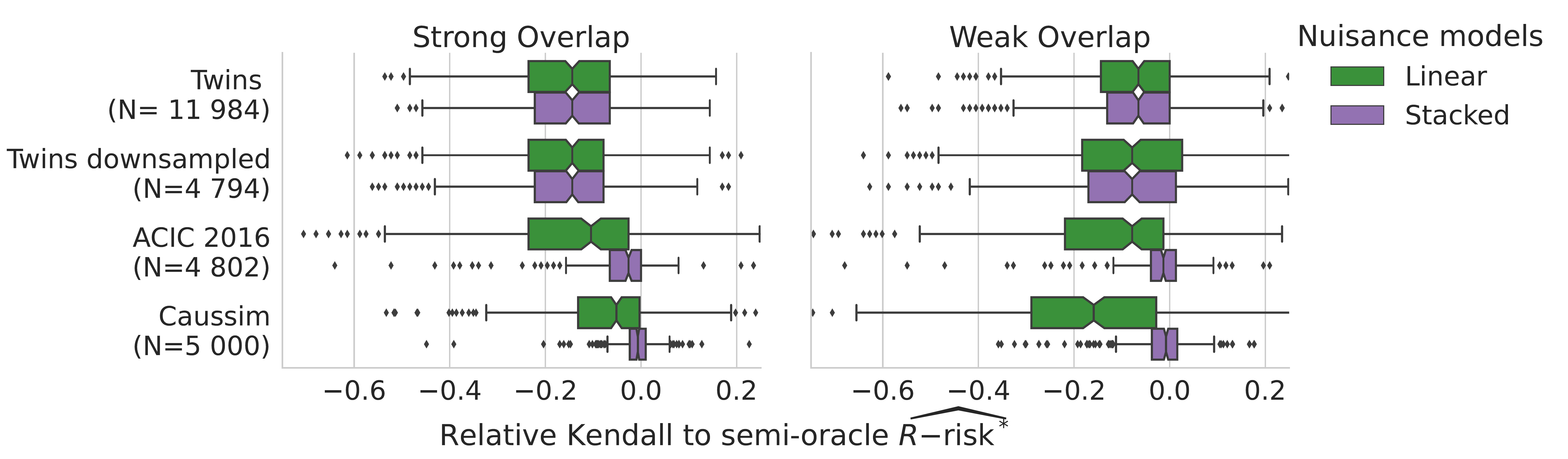}
    \end{subfigure}
    \hfill
    \caption{\textbf{\textcolor{DarkOrchid}{Stacked
                models} are good overall estimators of the nuisances}:
        Results are shown only for the
        R-risk. Details for every metrics are provided in Figure
        \ref{apd:fig:nuisances_comparison}. For Twins, where the true propensity
        model is linear, \textcolor{DarkOrchid}{stacked} and
        \textcolor{ForestGreen}{linear}
        estimations of the nuisances performs equivalently, even for a downsampled version
        (N=4794). }\label{fig:all_datasets_nuisances_comparison}
\end{figure}

Oracle versions of every risks recover more often the best estimator.
However, in configurations that we investigated,
stacked nuisances estimators (boosting and linear) lead to feasible
metrics with close performances to the oracles ones. This suggests that the
corresponding estimators recover well-enough the true nuisances.
One may wonder if it may be useful to use simpler models for the
nuisances, in particular in settings where there are less data or where
the true models are linear.
Figure \ref{fig:all_datasets_nuisances_comparison} compares causal model
selection estimating nuisances with stacked estimators or linear model.
It comprises the Twins data, where the true propensity model is linear,
and a downsampled version of this data, to study a situation favorable to
linear models. In these settings,
stacked and linear estimations of the nuisances performs equivalently.
Overall, Figure \ref{apd:fig:nuisances_comparison_twins} suggests that
to estimate nuisance it suffices to use adaptive models as built by
stacking linear models and gradient-boosted trees.
Figure \ref{apd:fig:nuisances_comparison} in the appendices, details
other causal metrics.

\paragraph{Use 90\% of the data to estimate outcome models, 10\% to
    select them}

For best causal modeling, the analyst often faces a compromise: given a finite
data sample, should she allocate most of the data to estimate the outcome model,
thus maximizing chances of achieving a high-quality outcome model but leaving
little data for model selection. Or, she could choose a bigger test set for
model selection and effect estimation. For causal model selection, there
is no established practice (as reviewed in Appendix \ref{apd:results:k_fold_choices}).

We investigate such tradeoff varying the ratio between train and test
data size. For this, we first split out 30\% of the data as a holdout set
$\mathcal{V}$ on which we use the oracle response functions to derive
silver-standard estimates of the causal quantities of interest. We then
use the standard estimation procedure on the remaining 70\% of the data,
splitting it into train $\mathcal{T}$ and test $\mathcal{S}$ of varying
sizes. We finally measure the error between this estimate and the
silver-standard one.

We consider two different analytic goals: estimating a average
treatment effect --a single number used for policy making-- and a
CATE --A full model of the treatment effect as a function of covariates
$X$. Given that the latter is a much more complex object than the former,
the optimal train/test ratio might vary. To measure errors, we use for
the ATE the relative absolute ATE bias between the ATE computed with the
selected outcome model on the test set, and the true ATE as evaluated on
the holdout set $\mathcal{V}$. For the CATE, we compare the
$\tau\text{-risk}$
of the best selected model applied on the holdout set $\mathcal{V}$. We explore this trade-off for the ACIC 2016 dataset and the R-risk.

\begin{figure}[!t]
    %\centering
    \begin{minipage}{.5\textwidth}
        \centerline{\textbf{a) CATE estimation error}}

        \includegraphics[height=.35\textwidth]{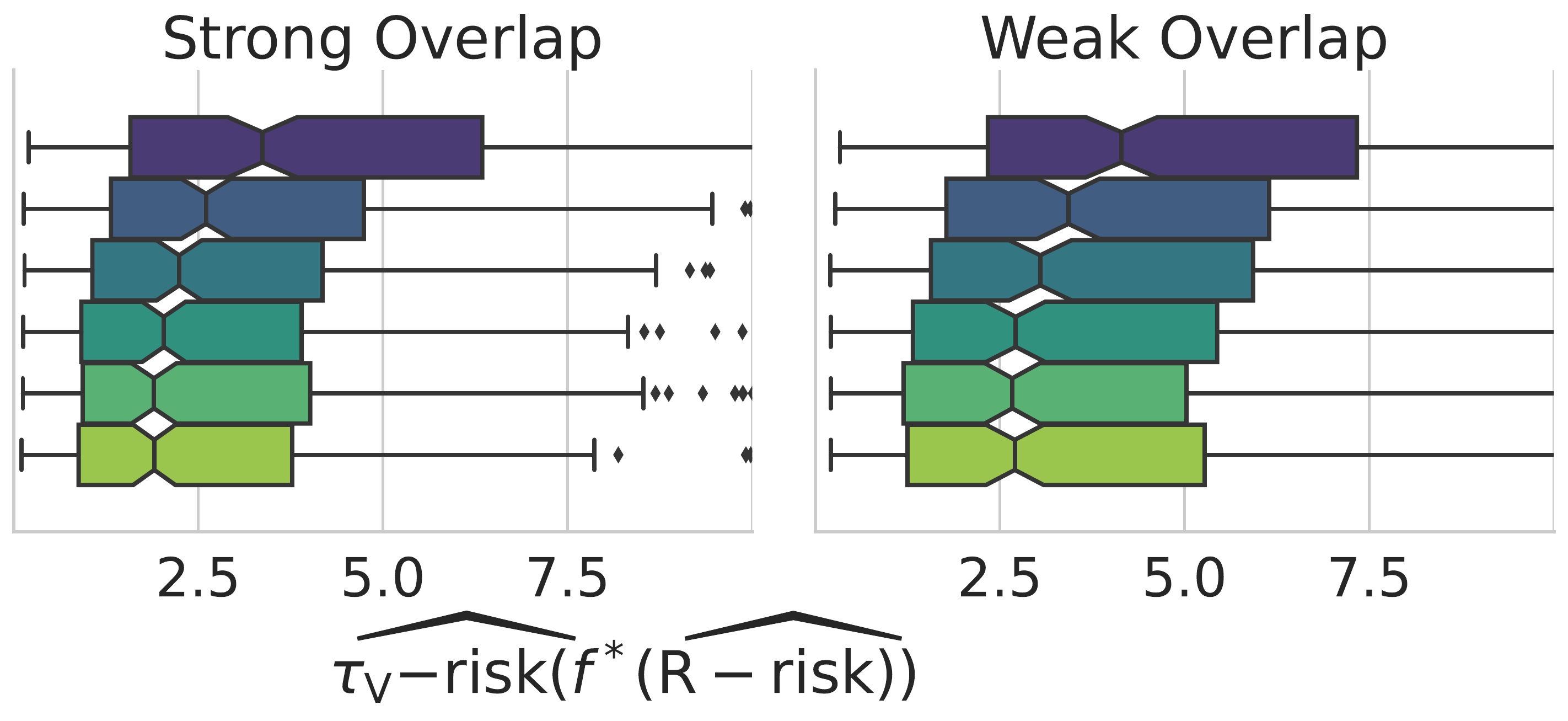}
    \end{minipage}
    \begin{minipage}{.5\textwidth}
        %\centering
        \centerline{\textbf{b) ATE estimation error}}
        \includegraphics[height=.35\textwidth]{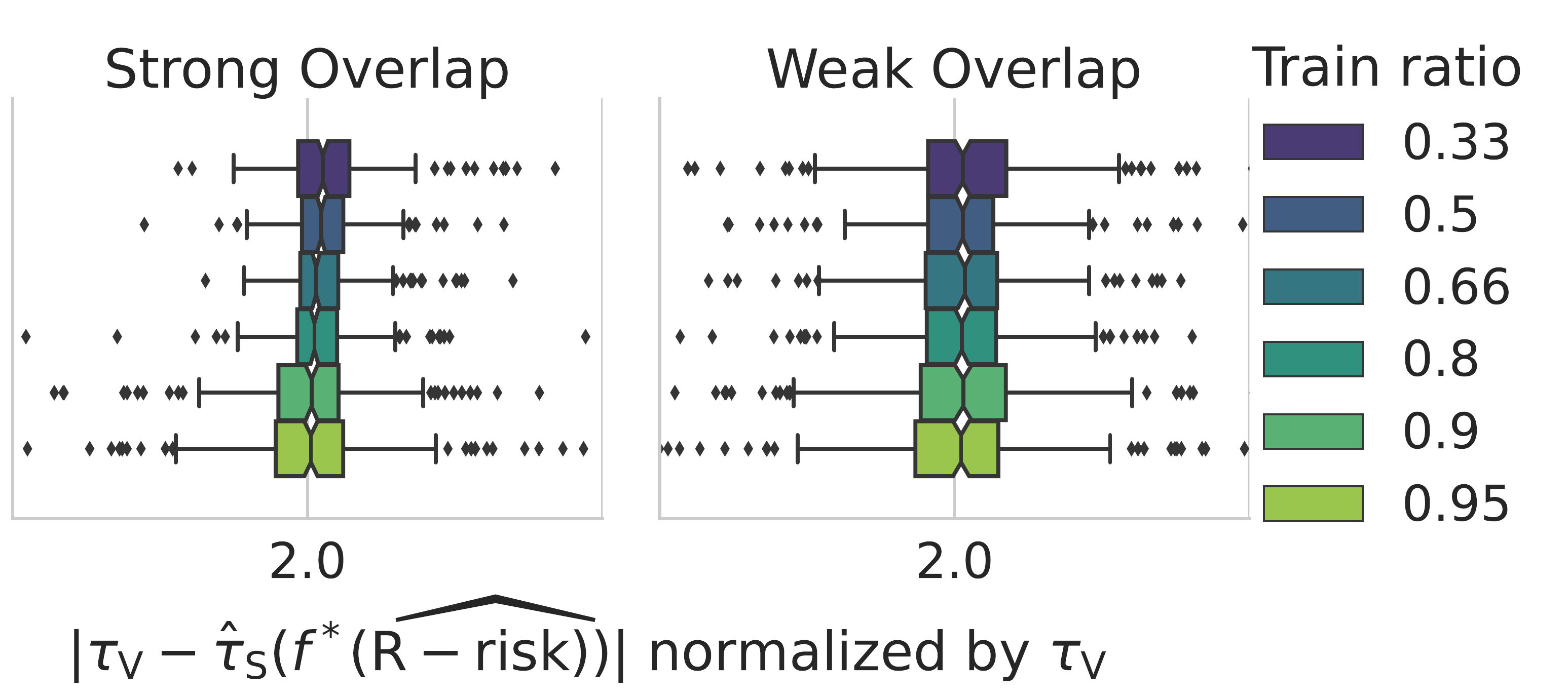}
    \end{minipage}%
    %\hspace{0.3\textwidth}
    \caption{\textbf{a) For CATE, a train/test ratio of 0.9/0.1 appears as a good
            trade-off.} b) For ATE, there is a small signal pointing also to
        0.9/0.1 (K=10).
        for ATE. Experiences on 10 replications of all 78 instances of the ACIC 2016
        data.}\label{fig:train_test_ratio}
\end{figure}

Figure \ref{fig:train_test_ratio} shows that a train/test ratio of
0.9/0.1 (K=10) appears as a good trade-off for CATE and ATE estimation,
thought there is little difference with a split of 0.8/0.2 (K=5).

\section{Discussion and conclusion}\label{sec:discussion}

Predictive models are increasingly used to reason about causal effects.
Our results highlight that they should be selected, validated, and tuned
using different procedures and error measures than those classically used
to assess prediction (estimating the so-called $\mu\text{-risk}$).
Rather, selecting the best outcome model according to the $R\text{-risk}$
(eq.\,\autoref{def:r_risk}) leads to more valid causal estimates.
Estimating this risk requires a markedly more complex procedure than
standard cross-validation used \emph{e.g.} in machine learning: it involves
fitting nuisance models necessary for model evaluation, though our
empirical results show that these can be learned on the same set of data
as the outcome model evaluated.
A poor estimation of the nuisance models may compromise the benefits of
the more complex $R\text{-risk}$ (as shown in Figure
\ref{fig:all_datasets_nuisances_comparison}). However controlling and selecting
these latter models is easier because they are associated to errors on
observed distributions. Our empirical results show that when selecting
these models in a flexible family of models the $R\text{-risk}$ dominates
simpler risks for model selection.
Results show that going from an oracle $R\text{-risk}$ --where the
nuisances are known-- to a feasible $R\text{-risk}$ --where the nuisances
are estimated-- decreases only very slightly the model-selection
performance of the $R\text{-risk}$. This may be explained by
theoretical results that suggest that estimation errors on both
nuisances partly compensate out in the
$R\text{-risk}$\cite{daniel2018double,naimi2021challenges,kennedy2020optimal,nie_quasioracle_2017,chernozhukov_double_2018}.
% 
%The usage of the $R\text{-risk}$ can also be understood as a
%$\tau \text{-risk}$ reweighted by the propensity score
%(prop \autoref{theory:prop:r_risk_rewrite}).
%

For strong overlap, the conventional procedure ($\mu\text{-risk}$) appears theoretically motivated
(\autoref{theory:mu_risk_ipw_bound}), however empirical results show that
even in this regime the $R\text{-risk}$ brings a sizeable benefit,
in agreement with \citet{schuler_comparison_2018}.

Our results points to K=5 or 10 for the choice of the number of folds used in
cross-validation for CATE and ATE, aligned with previous empirical evidence for
ATE \cite{chernozhukov_double_2018}.

\paragraph{Extension to binary outcome}
While we focused on continuous outcomes, in medicine, the target outcome
is often a categorical variable such as mortality status or diagnosis. In
this case, it may be interesting to focus on other estimands than the
Average Treatment Effect $\mathbb{E}[Y(1)] -\mathbb{E}[Y(0)] $, for instance the relative risk
$\frac{\mathbb P(Y(1) = 1)}{\mathbb P(Y(0) = 1)}$
or the odd ratio, $\frac{\mathbb P(Y(1) = 1) / [1 - \mathbb P(Y(1)
    =1)]}{\mathbb P(Y(0) = 1) / [1 - \mathbb P(Y(0) = 1]}$ are often used
\cite{austin2017estimating}. While the odds ratio is natural for
case-control studies \cite{rothman2008case}, a good choice of measure
can reduce heterogeneity \cite{colnet2023risk}.
Using as an estimand the log of
these values is suitable to additive models (for reasoning or noise
assumptions). In the log domain, the relative risk or the odds ratio are
written as a difference, as the ATE: $\log \mathbb{P}(Y(1)=1) - \log
    \mathbb{P}(Y(0=1)$ or $\log (\mathbb{P}(Y(1)=1) / [1 -\mathbb{P}(Y(1)=1)]
    ) - \log \mathbb{P}(Y(0=1) / [1 -\mathbb{P}(Y(0)=1)]$. Hence, the
framework studied here (\autoref{sec:neyman_rubin}) can directly apply.
It is particularly easy for the log odds ratio, as it is the output of a
logistic regression or any model with a cross-entropy loss.

\paragraph{Going further}
% Estimation of the nuisances, calibration
The $R\text{-risk}$ needs good estimation of nuisance models. The
propensity score $e$ calls for a control on the estimation of the
individual posterior probability. We have used the Brier score to select
these models, as it is minimized by the true individual probability. Regarding model-selection for
propensity score, an easy mistake is to use expected calibration errors
popular in machine learning
\cite{platt_probabilistic_1999,zadrozny_obtaining_2001,niculescu-mizil_predicting_2005,minderer_revisiting_2021}
as these select not for the individual posterior probability but for an
aggregate error rate \cite{perez2022beyond}. An
open question is whether a better metric than the brier score can be designed that controls for $e \, (1 - e)$, the quantity used in
the $R\text{-risk}$, rather than $e$.

%\idea{Treatment heterogeneity seems important.}
The quality of model selection varies substantially from one
data-generating mechanism to another. The overlap appears as an important
parameter: when the treated and untreated, causal model selection is very
hard. However, remaining variance in the empirical results suggests that
other parameters of the data generation processes come into play.
Intuitively, the complexity of the response surfaces and the treatment
heterogeneity interact with overlap violations: when extrapolations to
weak-overlap regions is hard, causal model selection is hard.

Nevertheless, from a practical perspective, our study establishes that
the $R\text{-risk}$ is the best option to select predictive models for
causal inference, without requiring assumptions on the data-generating
mechanism, the amount of data at hand, or the specific estimators used to
build predictive models.

% XXX: need a final positive
% remark
% Maybe that as our work has no
% requirement on the learning procedure,
% it contributes to ensuring the
% validity of causal inference on complex databases such as clinical
% records with a lot of open-ended text that must be analyzed with
% language models.

% Sentence stolen, adapt
% unlike parametric regression, machine learning (ML) methods do not
% generally require precise
% knowledge of the true data generating mechanisms

\section*{Acknowledgments}

We acknowledge fruitful discussions with Bénédicte Colnet.

\subsection*{Financial disclosure}

None reported.

\subsection*{Conflict of interest}

The authors declare no potential conflict of interests.

%\if@Mybibrefstyle%
\printbibliography
%\fi if using njdnatbib.sty \bibliography{bibliography}%

\clearpage

\appendix

%The following supporting information is available as part of the online article:

\section{Variability of ATE estimation on ACIC
  2016}\label{apd:toy_example:acic_2016_ate_variability}

Figure \ref{fig:acic_2016_ate_heterogeneity} shows ATE estimations for six
different models used in g-computation estimators on the 76 configurations of
the ACIC 2016 dataset. Outcome models are fitted on half of the data and
inference is done on the other half --ie. train/test with a split ratio of 0.5.
For each configuration, and each model, this train test split was repeated ten
times, yielding non parametric variance estimates
\cite{bouthillier_accounting_2021}.

Outcome models are implemented with
\href{https://scikit-learn.org/stable/}{scikit-learn}
\cite{pedregosa_scikitlearn_2011} and the following hyper-parameters:

\begin{table}[h!]
    \centering
    \begin{tabular}{llll}
        \toprule
        Outcome Model                                  & Hyper-parameters grid
        \\
        \midrule
        Random Forests                                 & Max depth: [2,
        10]                                                                    \\

        Ridge regression without treatment interaction & Ridge regularization:
        [0.1]                                                                  \\

        Ridge regression with treatment interaction    & Ridge regularization:
        [0.1]                                                                  \\
        \bottomrule
    \end{tabular}
    \caption{Hyper-parameters grid used for ACIC 2016 ATE variability}
    \label{apd:toy_example:acic_2016_ate_variability:table}
\end{table}

\section{Causal assumptions}\label{apd:causal_assumptions}

We assume the following four assumptions, referred as strong ignorability and
necessary to assure identifiability of the causal estimands with observational
data \cite{rubin_causal_2005}:
\begin{assumption}[Unconfoundedness]\label{assumption:ignorability}
    \begin{equation*}\label{eq:ignorability}
        \{Y(0), Y(1) \} \indep A | X
    \end{equation*}
    This condition --also called ignorability-- is equivalent to the conditional
    independence on $e(X)$ \cite{rosenbaum_central_1983}: $\{Y(0), Y(1) \}
        \indep  A | e(X)$.
\end{assumption}

\begin{assumption}[Overlap, also known as Positivity)]\label{assumption:overlap}
    \begin{equation*}\label{eq:overlap}
        \eta < e(x) < 1 - \eta \quad \forall x \in \mathcal X \text{ and some } \eta > 0
    \end{equation*}
    The treatment is not perfectly predictable. Or with different words, every
    patient has a chance to be treated and not to be treated. For a given set of
    covariates, we need examples of both to recover the ATE.
\end{assumption}

As noted by \cite{damour_overlap_2020}, the choice of covariates $X$ can
be viewed as a trade-off between these two central assumptions. A bigger
covariates set generally reinforces the ignorability assumption. In the
contrary, overlap can be weakened by large $\mathcal{X}$ because of the
potential inclusion of instruments: variables only linked to the treatment which
could lead to arbitrarily small propensity scores.

% remark: There is a major counter example to these colliders (variables which
% are caused by both the outcome and the treatment),

\begin{assumption}[Consistency]\label{assumption:consistency} The observed
    outcome is the potential outcome of the assigned treatment:
    \begin{equation*}\label{eq:consistancy}
        Y = A \, Y(1) + (1-A) \, Y(0)
    \end{equation*}
    Here, we assume that the intervention $A$ has been well defined. This
    assumption focuses on the design of the experiment. It clearly states the link
    between the observed outcome and the potential outcomes through the
    intervention \cite{hernan_causal_2020}.
\end{assumption}

\begin{assumption}[Generalization]\label{assumption:generalization} The training
    data on which we build the estimator and the test data on which we make the
    estimation are drawn from the same distribution $\mathcal D^*$, also known as
    the ``no covariate shift'' assumption \cite{jesson_identifying_2020}.
\end{assumption}

\section{Proofs: Links between feasible and oracle risks}\label{apd:proofs}

\subsection{Upper bound of $\tau\text{-risk}$ with
    $\mu\text{-risk}_{IPW}$}\label{apd:proofs:mu_risk_ipw_bound}

For the bound with the $\mu\text{-risk}_{IPW}$, we will decompose the CATE risk
on each factual population risks:

\begin{definition}[Population Factual $\mu\text{-risk}$]\label{mu_risk_a}
    \cite{shalit_estimating_2017}
    \begin{equation*}
        \mu\text{-risk}_{a}(f)= \int_{\mathcal Y \times \mathcal X} (y-f(x ; A=a))^{2}  p(y ; x=x \mid A=a) \; dy dx
    \end{equation*}
\end{definition}

Applying Bayes rule, we can decompose the $\mu\text{-risk}$ on each
intervention:
\begin{equation*}
    \mu\text{-risk}(f)
    =p_{A} \,\mu\text{-risk}_{1}(f)+\left(1-p_{A}\right) \,\mu\text{-risk}_{0}(f)
    \text{with } p_A=\mathbb P(A=1)
\end{equation*}

These definitions allows to state a intermediary result on each population:
\begin{lemma}[Mean-variance decomposition]\label{apd:proofs:mu_risk_ipw_link_mu}
    We need a reweighted version of the classical mean-variance decomposition.

    For an outcome model $f: x \times A \rightarrow \mathcal X$. Let the inverse
    propensity weighting function $w(a ; x)=a e(x)^{-1}+(1-a)(1-e(x))^{-1}$.
    \begin{align*}
         & \int_{\mathcal X}(\mu_{1}(x)-f(x ; 1))^{2} p(x) dx  = p_{A} \mu\text{-risk}_{IPW, 1}(w, f)  -\sigma^{2}_{Bayes}(1)
    \end{align*}
    And
    \begin{align*}
         & \int_{\mathcal X}(\mu_{0}(x)-f(x; 0))^{2} p(x) dx   = (1-p_A) \mu\text{-risk}_{IPW, 0}(w, f)  -\sigma^{2}_{Bayes}(0)
    \end{align*}

    \begin{proof}
        \begin{align*}
             & p_{A} \mu\text{-risk}_{IPW, 1}(w, f)  = \int_{\mathcal X \times \mathcal Y} \frac{1}{e(x)}(y-f(x ; 1))^{2} p(y \mid x ; A=1) p(x ; A=1) d y d x                                                      \\
             & = \int_{\mathcal X \times \mathcal Y} (y-f(x ; 1))^{2} p(y \mid x ; A=1) \frac{p(x ; A=1)}{p(x ; A=1)}p(x)dy dx                                                                                      \\
             & = \int_{\mathcal X \times \mathcal Y} \big[(y-\mu_1(x))^{2}+\left(\mu_{1}(x)-f(x ; 1)\right)^{2} + 2\left(y-\mu_{1}(x)\right)\left(\mu_{1}(x)-f(x, 1)\right) \big] p(y \mid x ; A=1) p(x) d y d x    \\
             & =\int_{\mathcal X} \big [ \int_{\mathcal Y} (y-\mu_1(x))^{2} p(y \mid x ; A=1) dy\big ] p(x)dx + \int_{\mathcal X \times \mathcal Y} \left(\mu_{1}(x)-f(x ; 1)\right)^{2} p(x)p(y \mid x ; A=1)dx dy \\
             & + \qquad 2 \int_{\mathcal X} \big [ \int_{\mathcal Y} \left(y-\mu_{1}(x)\right) p(y \mid x ; A=1) dy \big ] \left(\mu_{1}(x)-f(x, 1)\right)p(x)dx                                                    \\
             & =\int_{\mathcal X} \sigma_{y}^{2}(x, 1) p(x) d x +\int_{\mathcal X} \left(\mu_{1}(x)-f(x ; 1)\right)^{2} p(x) d x+0
        \end{align*}
    \end{proof}
\end{lemma}

\begin{proposition*}[Upper bound with mu-IPW]\label{apd:proofs:prop:upper_bound}
    Let f be a given outcome model, let the weighting function $w$ be the Inverse
    Propensity Weight $w(x; a) = \frac{a}{e(x)} + \frac{1-a}{1-e(x)}$. Then, under
    overlap (assumption \ref{assumption:overlap}),
    \begin{align*}
        \tau\text{-risk}(f) \leq & \; 2 \, \mu\text{-risk}_{IPW}(w, f)  \; - 2 \, (\sigma^2_{Bayes}(1) +  \sigma^2_{Bayes}(0))
    \end{align*}

    \begin{proof}
        \begin{align*}
             & \tau\text{-risk}(f) =\int_{\mathcal X}(\mu_{1}(x)-\mu_{0}(x)-(f(x ; 1)-f(x ; 0))^{2} p(x) d x
        \end{align*}
        By the triangle inequality $(u+v)^2 \leq 2(u^2 + v^2)$:
        \begin{align*}
             & \tau\text{-risk}(f) \leq
            2 \int_{\mathcal X}\big[\left(\mu_{1}(x)-f(x ; 1)\right)^{2}+\left(\mu_{0}(x)-f(x ; 0)\right)^{2}\big] p(x) d x \\
        \end{align*}
        Applying Lemma \ref{apd:proofs:mu_risk_ipw_link_mu},
        \begin{align*}
             & \tau\text{-risk}(f) \leq 2\big[p_A \mu\text{-risk}_{IPW, 1}(w, f) +(1-p_A) \mu\text{-risk}_{IPW, 0}(w, f)(w, f)\big] -2(\sigma^{2}_{Bayes}(0) + \sigma^{2}_{Bayes}(1)) \\
             & = 2 \mu\text{-risk}_{IPW}(w, f)-2(\sigma^{2}_{Bayes}(0) + \sigma^{2}_{Bayes}(1))
        \end{align*}
    \end{proof}
\end{proposition*}

\subsection{Reformulation of the $R\text{-risk}$ as reweighted
    $\tau\text{-risk}$}\label{apd:proofs:r_risk_rewrite}

\begin{proposition*}[$R\text{-risk}$ as reweighted $\tau
            \text{-risk}$]\label{apd:proofs:prop:r_risk_rewrite}

    \begin{proof}

        We consider the R-decomposition: \cite{robinson_rootnconsistent_1988},
        \begin{equation}\label{apd:eq:r_decomposition}
            y(a) = m(x) + \big( a - e(x) \big) \tau(x) + \varepsilon(x; a)
        \end{equation}
        Where $\mathbb E[\varepsilon(X; A)|X, A] = 0$ We can use it as plug in the
        $R\text{-risk}$ formula:

        \begin{align*}
             & R\text {-risk}(f) =\int_{\mathcal{Y} \times \mathcal{X} \times \mathcal{A}}[(y-m(x))-\big(a-e(x)\big) \tau_f(x)]^{2} p(y ; x ; a) d y d x d a                                     \\
             & =\int_{\mathcal{Y} \times \mathcal{X} \times \mathcal{A}} \left[\big(a-e(x)\big)\tau(x)+\varepsilon(x ; a)-\big(a-e(x)\big) \tau_f(x)\right]^{2} p(y ; x ; a) d y d x da          \\
             & =\int_{\mathcal{X} \times \mathcal{A}}\big(a-e(x)\big)^{2}\big(\tau(x)- \tau_f(x)\big)^{2} p(x ; a) d x d a                                                                       \\
             & + 2  \int_{\mathcal{Y} \times \mathcal{X} \times \mathcal{A}}\big(a-e(x)\big)\big(\tau(x)-\tau_f(x)\big)  \int_{\mathcal{Y}} \varepsilon(x ; a) p(y \mid x ; a) d y p(x ; a)dx da \\
             & +\int_{\mathcal{X} \times \mathcal{A}} \int_{\mathcal{Y}} \varepsilon^{2}(x ; a) p(y \mid x ; a) d y p(x ; a) d x d a
        \end{align*}

        The first term can be decomposed on control and treated populations to force
        $e(x)$ to appear:
        \begin{align*}
             & \int_{\mathcal{X}}\big(\tau(x)-\tau_f(x)\big)^{2}\left[e(x)^{2}p(x;0) + \big(1-e(x)\big)^{2} p(x;1)\right] d x                    \\
             & =\int_{\mathcal{X}}\big(\tau(x)-\tau_f(x)\big)^{2}  \left[e(x)^{2}\big(1-e(x)\big)p(x) + \big(1-e(x)\big)^{2}e(x) p(x)\right] d x \\
             & =\int_{\mathcal{X}}(\tau(x)-\tau_f(x))^{2}(1-e(x)) e(x)[1-e(x)+e(x)] p(x) d x                                                     \\ &=\int_{\mathcal{X}}(\tau(x)-\tau_f(x))^{2}(1-e(x)) e(x) p(x) d x.
        \end{align*}

        The second term is null since, $\mathbb E[\varepsilon(x, a) |X, A]=0$.

        The third term corresponds to the modulated residuals \ref{eq:residuals} :
        $\tilde{\sigma}_B^2(0) + \tilde{\sigma}_B^2(1)$

    \end{proof}
\end{proposition*}

\section{Measuring overlap}\label{apd:motivation_ntv}

\paragraph{Motivation of the Normalized Total Variation}
%\idea{A simplier measure, NTV seems to capture what we want}
Computing overlap when working only on samples of the observed distribution,
outside of simulation, requires a sophisticated estimator of discrepancy
between distributions, as two data points never have the same exact set of
features. Maximum Mean Discrepancy \cite{gretton2012kernel} is typically
used in the context of causal inference
\cite{shalit_estimating_2017,johansson_generalization_2021}. However it
needs a kernel, typically Gaussian, to extrapolate across neighboring
observations. We prefer avoiding the need to specify such a kernel, as it must
be adapted to the data which is tricky with categorical or non-Gaussian
features, a common situation for medical data.

For simulated and some semi-simulated data, we have access to the probability of
treatment for each data point, which sample both densities in the same data
point. Thus, we can directly use distribution discrepancy measures and rely on
the Normalized Total Variation (NTV) distance to measure the overlap between the
treated and control propensities. This is the empirical measure of the total
variation distance \cite{sriperumbudur_integral_2009} between the distributions,
$TV(\mathbb{P}(X|A=1), \mathbb{P}(X|A=0))$. As we have both distribution sampled
on the same points, we can rewrite it a sole function of the propensity score, a
low dimensional score more tractable than the full distribution $\mathbb
    P(X|A)$:

\begin{equation}\label{eq:ntv}
    \widehat{NTV}(e, 1-e) = \frac{1}{2N} \sum_{i =1}^{N} \big |\frac{e(x_i)}{p_A}-\frac{1-e(x_i)}{1-{p_A}} \big |
\end{equation}

Formally, we can rewrite NTV as the Total Variation distance
between the two population distributions. For a population $O = (Y(A), X, A)
    \sim \mathcal{D}$:

\begin{align*}
    NTV(O) & = \frac{1}{2N} \sum_{i =1}^{N} \big | \frac{e(x_i)}{p_A}-\frac{1-e(x_i)}{1-{p_A}} \big|         \\
           & = \frac{1}{2N} \sum_{i =1}^{N} \big|\frac{P(A=1|X=x_i)}{p_A}-\frac{P(A=0|X=x_i)}{1-{p_A}} \big|
\end{align*}

Thus NTV approximates the following quantity in expectation over the data
distribution $\mathcal{D}$:

\begin{align*}
    NTV(\mathcal{D}) & = \int_{\mathcal{X}} \big | \frac{p(A=1|X=x)}{p_A}-\frac{p(A=0|X=x)}{1-{p_A}} \big | p(x)dx \\
                     & = \int_{\mathcal{X}} \big | \frac{p(A=1, X=x)}{p_A}-\frac{p(A=0, X=x)}{1-{p_A}} \big |dx    \\
                     & = \int_{\mathcal{X}} \big | p(X=x|A=1)- p(X=x|A=0) \big |dx
\end{align*}

For countable sets, this expression corresponds to the Total Variation distance
between treated and control populations covariate distributions : $TV(p_0(x),
    p_1(x))$.

\begin{figure}
    \begin{subfigure}[b]{\textwidth}
        \centering
        \caption{\textbf{Uncalibrated classifiers}}
        \includegraphics[width=\linewidth]{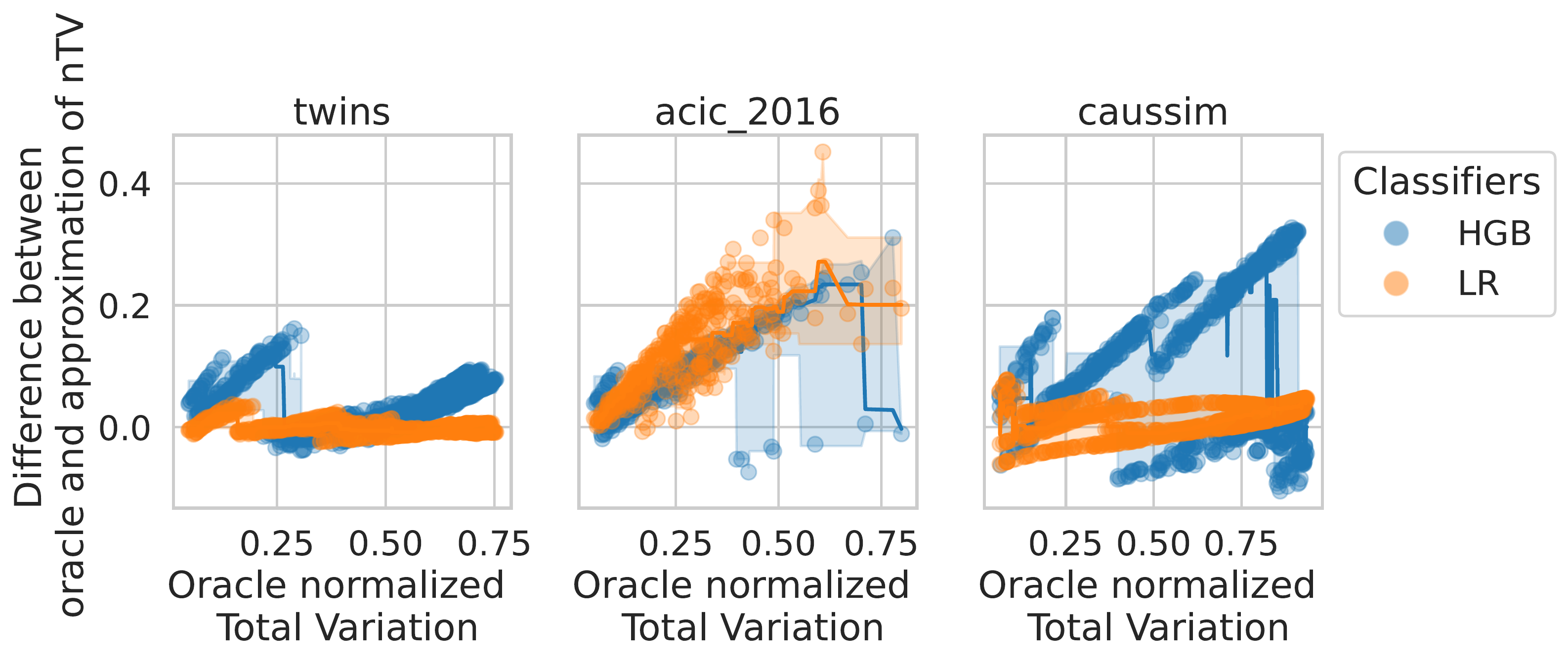}
    \end{subfigure}
    \begin{subfigure}[b]{\textwidth}
        \centering
        \caption{\textbf{Calibrated
                classifiers}}
        \includegraphics[width=\linewidth]{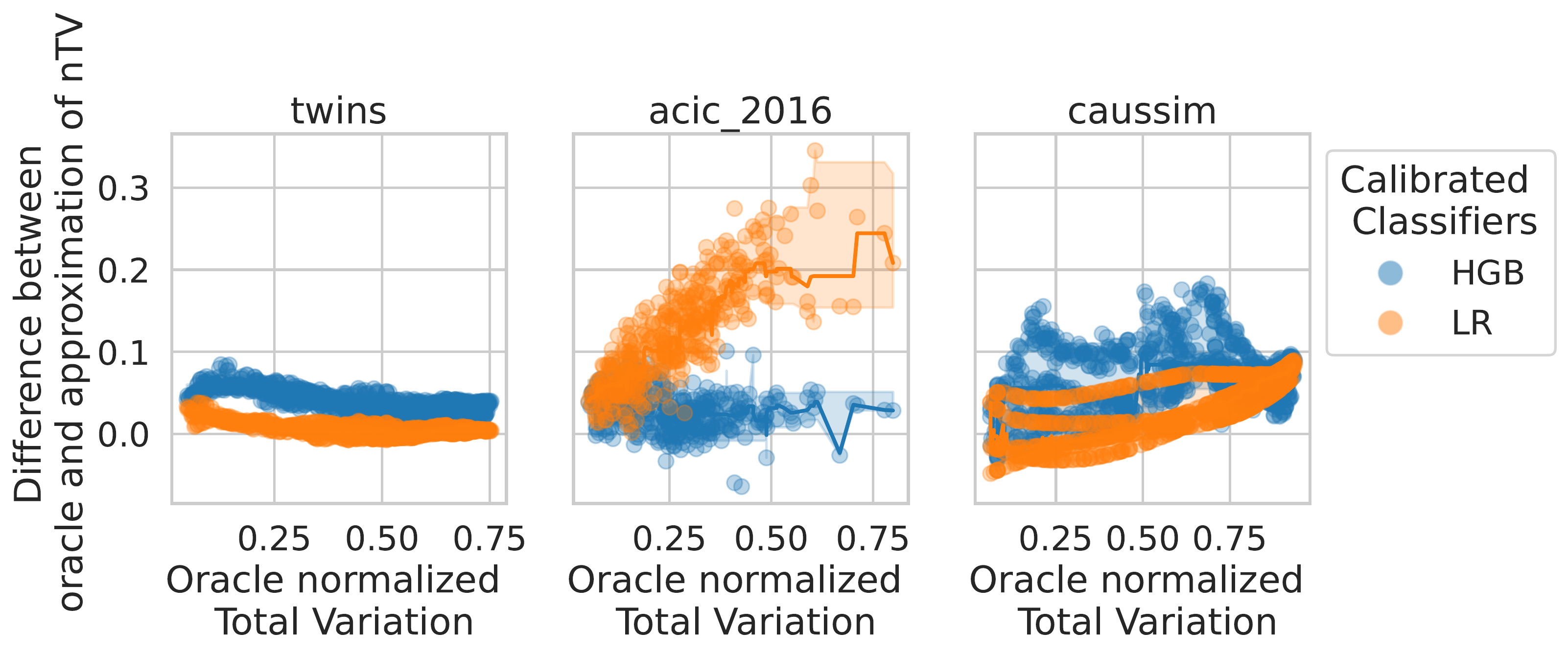}
    \end{subfigure}
    \caption{a) Without calibration, estimation of NTV is not trivial even for
        boosting models. b) Calibrated classifiers are able to recover the true
        Normalized Total Variation for all datasets where it is
        available.}\label{apd:overlap:ntv_approximation}
\end{figure}

\paragraph{Measuring overlap without the oracle propensity scores:} For ACIC
2018, or for non-simulated data, the true propensity scores are not known. To
measure overlap, we rely on flexible estimations of the Normalized Total
Variation, using gradient boosting trees to approximate the propensity score.
Empirical arguments for this plug-in approach is given in Figure
\ref{apd:overlap:ntv_approximation}.

\paragraph{Empirical arguments}

\begin{figure}
    \centering
    \caption{NTV recovers well the overlap settings described in the ACIC paper
        \cite{dorie_automated_2019}}\label{apd:overlap:penalized_overlap}
    \includegraphics[width=0.7\linewidth]{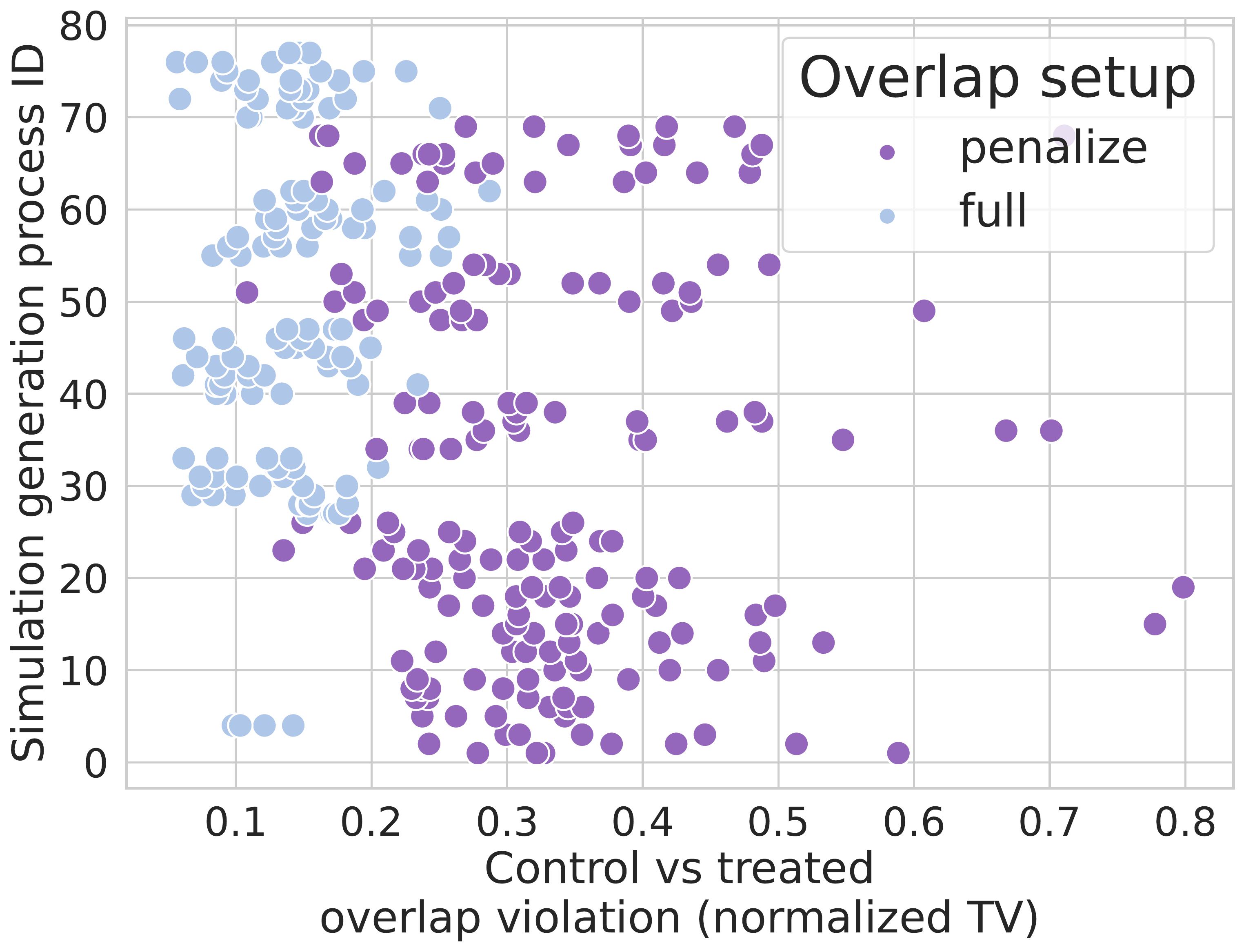}
\end{figure}

\begin{figure}[htbp]
    \centering
    \caption{Good correlation between overlap measured as normalized Total
        Variation and Maximum Mean Discrepancy (200 sampled Caussim
        datasets)}\label{apd:overlap:caussim:mmd_vs_ntv}
    \includegraphics[width=0.7\linewidth]{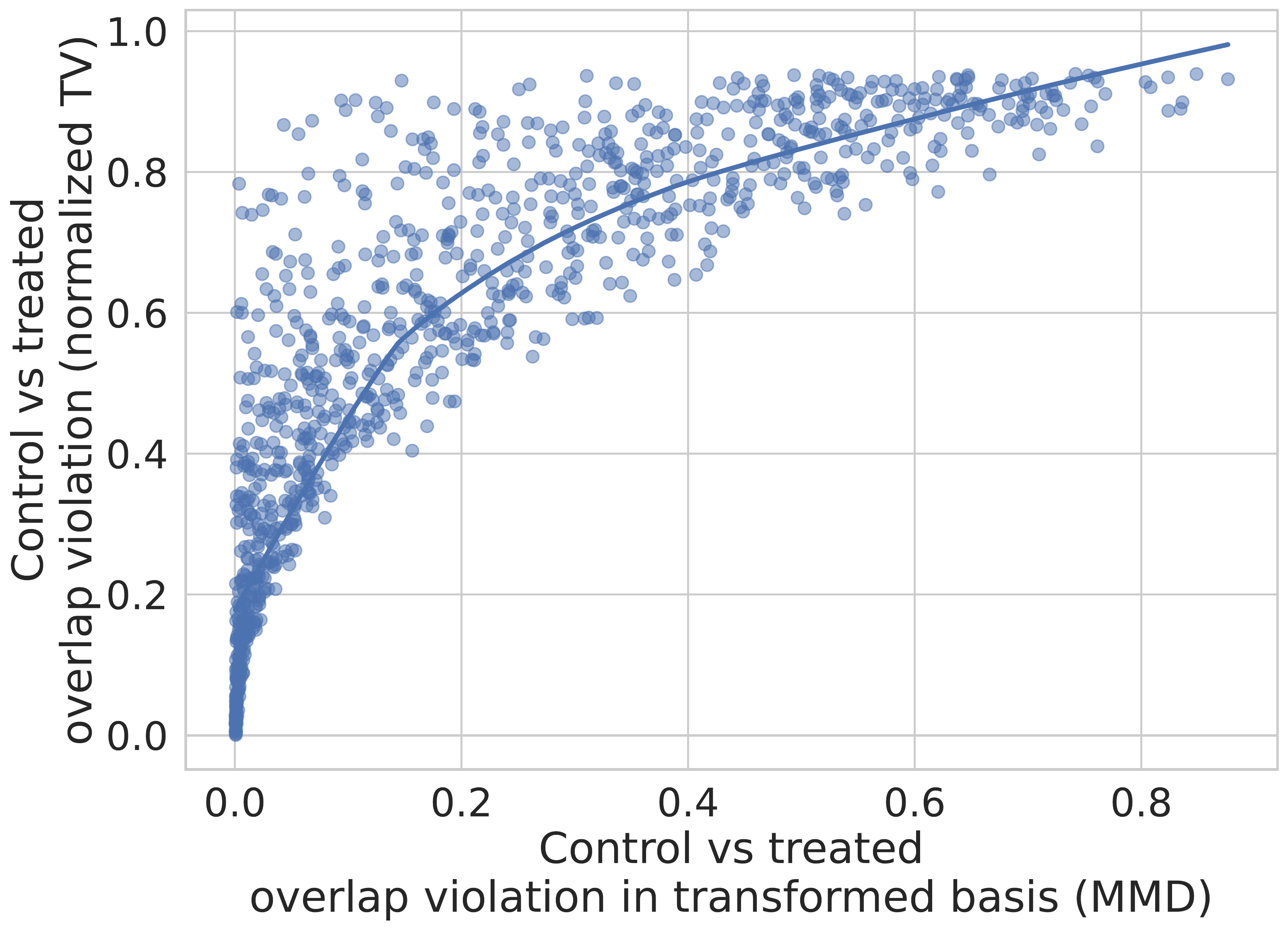}
\end{figure}

\begin{figure}[ht]
    \centering
    \begin{subfigure}[b]{0.47\textwidth}
        \centering
        \caption{\textbf{Caussim}}
        \includegraphics[width=\textwidth]{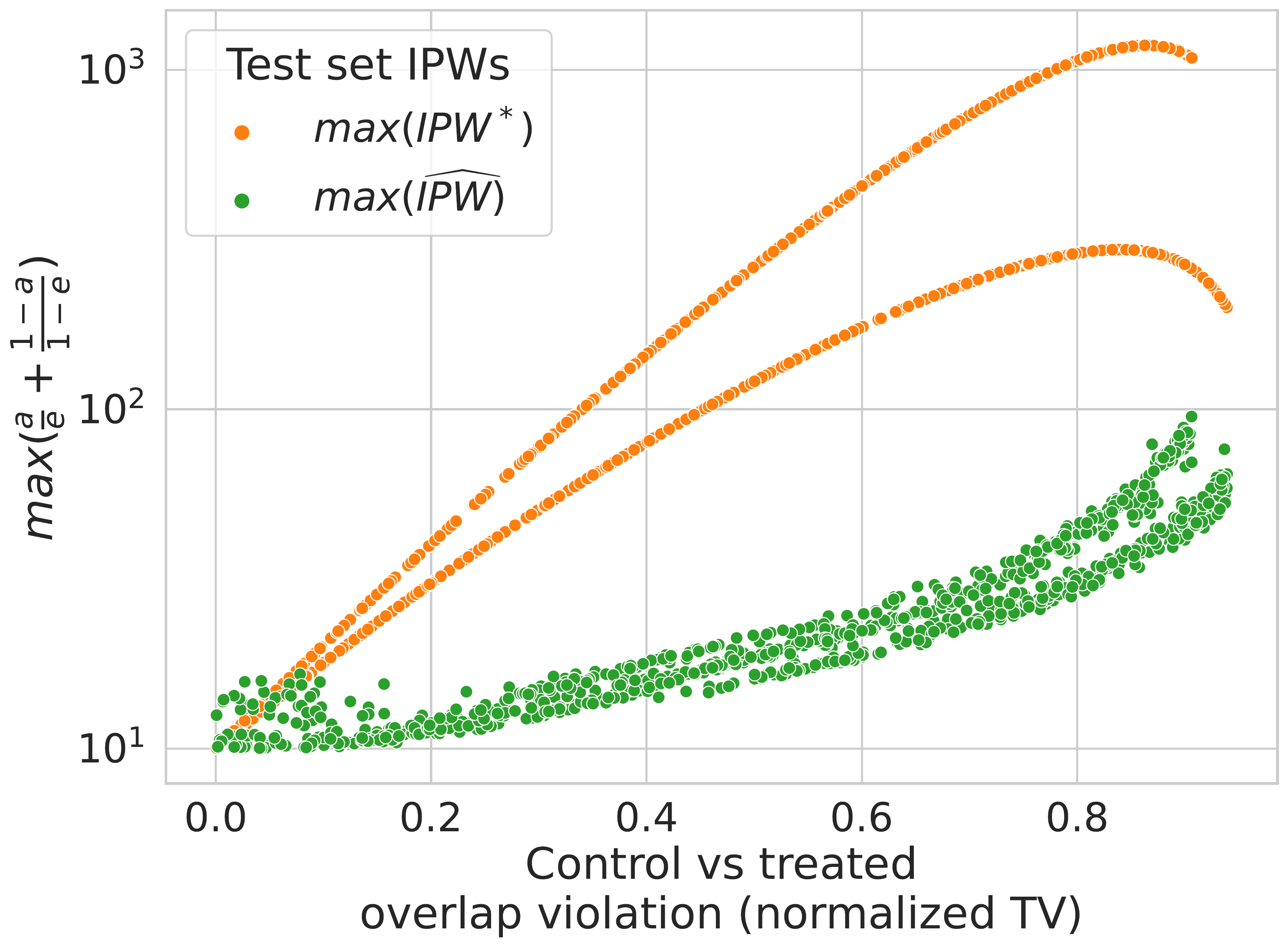}
        \label{apd:caussim:max_ipw_vs_ntv}
    \end{subfigure}
    \hfill
    \begin{subfigure}[b]{0.47\textwidth}
        \centering
        \caption{\textbf{ACIC 2016}}
        \includegraphics[width=\textwidth]{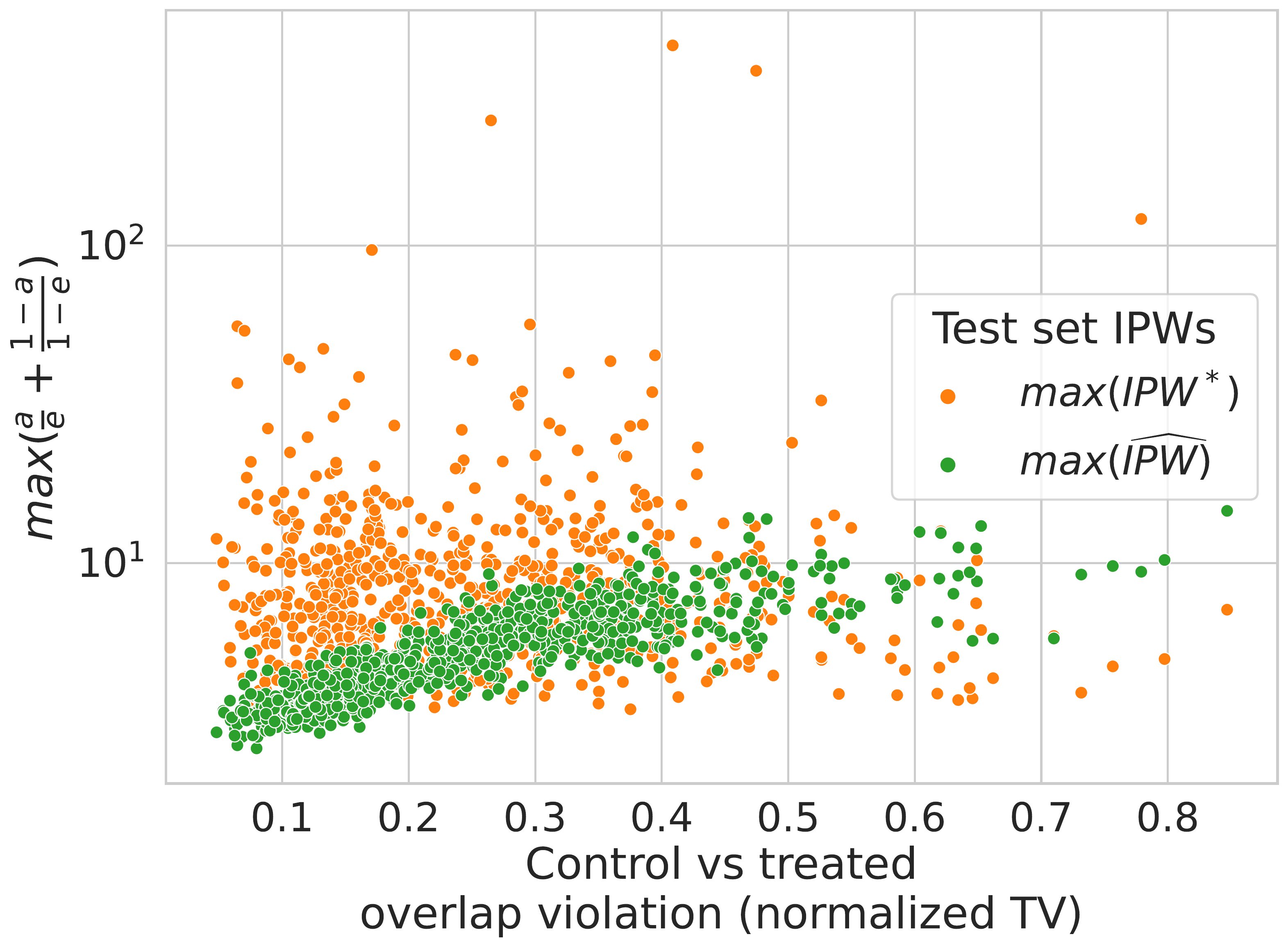}
        \label{apd:acic_2016:ntv_vs_max_ipw}
    \end{subfigure}
    \begin{subfigure}[b]{0.47\textwidth}
        \centering
        \caption{\textbf{ACIC 2018}}
        \includegraphics[width=\textwidth]{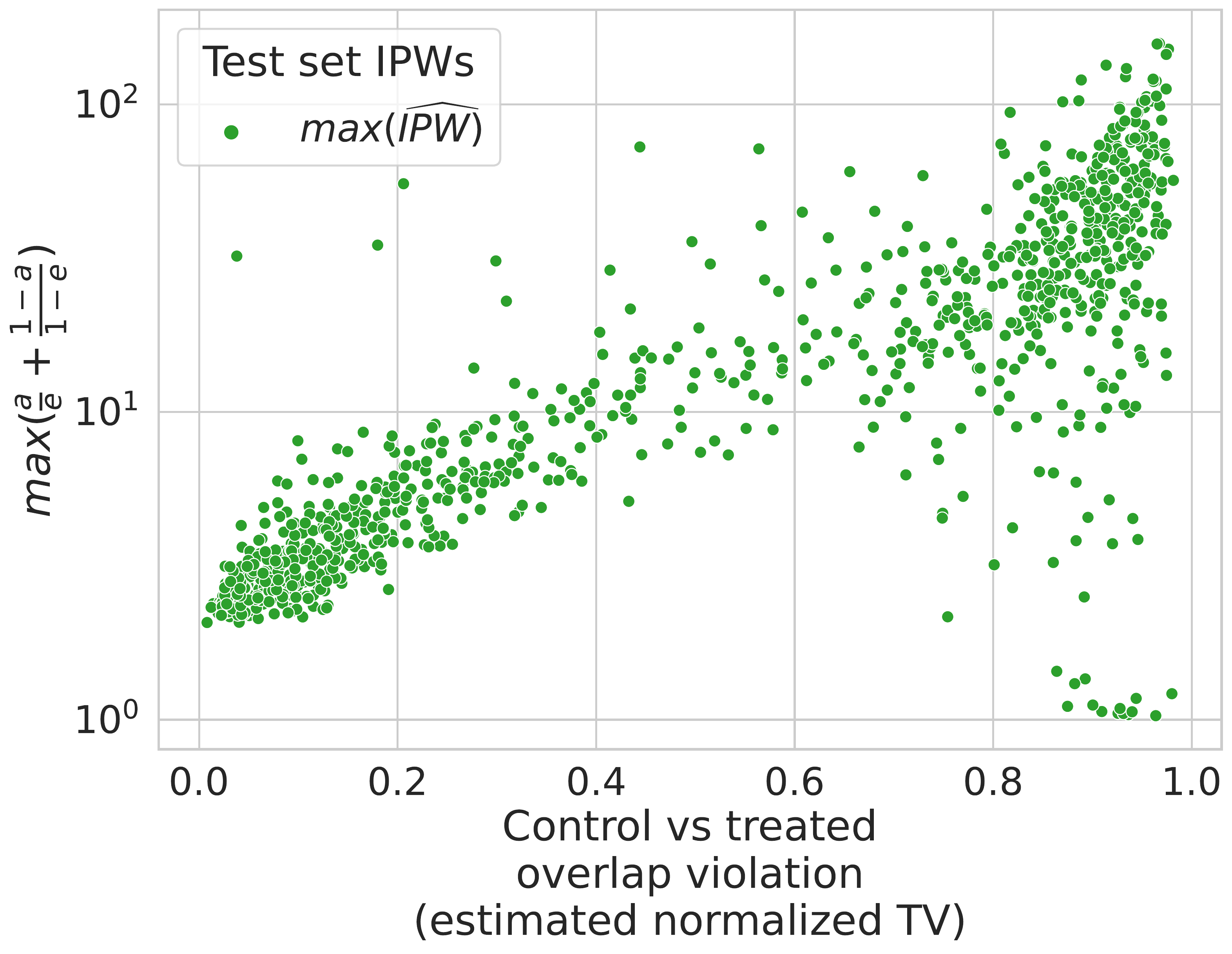}
        \label{apd:acic_2018:ntv_vs_max_ipw}
    \end{subfigure}
    \hfill
    \begin{subfigure}[b]{0.49\textwidth}
        \centering
        \caption{\textbf{TWINS}}
        \includegraphics[width=\textwidth]{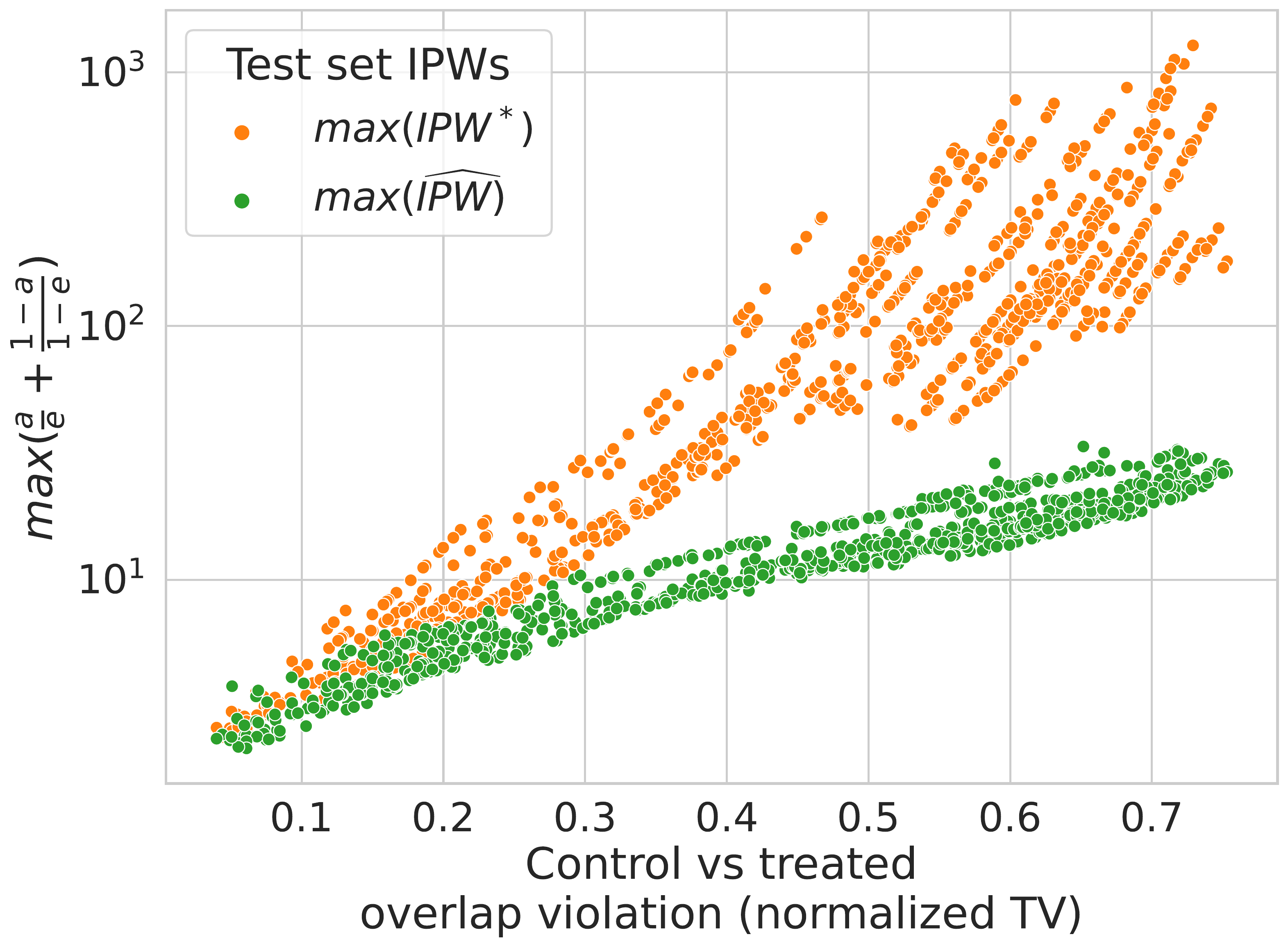}
        \label{apd:twins:ntv_vs_max_ipw}
    \end{subfigure}
    \caption{Maximal value of Inverse Propensity Weights increases exponentially with the overlap as measure by Normalized Total Variation.}
    \label{apd:ntv_vs_max_ipw}
\end{figure}

We show empirically that NTV is an appropriate measure of overlap by :
\begin{itemize}
    \item Comparing the NTV distance with the MMD for Caussim which is gaussian
          distributed in Figure \ref{apd:overlap:caussim:mmd_vs_ntv},
    \item Verifying that setups with penalized overlap from ACIC 2016 have a
          higher total variation distance than unpenalized setups in Figure
          \ref{apd:overlap:penalized_overlap}.
    \item Verifying that the Inverse Propensity Weights extrema (the inverse of
          the $\nu$ overlap constant appearing in the overlap Assumption
          \ref{assumption:overlap}) positevely correlates with NTV for Caussim,
          ACIC 2016 and Twins in Figure \ref{apd:ntv_vs_max_ipw}. Even if the same
          value of the maximum IPW could lead to different values of NTV, we
          expect both measures to be correlated : the higher the extrem propensity
          weights, the higher the NTV.
\end{itemize}

\paragraph{Estimating NTV in practice}

Finally, we verify that approximating the NTV distance with a learned plug-in
estimates of e(x) is reasonnable. We used either a logistic regression or a
gradient boosting classifier to learn the propensity models for the three
datasets where we have access to the ground truth propensity scores: Caussim,
Twins and ACIC 2016. We respectively sampled 1000, 1000 and 770 instances of
these datasets with different seeds and overlap settings. We first run a
hyperparameter search with cross-validation on the train set, then select the
best estimator. We refit on the train set this estimator with or without
calibration by cross validation and finally estimate the normalized TV with the
obtained model. This training procedure reflects the one described in Algorithm
\ref{problem:estimation_procedure:algo} where nuisance models are fitted only on
the train set.

The hyper parameters are : learning rate $ \in [1e-3, 1e-2, 1e-1, 1]$, minimum
samples leaf $\in [2, 10, 50, 100, 200]$ for boosting and L2 regularization $\in
    [1e-3, 1e-2, 1e-1, 1]$ for logistic regression.

Results in Figure \ref{apd:overlap:ntv_approximation} comparing bias to the true
normalized Total Variation of each dataset instances versus growing true NTV
indicate that calibration of the propensity model is crucial to recover a good
approximation of the NTV.

\section{Experiments}

\subsection{Details on the data generation process}
\label{apd:experiments:generation}

We use Gaussian-distributed covariates and random basis expansion based on
Radial Basis Function kernels. A random basis of RBF kernel enables modeling
non-linear and complex relationships between covariates in a similar way to the
well known spline expansion. The estimators of the response function are learned
with a linear model on another random basis (which can be seen as a stochastic
approximation of the full data kernel \cite{rahimi_random_2008}). We
carefully control the amount of overlap between treated and control populations,
a crucial assumption for causal inference.

\begin{itemize}
    \item The raw features for both populations are drawn from a mixture of
          Gaussians:
          $\mathbb P(X) = p_A \mathbb P(X|A=1) + (1- p_A) \mathbb P(X|A=0)$
          where $\mathbb P(x|A=a)$ is a rotated Gaussian:
          \begin{equation}
              \mathbb P(x|A=a) = W \cdot \mathcal N \Big( \begin{bmatrix} (1-2a) \theta \\ 0\end{bmatrix} ; \begin{bmatrix} \sigma_0 & 0 \\ 0 & \sigma_1\end{bmatrix} \Big)
          \end{equation}
          with $\theta$ a parameter controlling overlap (bigger yields poorer
          overlap), $W$ a random rotation matrix and $\sigma_0^2=2;\sigma_1^2=5$.

          This generation process allows to analytically compute the oracle
          propensity scores $e(x)$, to simply control for overlap with the
          parameter $\theta$, the distance between the two Gaussian main axes and
          to  visualize response surfaces.

    \item A basis expansion of the raw features increases the problem dimension.
          Using Radial Basis Function (RBF) Nystroem transformation \footnote{We use the
              \href{https://scikit-learn.org/stable/modules/generated/sklearn.kernel_approximation.Nystroem.html}{Sklearn
                  implementation, \cite{pedregosa_scikitlearn_2011}}}, we expand the raw
          features into a transformed space. The basis expansion samples randomly a
          small number of representers in the raw data. Then,  it computes an
          approximation of the full N-dimensional kernel with these basis components,
          yielding the transformed features $z(x)$.

          We generate the basis following the original data distribution, $\left [
                  b_1 .. b_D \right ] \sim \mathbb P(x)$, with D=2 in our simulations. Then, we
          compute an approximation of the full kernel of the data generation
          process $RBF(x, \cdot) \;  with \; x \sim \mathbb P(x)$ with these
          representers: $z(x) = [RBF_{\gamma}(x, b_d)]_{d=1..D}
              \cdot Z^T \in \mathbb{R}^D$
          with $RBF_{\gamma}$ being the Gaussian kernel $K(x, y) = exp(-\gamma
              ||x-y||^2)$ and Z the normalization constant of the kernel basis,
          computed as the root inverse of the basis kernel $Z=[K(b_i, b_j)]_{i, j
              \in {1..D}}^{-1/2}$

    \item Functions $\mu_0$, $\tau$ are distinct linear functions of the
          transformed features:
          \begin{equation*}
              \mu_0(x) = \begin{bmatrix} z(x); 1 \end{bmatrix} \cdot \beta_{\mu}^T
          \end{equation*}
          \begin{equation*}
              \tau(x) = \begin{bmatrix} z(x); 1 \end{bmatrix} \cdot \beta_{\tau}^T
          \end{equation*}
    \item Adding a Gaussian noise, $\varepsilon \sim \mathcal N(0, \sigma(x;a))$,
          we construct the potential outcomes:
          $y(a) = \mu_0(x) + a\,\tau(x) + \varepsilon(x, a)$
\end{itemize}
We generated 1000 instances of this dataset with uniformly random overlap
parameters $\theta \in \left[ 0, 2.5 \right]$.

\subsection{Model selection procedures}

\paragraph{Nuisances estimation}\label{apd:experiments:nuisances_hp}

The nuisances are estimated with a stacked regressor inspired by the Super
Learner framework, \cite{laan_super_2007}). The hyper-parameters are
optimized with a random search with following search grid detailed in Table
\ref{apd:experiments:nuisances_hp_grid}. All implementations come from
\href{https://scikit-learn.org/stable/}{scikit-learn}
\cite{pedregosa_scikitlearn_2011}.

\begin{table}[h!]
    \begin{tabular}{llll}
        \toprule
        Model                & Estimator
                             & Hyper-parameters grid                                         \\
        \midrule
        Outcome, m           & StackedRegressor
                             & ridge regularization: [0.0001, 0.001, 0.01, 0.1, 1, 10, 100]  \\
        \multirow[c]{3}{*}{} & (HistGradientBoostingRegressor, ridge)
                             & HistGradientBoostingRegressor  learning rate: [0.01, 0.1, 1]  \\
                             &
                             & HistGradientBoostingRegressor  max leaf nodes: [10,
        20, 30, 50]                                                                          \\
        \midrule
        Treatment, e         & StackedClassifier
                             & LogisticRegression  C: [0.0001, 0.001, 0.01, 0.1, 1, 10, 100] \\
        \multirow[c]{3}{*}{} & (HistgradientBoostingClassifier, LogisticRegression)
                             & HistGradientBoostingClassifier  learning rate: [0.01, 0.1, 1] \\
                             &
                             & HistGradientBoostingClassifier  max leaf nodes: [10,
        20, 30, 50]                                                                          \\
        \bottomrule
    \end{tabular}
    \caption{Hyper-parameters grid used for nuisance models}
    \label{apd:experiments:nuisances_hp_grid}
\end{table}

\subsection{Additional Results}\label{apd:experiments:additional_results}

\paragraph{Definition of the Kendall's tau, $\kappa$}

The Kendall's tau  is a widely used statistics to measure the rank correlation
between two set of observations. It measures the number of concordant pairs
minus the discordant pairs normalized by the total number of pairs. It takes values in the
$[-1, 1]$ range.
\begin{equation}\label{eq:kendall_tau}
    \kappa=\frac{\text { (number of concordant pairs })-\text { (number of discordant pairs) }}{\text { (number of pairs) }}
\end{equation}

\paragraph{Values of relative $\kappa(\ell,\tau\mathrm{{-risk}})$ compared to
    the mean over all metrics Kendall's as shown in the boxplots of Figure \ref{fig:relative_kendalls_all_datasets}}

\begin{table}
    \centering
    \resizebox{0.7\textwidth}{!}{
        \begin{tabular}{llrrrr}
    \toprule
                                                                 &           &
    \multicolumn{2}{r}{Strong
    Overlap}                                                     &
    \multicolumn{2}{r}{Weak
        Overlap}
    \\ \midrule
                                                                 &           & Median & IQR   & Median & IQR \\
    Metric                                                       & Dataset   &        &       &        &     \\
    \midrule
    \multirow[c]{4}{*}{$\widehat{\mu\mathrm{-risk}}$}            & Twins
    (N=11 984)                                                   & -0.32     & 0.12   & -0.19 & 0.12         \\
    \cline{2-6}
                                                                 & ACIC 2016
    (N=4 802)                                                    & -0.03     & 0.13   & 0.11  & 0.19         \\
    \cline{2-6}
                                                                 & Caussim
    (N=5 000)                                                    & -0.40     & 0.55   & -0.16 & 0.31         \\
    \cline{2-6}
                                                                 & ACIC 2018
    (N=5 000)                                                    & 0.00      & 0.30   & 0.01  & 0.40         \\
    \cline{1-6} \cline{2-6}
    \multirow[c]{4}{*}{$\widehat{\mu\mathrm{-risk}}_{IPW}$}      & Twins
    (N= 11 984)                                                  & -0.31     & 0.13   & -0.17 & 0.12         \\
    \cline{2-6}
                                                                 & ACIC 2016
    (N=4 802)                                                    & -0.02     & 0.13   & 0.11  & 0.19         \\
    \cline{2-6}
                                                                 & Caussim
    (N=5 000)                                                    & -0.34     & 0.50   & 0.09  & 0.31         \\
    \cline{2-6}
                                                                 & ACIC 2018
    (N=5 000)                                                    & 0.00      & 0.30   & -0.01 & 0.43         \\
    \cline{1-6} \cline{2-6}
    \multirow[c]{3}{*}{$\widehat{\mu\mathrm{-risk}}^{*}_{IPW}$}  & Twins
    (N= 11 984)                                                  & -0.32     & 0.13   & -0.17 & 0.13         \\
    \cline{2-6}
                                                                 & ACIC 2016
    (N=4 802)                                                    & -0.02     & 0.13   & 0.11  & 0.21         \\
    \cline{2-6}
                                                                 & Caussim
    (N=5 000)                                                    & -0.33     & 0.54   & 0.26  & 0.27         \\
    \cline{1-6} \cline{2-6}
    \multirow[c]{4}{*}{$\widehat{\tau\mathrm{-risk}}_{IPW}$}     & Twins
    (N= 11 984)                                                  & 0.13      & 0.12   & 0.27  & 0.12         \\
    \cline{2-6}
                                                                 & ACIC 2016
    (N=4 802)                                                    & -0.07     & 0.18   & 0.05  & 0.31         \\
    \cline{2-6}
                                                                 & Caussim
    (N=5 000)                                                    & -0.19     & 0.43   & -0.14 & 0.18         \\
    \cline{2-6}
                                                                 & ACIC 2018
    (N=5 000)                                                    & -0.16     & 0.40   & -0.11 & 0.66         \\
    \cline{1-6} \cline{2-6}
    \multirow[c]{3}{*}{$\widehat{\tau\mathrm{-risk}}_{IPW}^{*}$} & Twins
    (N= 11 984)                                                  & 0.12      & 0.14   & 0.20  & 0.16         \\
    \cline{2-6}
                                                                 & ACIC 2016
    (N=4 802)                                                    & -0.03     & 0.16   & -0.09 & 0.43         \\
    \cline{2-6}
                                                                 & Caussim
    (N=5 000)                                                    & -0.15     & 0.46   & -0.17 & 0.19         \\
    \cline{1-6} \cline{2-6}
    \multirow[c]{4}{*}{$\widehat{\mathrm{U-risk}}$}              & Twins
    (N= 11 984)                                                  & 0.13      & 0.12   & 0.02  & 0.25         \\
    \cline{2-6}
                                                                 & ACIC 2016
    (N=4 802)                                                    & 0.04      & 0.11   & 0.11  & 0.26         \\
    \cline{2-6}
                                                                 & Caussim
    (N=5 000)                                                    & 0.04      & 0.43   & -0.04 & 0.17         \\
    \cline{2-6}
                                                                 & ACIC 2018
    (N=5 000)                                                    & 0.12      & 0.26   & -0.02 & 0.50         \\
    \cline{1-6} \cline{2-6}
    \multirow[c]{3}{*}{$\widehat{\mathrm{U-risk}}^{*}$}          & Twins
    (N= 11 984)                                                  & 0.25      & 0.08   & -0.41 & 0.45         \\
    \cline{2-6}
                                                                 & ACIC 2016
    (N=4 802)                                                    & 0.08      & 0.13   & -0.59 & 0.57         \\
    \cline{2-6}
                                                                 & Caussim
    (N=5 000)                                                    & 0.46      & 0.12   & 0.02  & 0.44         \\
    \cline{1-6} \cline{2-6}
    \multirow[c]{4}{*}{$\widehat{\mathrm{R-risk}}$}              & Twins
    (N= 11 984)                                                  & 0.15      & 0.10   & 0.25  & 0.18         \\
    \cline{2-6}
                                                                 & ACIC 2016
    (N=4 802)                                                    & 0.07      & 0.12   & 0.22  & 0.15         \\
    \cline{2-6}
                                                                 & Caussim
    (N=5 000)                                                    & 0.34      & 0.26   & 0.13  & 0.21         \\
    \cline{2-6}
                                                                 & ACIC 2018
    (N=5 000)                                                    & 0.13      & 0.27   & 0.21  & 0.47         \\
    \cline{1-6} \cline{2-6}
    \multirow[c]{3}{*}{$\widehat{\mathrm{R-risk}}^{*}$}          & Twins
    (N= 11 984)                                                  & 0.25      & 0.10   & 0.32  & 0.15         \\
    \cline{2-6}
                                                                 & ACIC 2016
    (N=4 802)                                                    & 0.12      & 0.12   & 0.25  & 0.15         \\
    \cline{2-6}
                                                                 & Caussim
    (N=5 000)                                                    & 0.47      & 0.11   & 0.16  & 0.14         \\
    \cline{1-6} \cline{2-6}
    \bottomrule
\end{tabular}

    }
    \caption{Values of relative $\kappa(\ell,\tau\mathrm{{-risk}})$ compared to
        the mean over all metrics Kendall's as shown in the boxplots of Figure
        \ref{fig:relative_kendalls_all_datasets}}\label{apd:table:relative_kendalls_all_datasets}
\end{table}

\begin{figure}
    \centering
    \begin{subfigure}[b]{0.44\textwidth}
        \centering
        \caption{\textbf{Caussim}}
        \includegraphics[width=\textwidth]{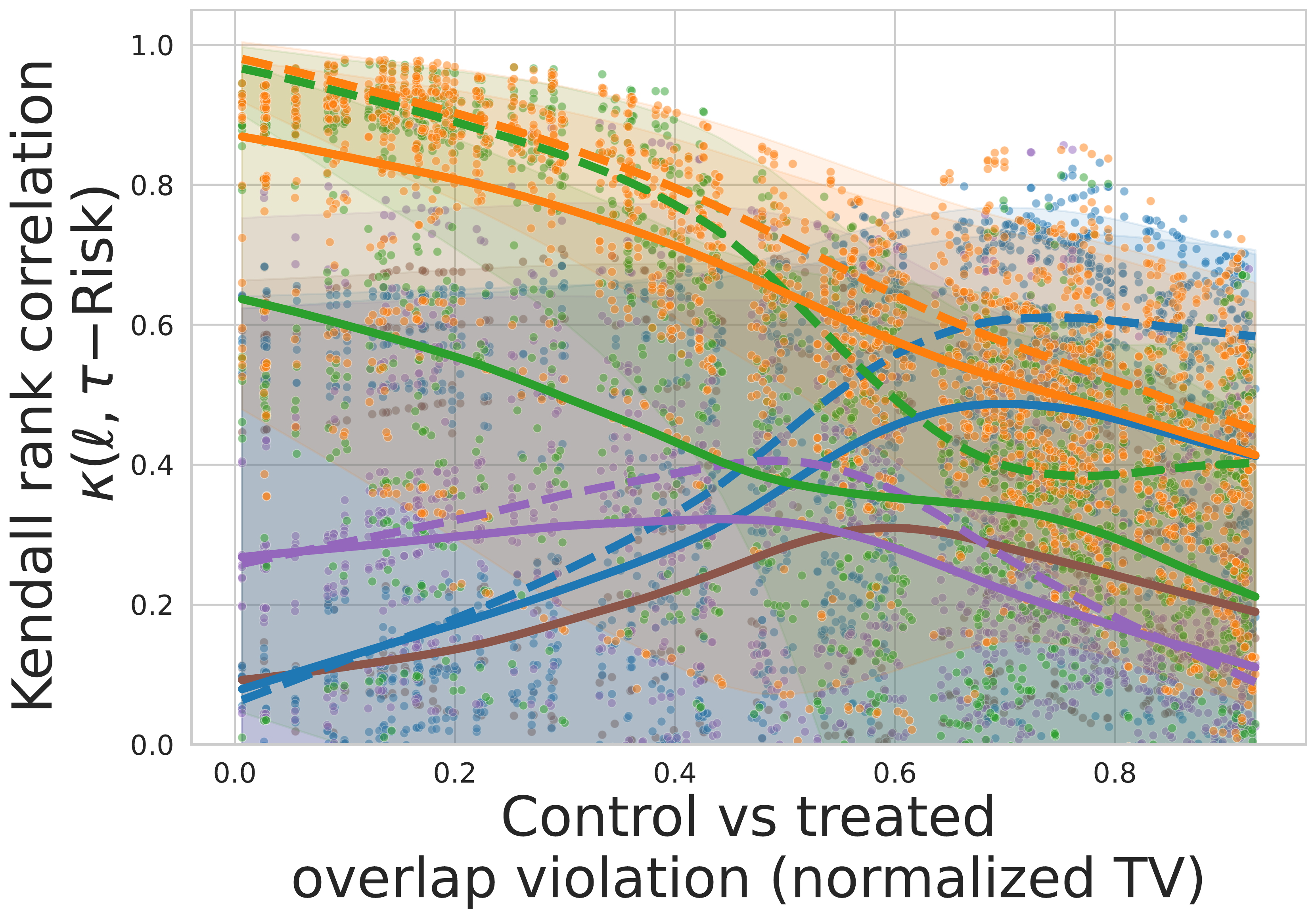}
        \label{fig:ranking_agreement_w_tau_risk_caussim}
    \end{subfigure}
    \hfill
    \begin{subfigure}[b]{0.44\textwidth}
        \centering
        \caption{\textbf{ACIC 2016}}
        \includegraphics[width=\textwidth]{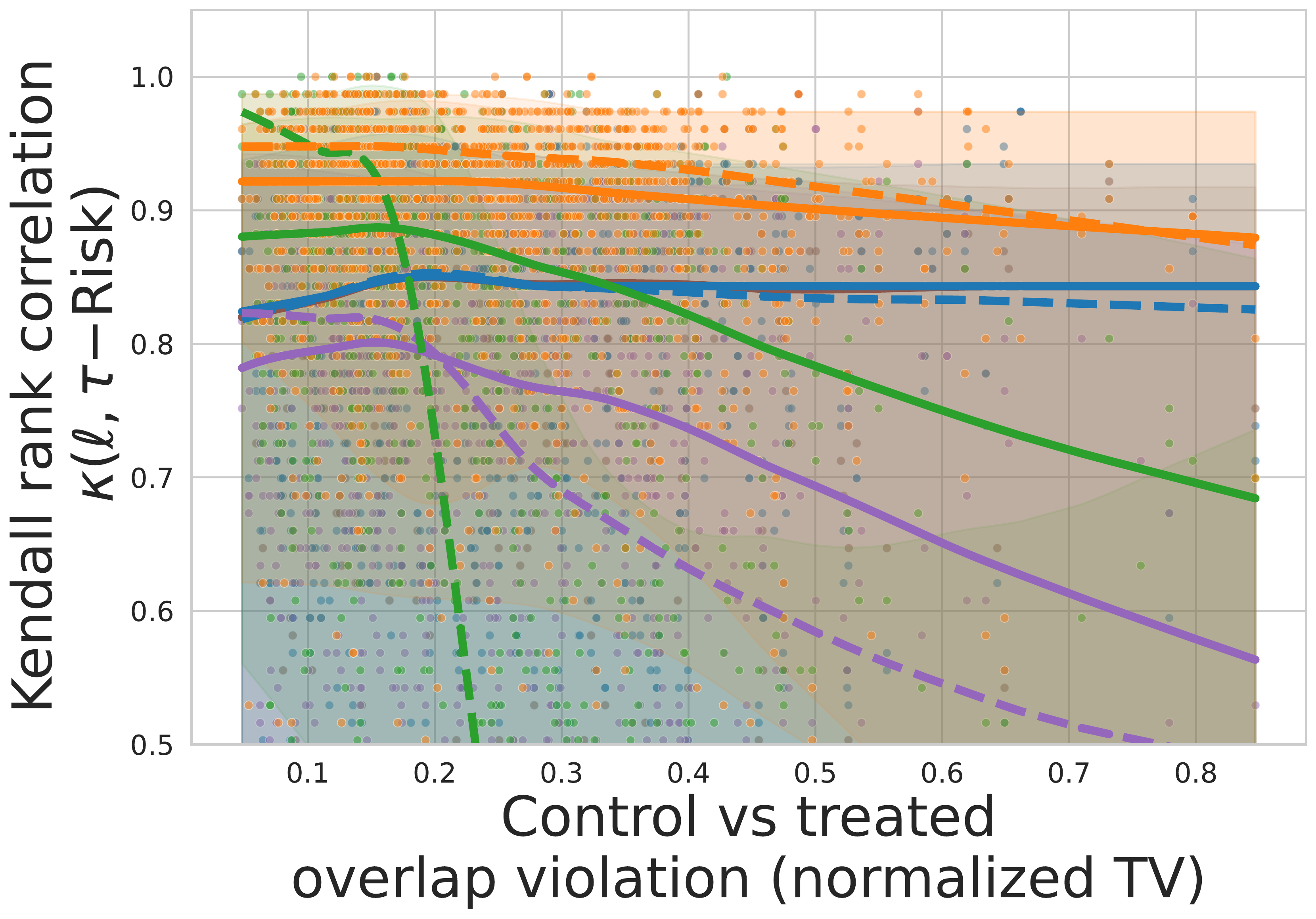}
        \label{fig:ranking_agreement_tau_risk_acic_2016}
    \end{subfigure}
    \hfill
    \begin{subfigure}[b]{0.10\textwidth}
        \centering
        \includegraphics[width=1.5\textwidth]{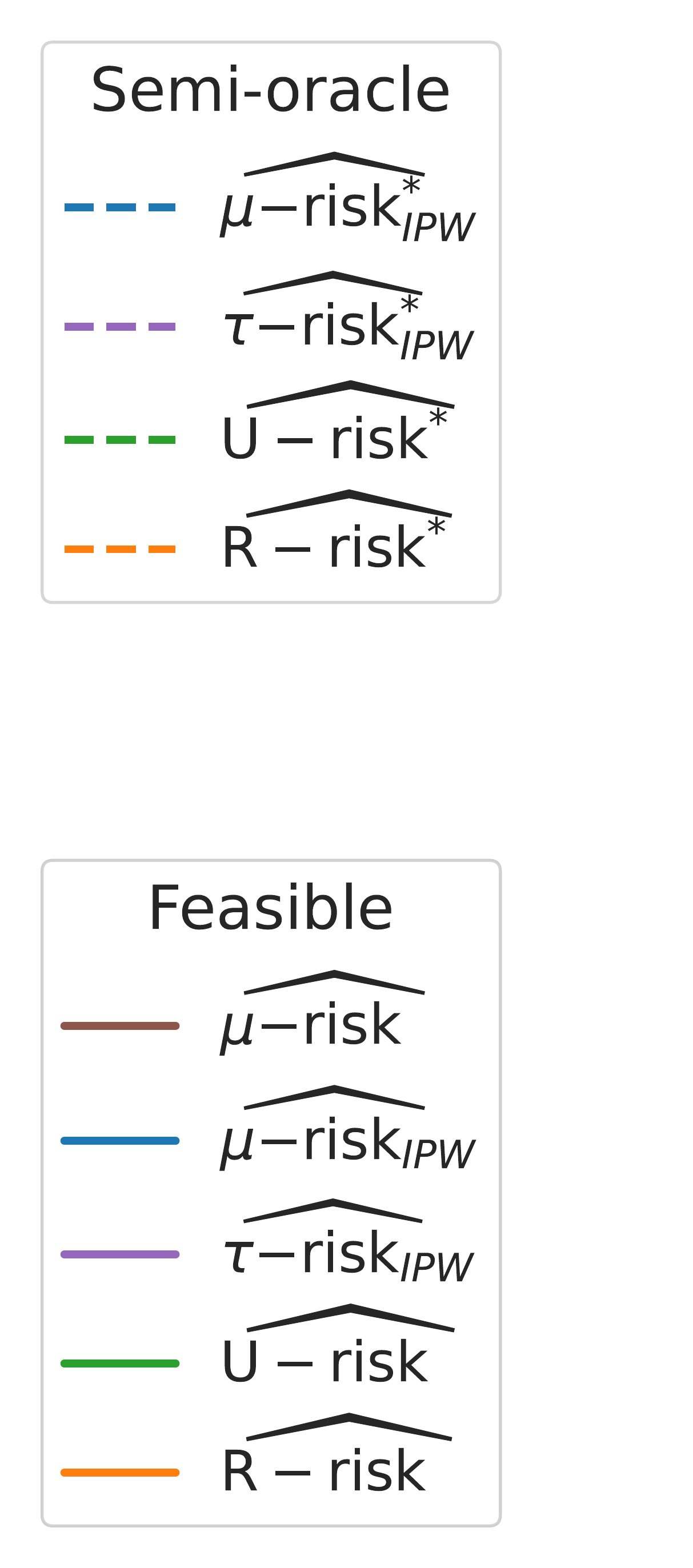}
    \end{subfigure}
    \begin{subfigure}[b]{0.44\textwidth}
        \centering
        \caption{\textbf{ACIC 2018}}
        \includegraphics[width=\textwidth]{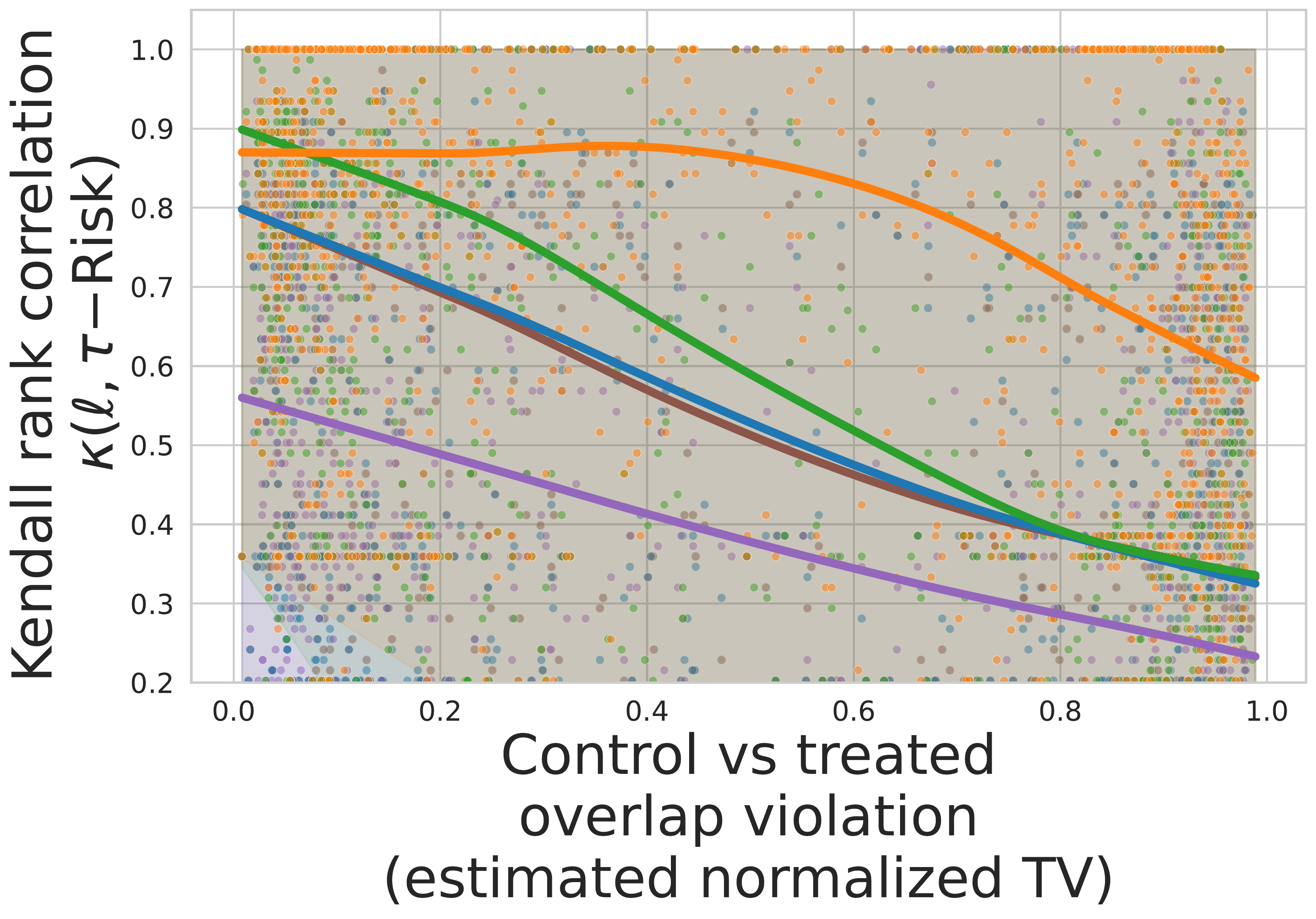}
        \label{fig:ranking_agreement_w_tau_risk_acic_2018}
    \end{subfigure}
    \hfill
    \begin{subfigure}[b]{0.44\textwidth}
        \centering
        \caption{\textbf{TWINS}}
        \includegraphics[width=\textwidth]{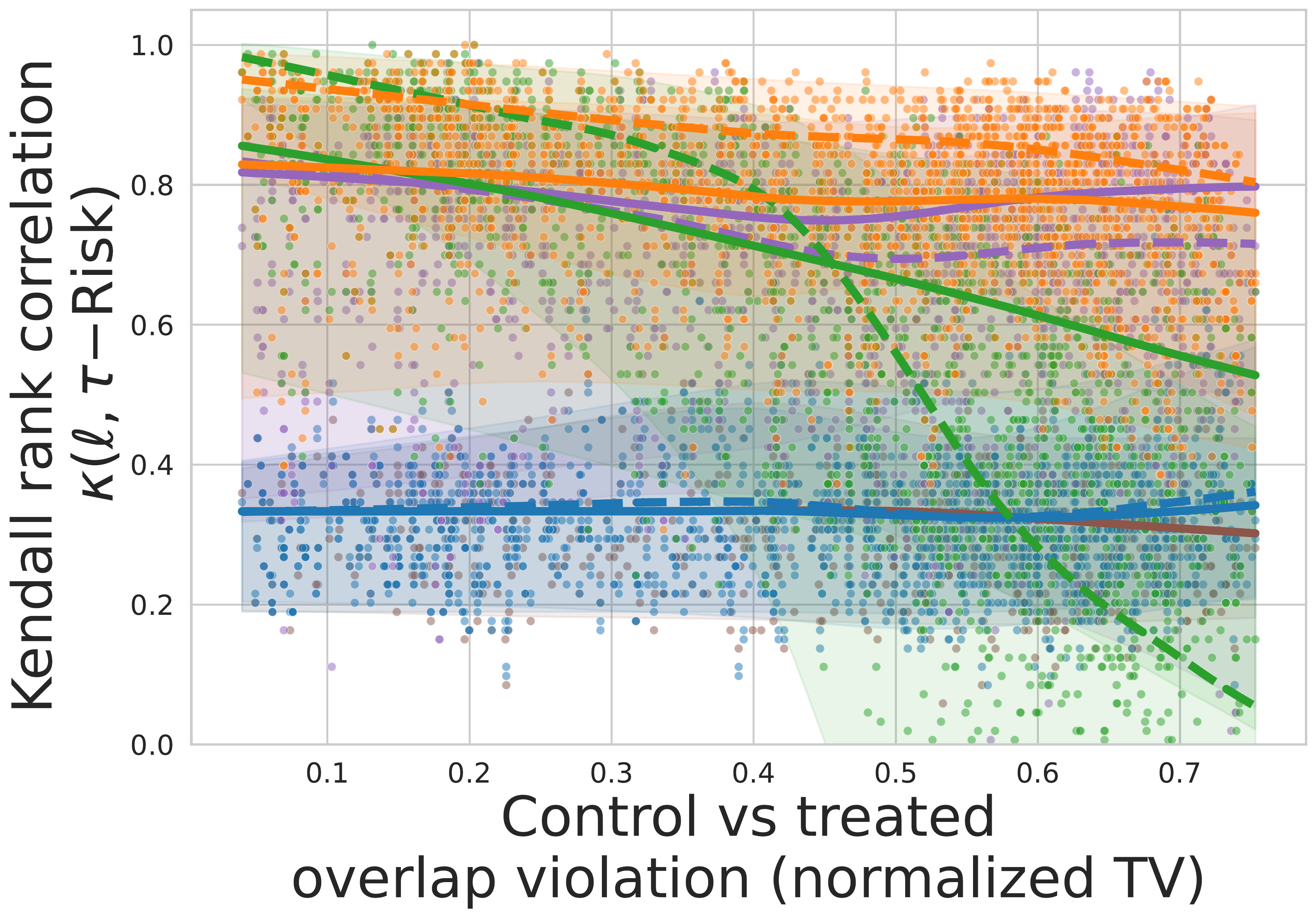}
        \label{fig:ranking_agreement_tau_risk_twins}
    \end{subfigure}
    \hfill
    \begin{subfigure}[b]{0.10\textwidth}
        ~
    \end{subfigure}
    % \begin{subfigure}[b]{\textwidth}
    %   \centering
    %   \includegraphics[width=0.5\textwidth]{images/legend_metrics_horizontal.pdf}
    % \end{subfigure}
    \caption{Agreement with $\tau\text{-risk}$ ranking of methods function
        of overlap violation. The lines represent medians, estimated with a
        lowess. The transparent
        bands denote the 5\% and 95\% confidence intervals.}\label{apd:fig:all_datasets_tau_risk_ranking_agreement}
\end{figure}

\paragraph{Figure \ref{apd:fig:all_datasets_tau_risk_ranking_agreement} -
    Results measured in absolute Kendall's}

\begin{figure}
    \centering
    \begin{subfigure}[b]{0.44\textwidth}
        \centering
        \caption{\textbf{Caussim}}
        \includegraphics[width=\textwidth]{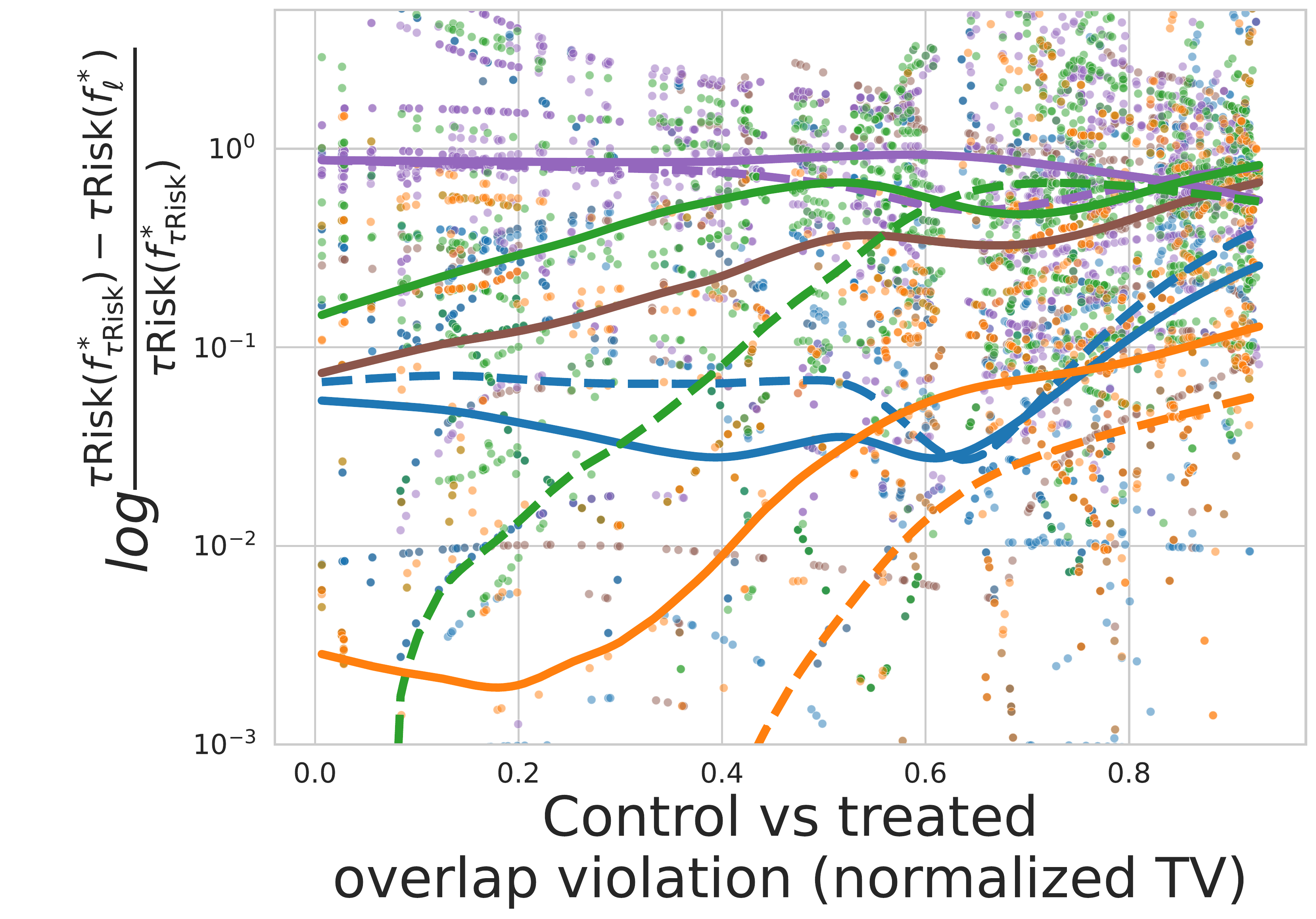}
        \label{fig:normalized_bias_tau_risk_to_best_method_caussim}
    \end{subfigure}
    \hfill
    \begin{subfigure}[b]{0.44\textwidth}
        \centering
        \caption{\textbf{ACIC 2016}}
        \includegraphics[width=\textwidth]{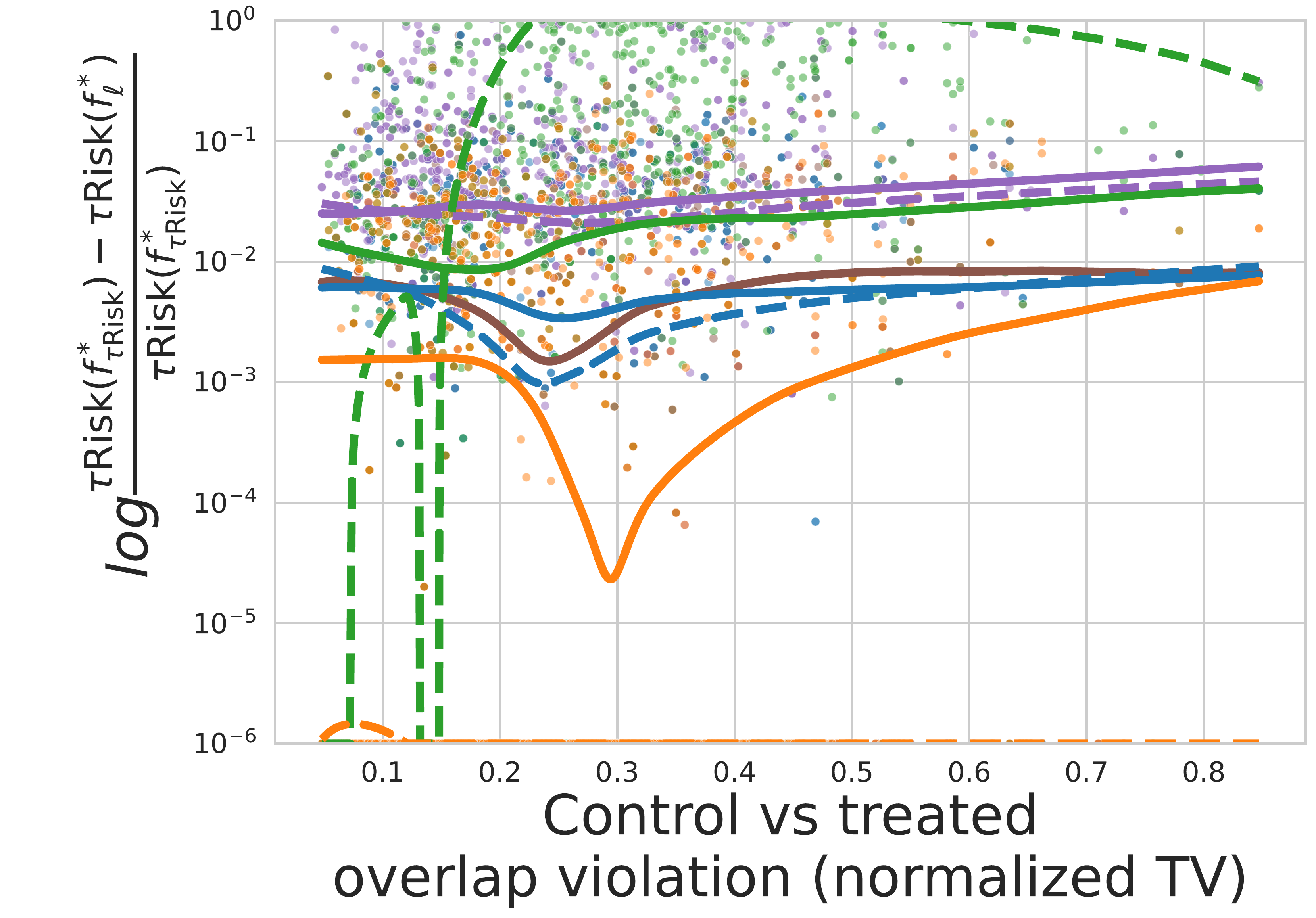}
        \label{fig:normalized_bias_tau_risk_to_best_method_acic_2016}
    \end{subfigure}
    \hfill
    \begin{subfigure}[b]{0.10\textwidth}
        \centering
        \includegraphics[width=1.5\textwidth]{images/legend_metrics.pdf}
    \end{subfigure}
    \begin{subfigure}[b]{0.44\textwidth}
        \centering
        \caption{\textbf{ACIC 2018}}
        \includegraphics[width=\textwidth]{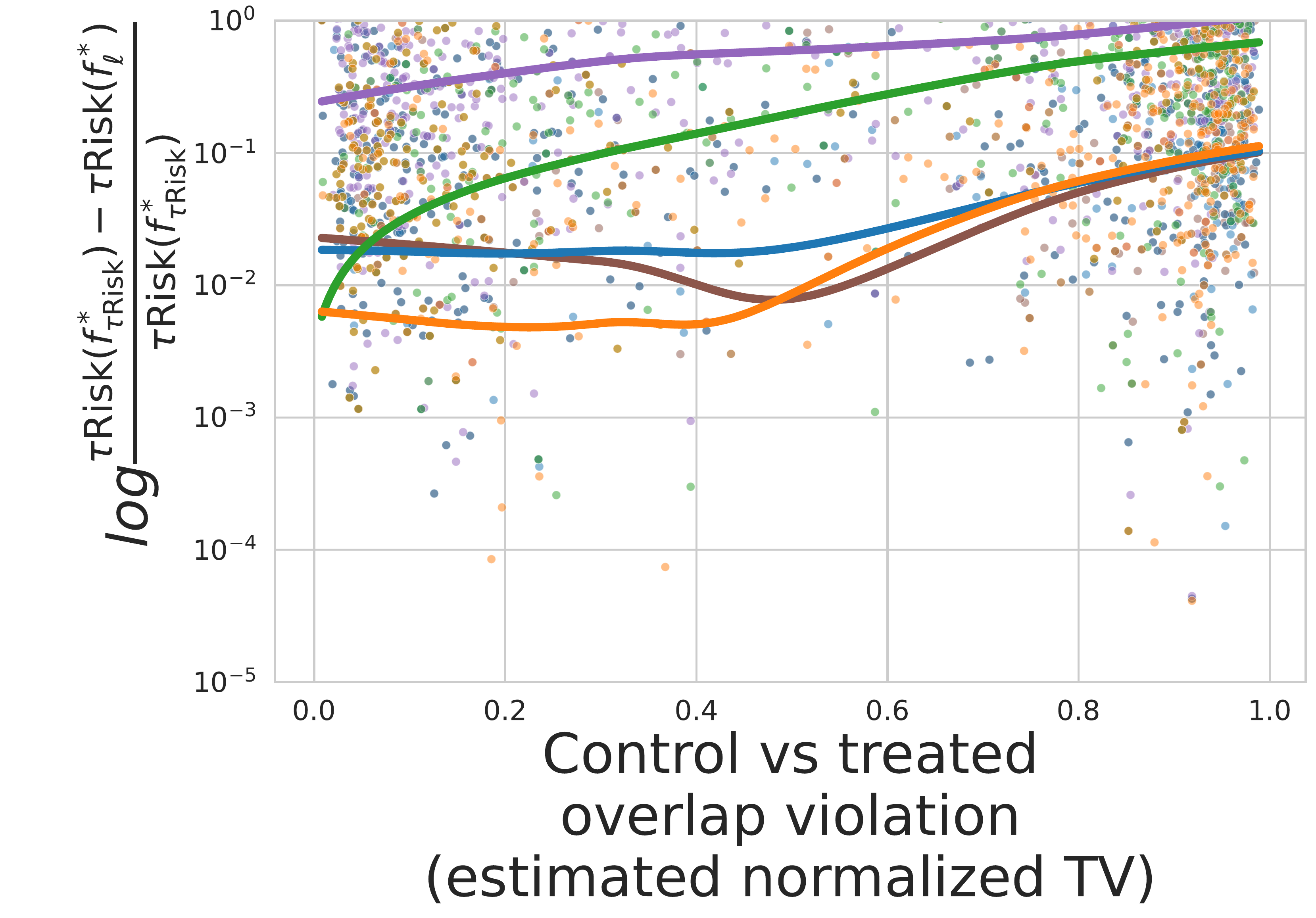}
        \label{fig:normalized_bias_tau_risk_to_best_method_acic_2018}
    \end{subfigure}
    \hfill
    \begin{subfigure}[b]{0.44\textwidth}
        \centering
        \caption{\textbf{TWINS}}
        \includegraphics[width=\textwidth]{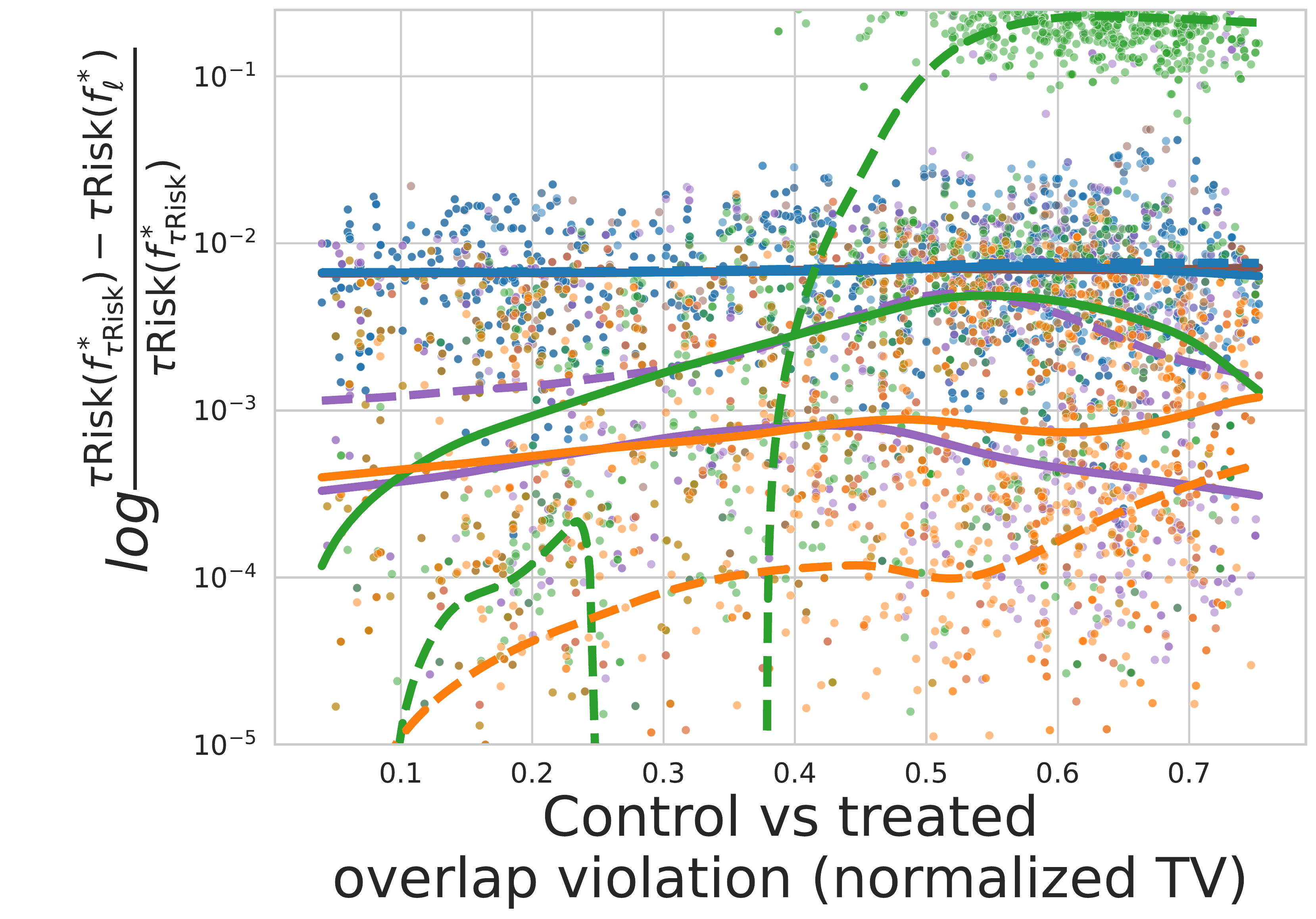}
        \label{fig:normalized_bias_tau_risk_to_best_method_twins}
    \end{subfigure}
    \hfill
    \begin{subfigure}[b]{0.10\textwidth}
        ~
    \end{subfigure}
    \caption{Metric performances by normalized tau-risk distance to the best
        method selected with $\tau\text{-risk}$. All nuisances are learned with the
        same estimator stacking gradient boosting and ridge regression. Doted and
        plain lines corresponds to 60\% lowess quantile estimates. This choice of
        quantile allows to see better the oracle metrics lines for which outliers with a value
        of 0 distord the curves.}
    \label
    {apd:all_datasets_normalized_bias_tau_risk_to_best_method}
\end{figure}

\paragraph{Figure \ref{apd:all_datasets_normalized_bias_tau_risk_to_best_method}
    - Results measured as distance to the oracle tau-risk}

To see practical gain in term of $\tau\text{-risk}$, we plot the results as the
normalized distance between the estimator selected by the oracle
$\tau\text{-risk}$ and the estimator selected by each causal metric.

Then, $\widehat{R\text{-risk}}^*$ is more efficient than all other metrics. The
gain are substantial for every datasets.

\begin{figure}
    \begin{subfigure}[b]{0.9\textwidth}
        %\centering
        \caption{\textbf{Caussim}}
        \includegraphics[width=1.15\textwidth]{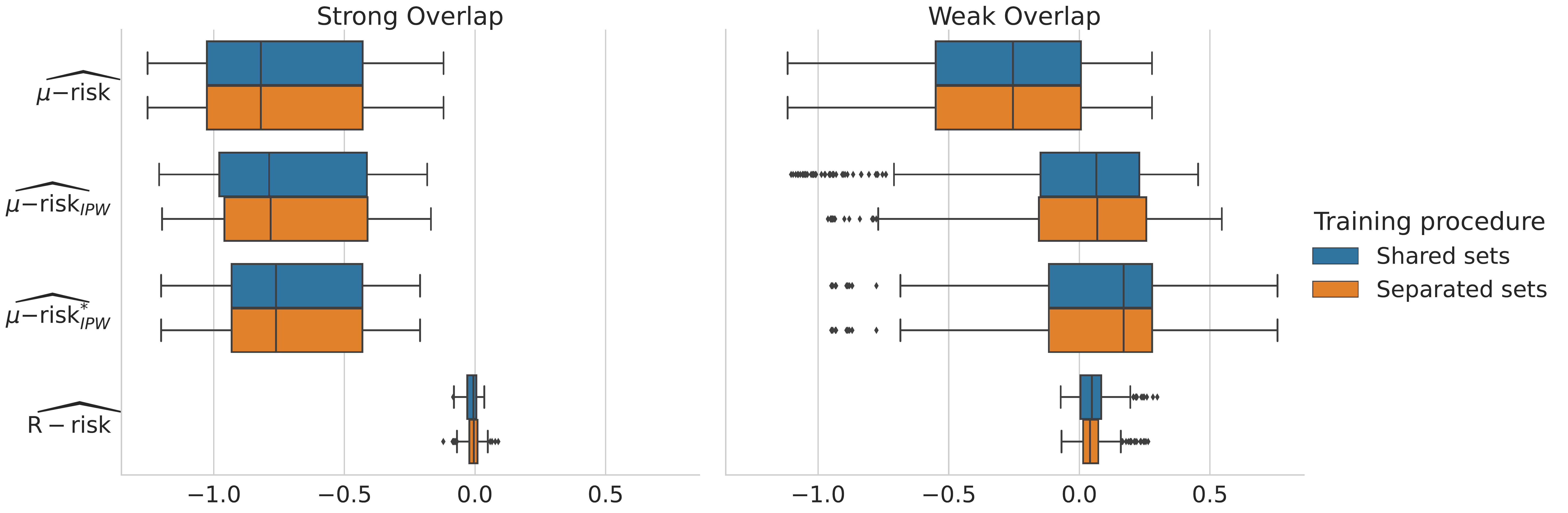}
        \label{fig:experiments:procedures_comparison:caussim}
    \end{subfigure}
    \hfill
    \begin{subfigure}[b]{0.9\textwidth}
        \centering
        \caption{\textbf{ACIC 2016}}
        \includegraphics[width=1\textwidth]{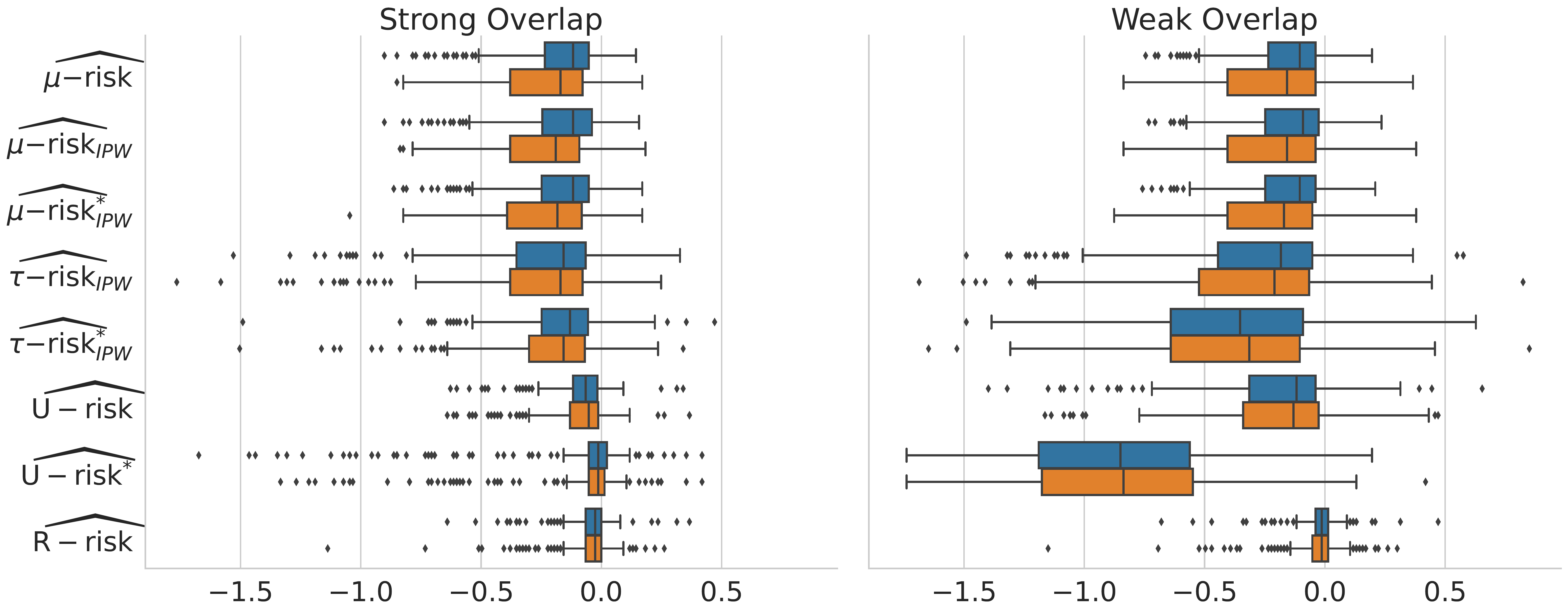}
        \label{fig:experiments:procedures_comparison:acic_2016}
    \end{subfigure}
    \hfill
    \begin{subfigure}[b]{0.9\textwidth}
        \centering
        \caption{\textbf{Twins}}
        \includegraphics[width=1\textwidth]{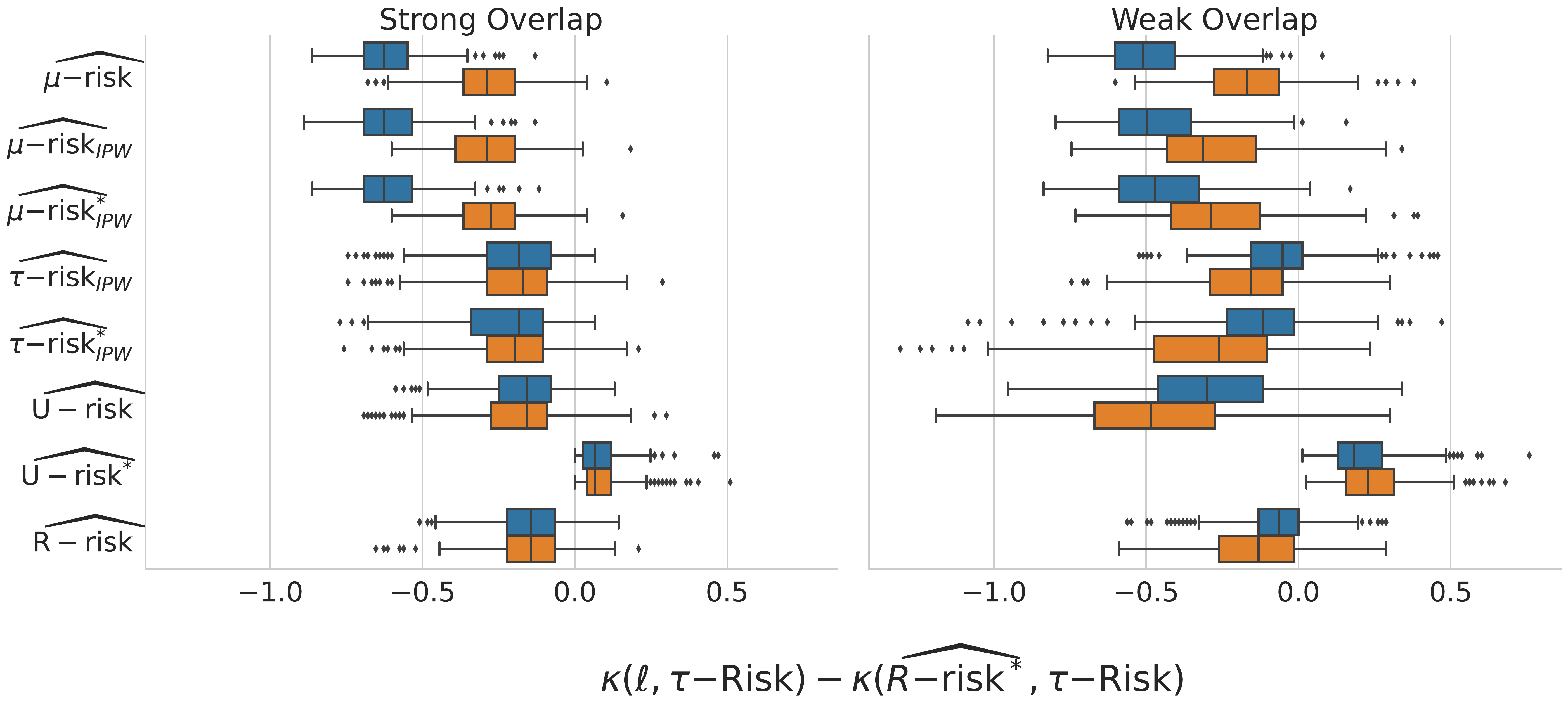}
        \label{fig:experiments:procedures_comparison:twins}
    \end{subfigure}
    \hfill
    \caption{Results are similar between the \textcolor{MidnightBlue}{Shared
            nuisances/candidate set} and
        the \textcolor{RedOrange}{Separated nuisances set} procedure. The
        experience has not been run on the full metrics for Caussim due to computation costs.}\label
    {apd:fig:procedures_comparison_all_metrics}
\end{figure}

\paragraph{Figure \ref{apd:fig:procedures_comparison_all_metrics} - Stacked models for the nuisances is more efficient}
For each metrics the benefit of
using a stacked model of linear and boosting estimators for nuisances compared
to a linear model. The evaluation measure is Kendall's tau relative to the
oracle $R\text{-risk}^{\star}$ to have a stable reference between exepriments.
Thus, we do not include in this analysis the ACIC 2018 dataset since
$R\text{-risk}^{\star}$ is not available due to the lack of the true propensity
score.

\begin{figure}
    \centering
    \includegraphics[width=\textwidth]{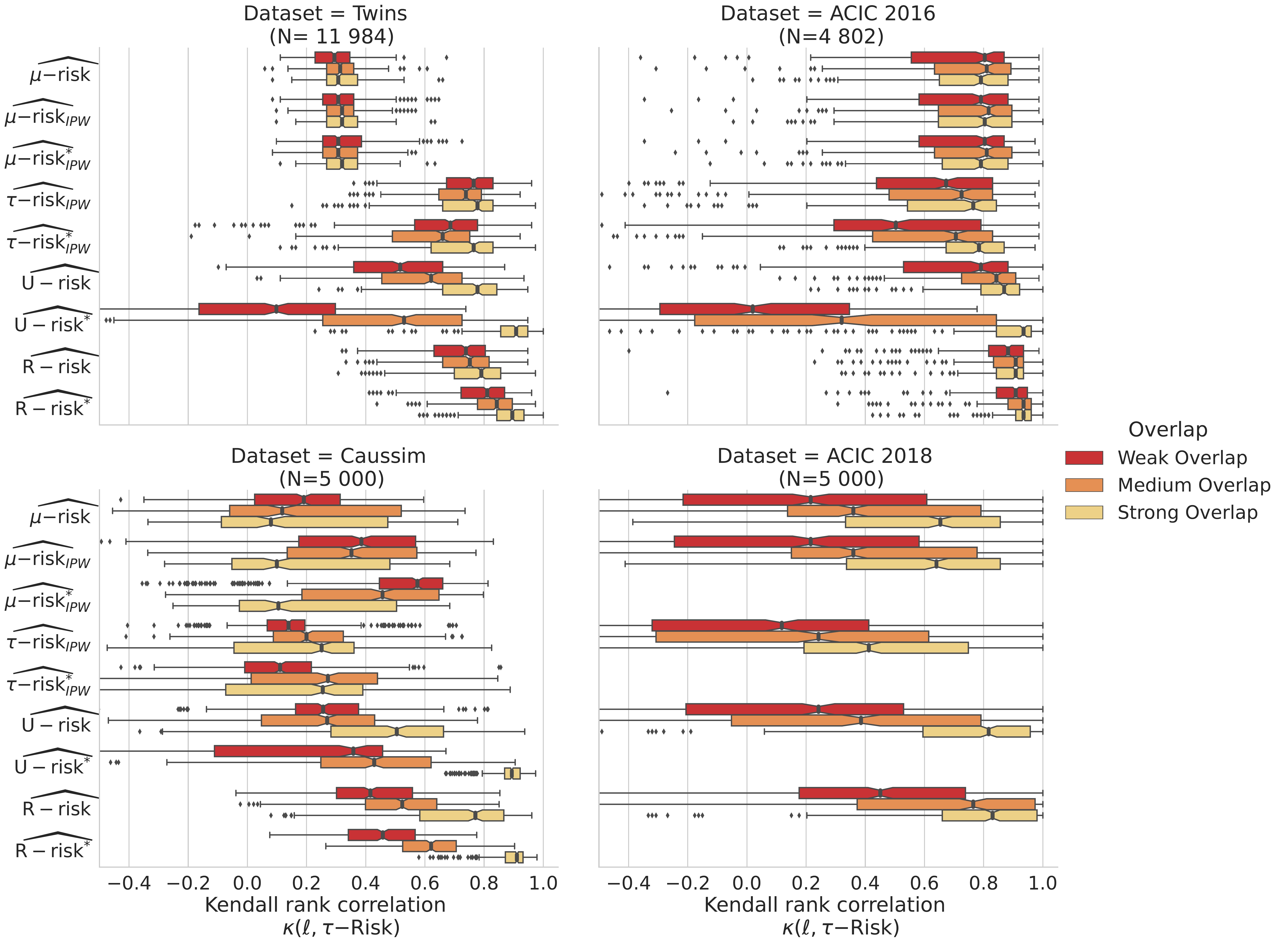}
    \hfill
    \caption{\textbf{Low population overlap hinders causal model selection for all
            metrics}:
        Kendall's $\tau$ agreement with $\tau\text{-risk}$. Strong, medium and Weak overlap
        correspond
        to the tertiles of the overlap distribution measured with Normalized Total
        Varation eq. \ref{eq:ntv}.}\label{apd:fig:all_datasets_overlap_effect}
\end{figure}

\paragraph{Figure \ref{apd:fig:all_datasets_overlap_effect} Low population
    overlap hinders model selection for all metrics}

\begin{figure}
    \begin{subfigure}[b]{0.49\textwidth}
        \centering
        \caption{\textbf{Twins}}
        \includegraphics[width=1\textwidth]{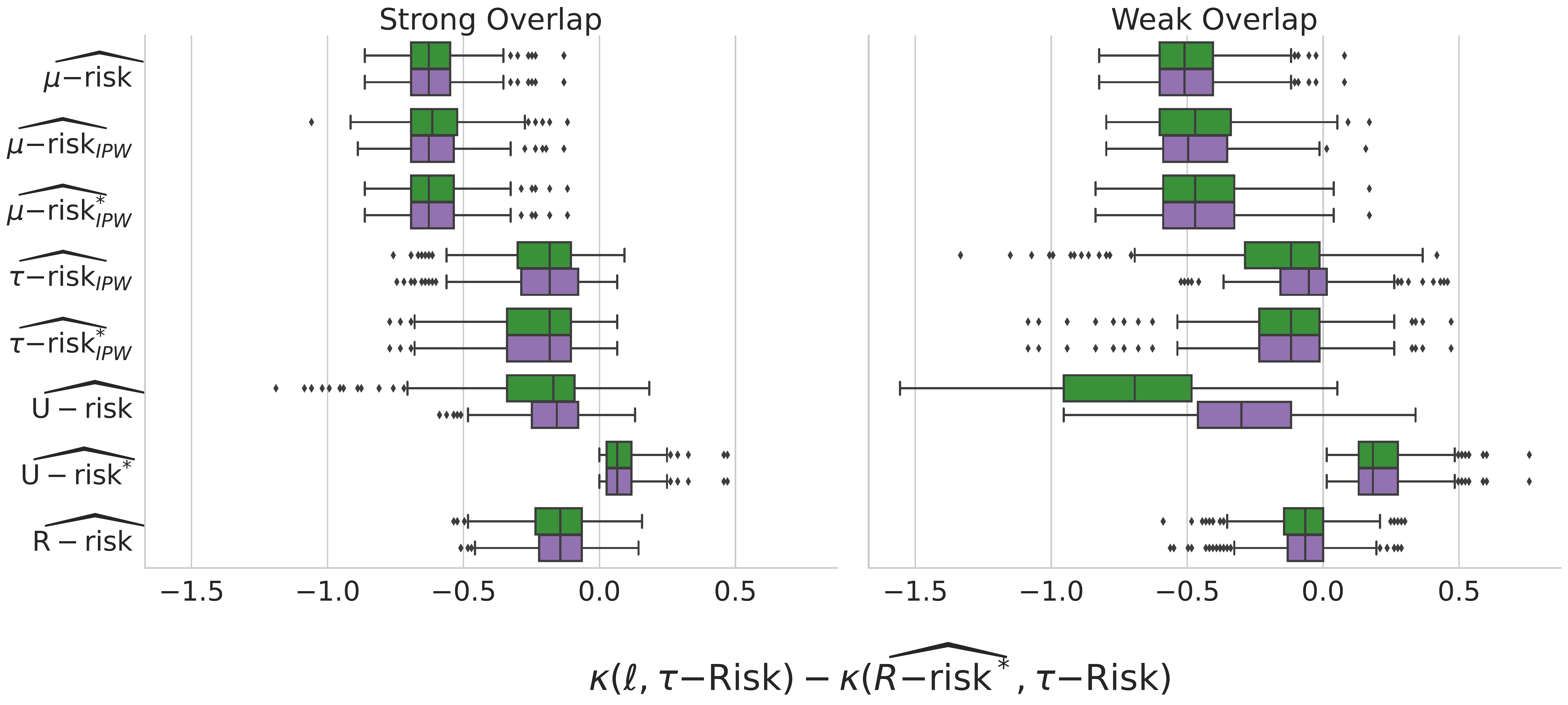}
        \label{fig:experiments:nuisance_comparison:twins}
    \end{subfigure}
    \hfill
    \begin{subfigure}[b]{0.49\textwidth}
        \centering
        \caption{\textbf{Twins downsampled}}
        \includegraphics[width=1\textwidth]{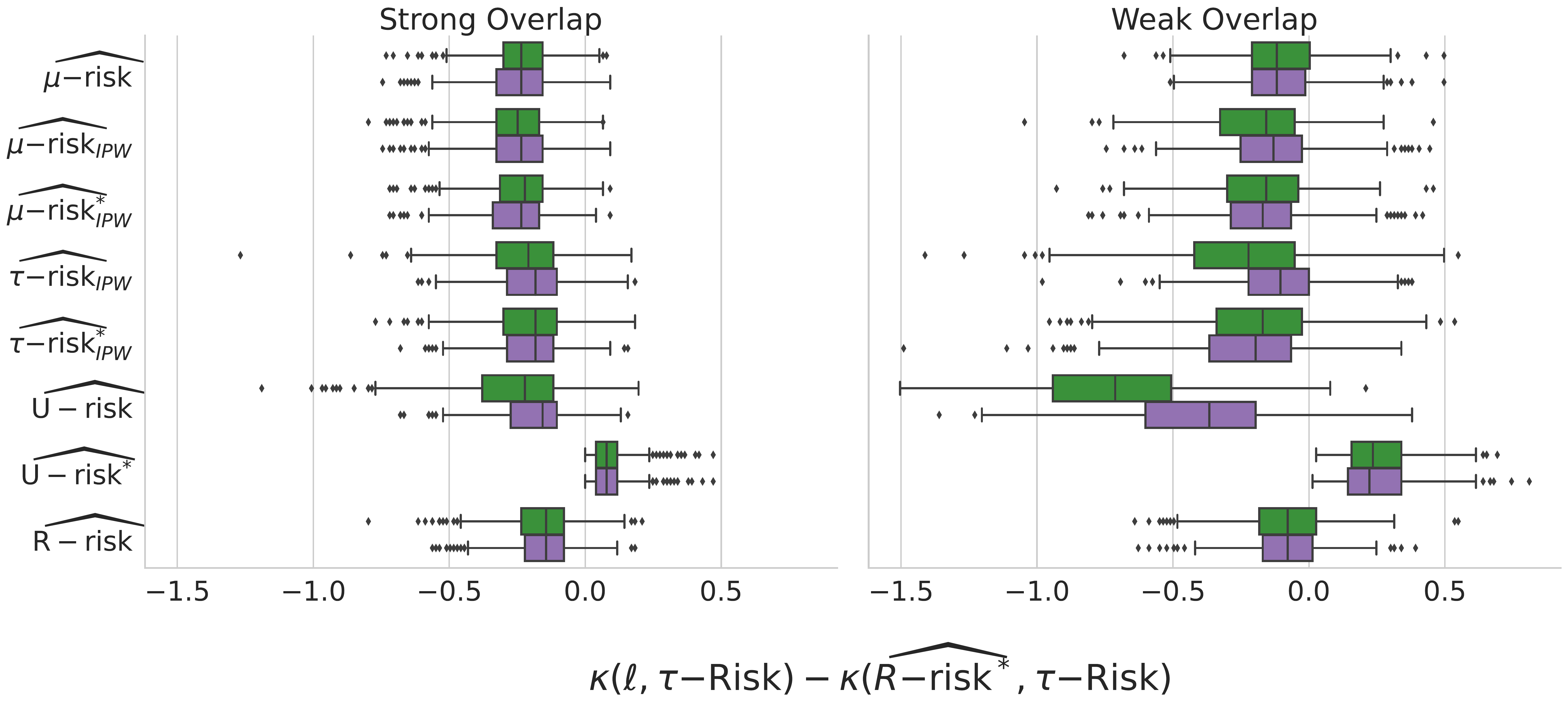}
        \label{fig:experiments:nuisance_comparison:twins_ds}
    \end{subfigure}
    \hfill
    \begin{subfigure}[b]{0.49\textwidth}
        %\centering
        \caption{\textbf{Caussim}}
        \includegraphics[width=1.15\textwidth]{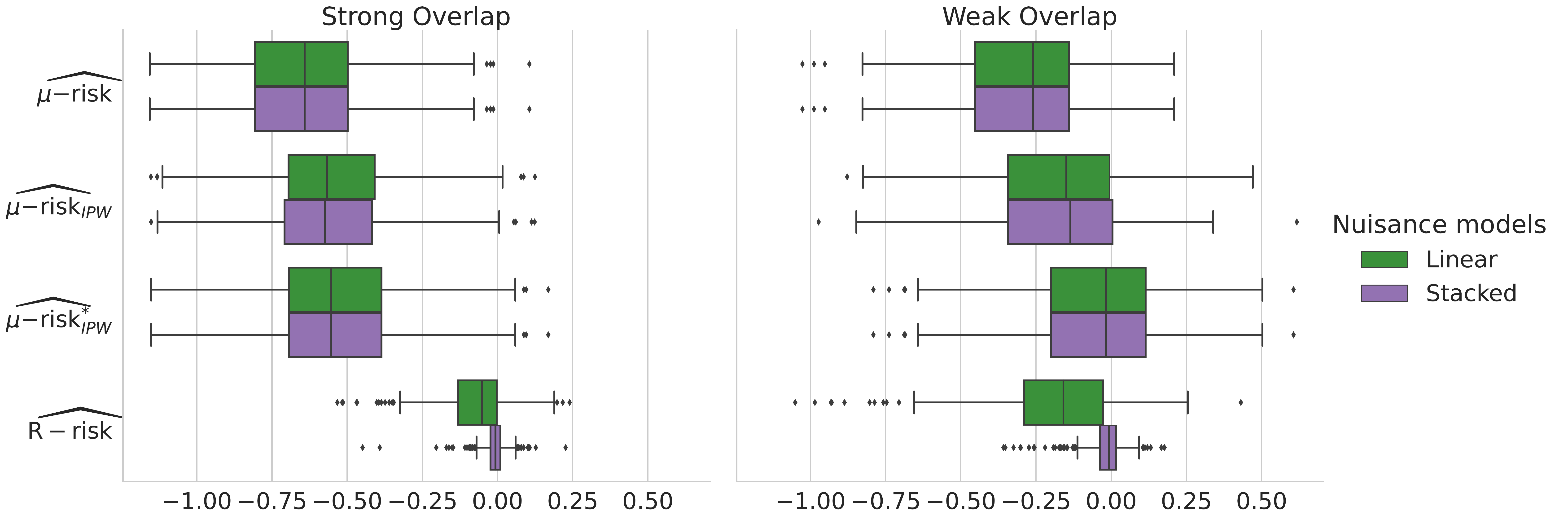}
        \label{fig:experiments:nuisance_comparison:caussim}
    \end{subfigure}
    \hfill
    \begin{subfigure}[b]{0.49\textwidth}
        \centering
        \caption{\textbf{ACIC 2016}}
        \includegraphics[width=1\textwidth]{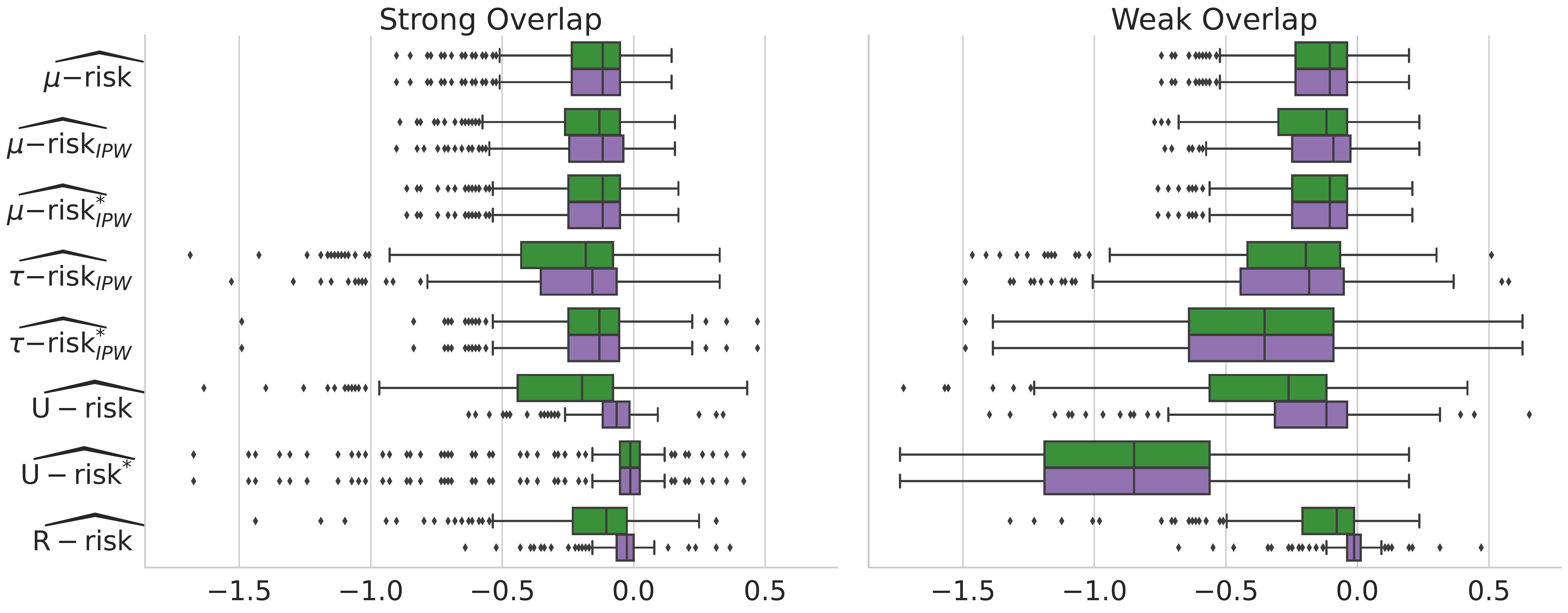}
        \label{fig:experiments:nuisance_comparison:acic_2016}
    \end{subfigure}
    \hfill
    \caption{Learning the nuisances with \textcolor{DarkOrchid}{stacked models} (linear and
        gradient boosting) is important for successful model selection with R-risk.
        For Twins dataset, there is no improvement for \textcolor{DarkOrchid}{stacked models} compared to
        \textcolor{ForestGreen}{linear models} because of the linearity of the propensity model.}\label
    {apd:fig:nuisances_comparison}
\end{figure}

\paragraph{Figure \ref{apd:fig:nuisances_comparison} - Stacked models for the nuisances is more efficient}
For each metrics the benefit of
using a stacked model of linear and boosting estimators for nuisances compared
to a linear model. The evaluation measure is Kendall's tau relative to the
oracle $R\text{-risk}^{\star}$ to have a stable reference between exepriments.
Thus, we do not include in this analysis the ACIC 2018 dataset since
$R\text{-risk}^{\star}$ is not available due to the lack of the true propensity
score.

\begin{figure}
    \centering
    \includegraphics[width=\textwidth]{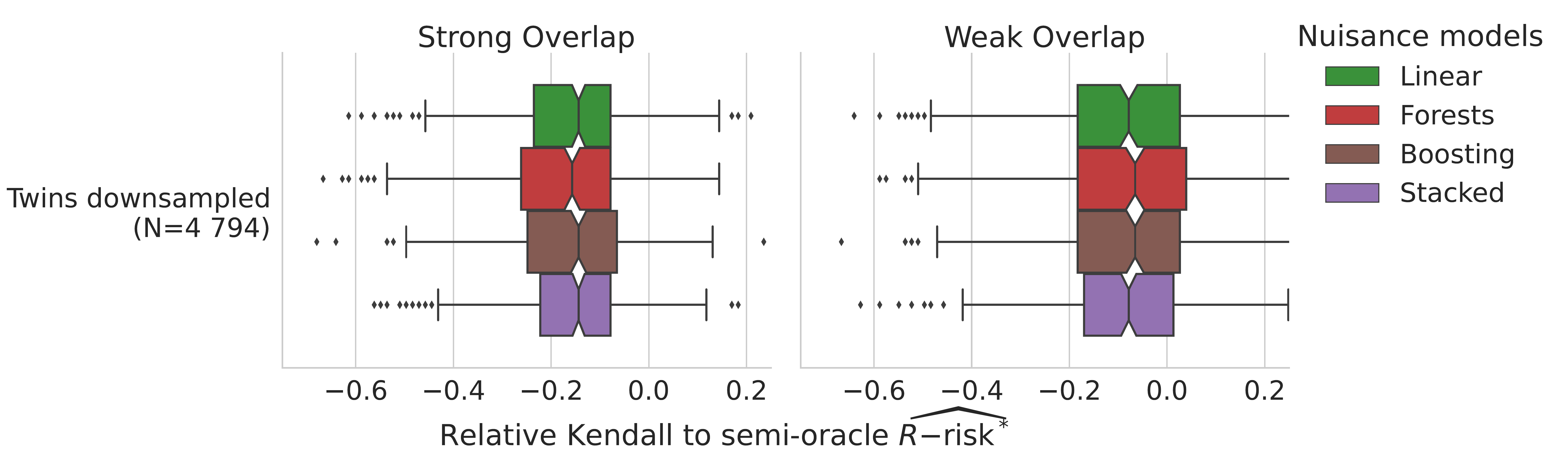}
    \hfill
    \caption{\textbf{Flexible models are performant in recovering nuisances
            in the downsampled Twins dataset.} The propensity score is linear in this
        setup, making it particularly challenging for flexible models compared to
        linear methods.}\label{apd:fig:nuisances_comparison_twins}
\end{figure}

\paragraph{Figure \ref{apd:fig:nuisances_comparison_twins} - Flexible models are performant in recovering nuisances even
    in linear setups}

\paragraph{Selecting different seeds and parameters is crucial to draw
    conclucions}\label{{apd:results:seed_effect}}

One strength of our study is the various number of different simulated and
semi-simulated datasets. We are convinced that the usual practice of using only
a small number of generation processes does not allow to draw statistically
significant conclusions.

\begin{figure}
    \centering
    \caption{Kendall correlation coefficients for each causal metric. Each (color,
        shape) pair indicates a different (treatment ratio, seed) of the generation
        process.}\label {apd:results:fig:seed_effect}
    \includegraphics[width=\linewidth]{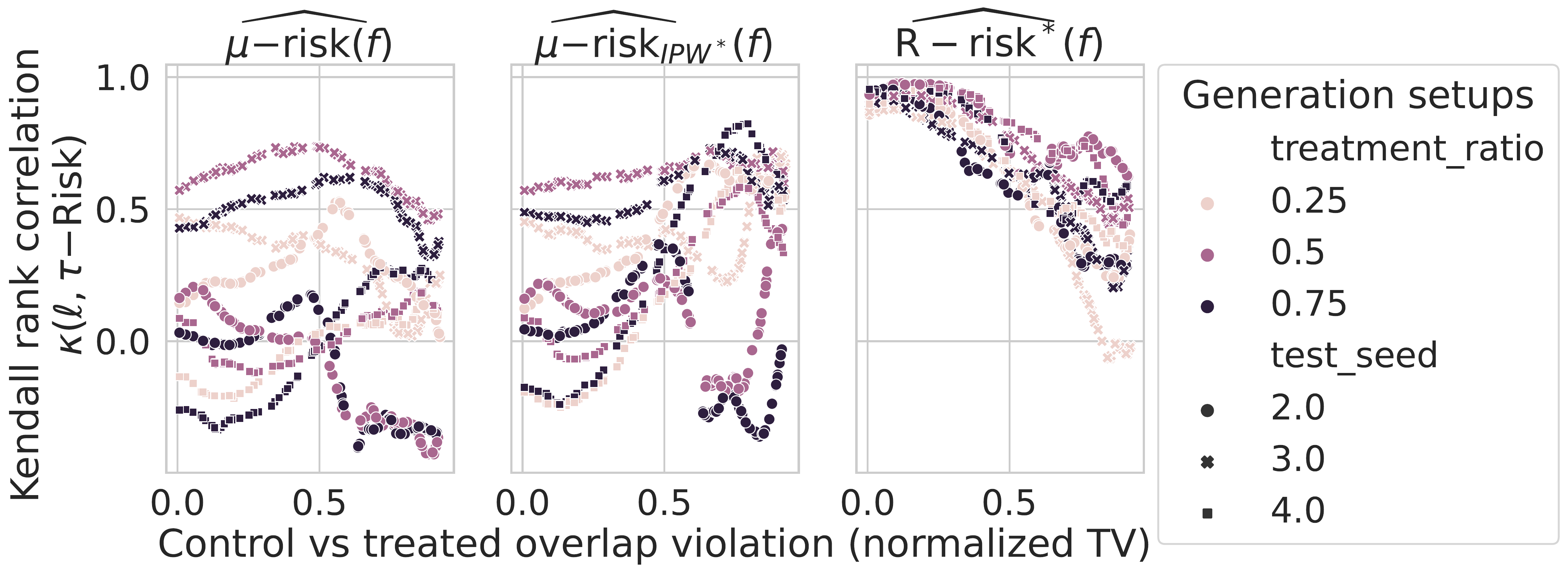}
\end{figure}

Figure \ref{apd:results:fig:seed_effect} illustrate the dependence of the
results on the generation process for caussim simulations. We highlighted the
different trajectories induced by three different seeds for data generation and
three different treatment ratio instead of 1000 different seeds. The result
curves are relatively stable from one setup to another for $R{-risk}$, but vary
strongly for $\mu\text{-risk}$ and $\mu\text{-risk}_{IPW}$.

\FloatBarrier

\section{Heterogeneity in practices for data split}\label{apd:results:k_fold_choices}

Splitting the data is common when using machine learning for causal
inference, but practices vary widely in terms of the fraction of data to
allocate to train models, outcomes and nuisances, and to evaluate them.

Before even model selection, data splitting is often required for
estimation of the treatment effect, ATE or CATE, for instance to compute
the nuisances required to optimize the outcome model (as the
$R\text{-risk}$, definition \autoref{def:r_risk}).
The most frequent choice is use 80\% of the data to fit the models,
and 20\% to evaluate them.
For instance, for CATE estimation, the R-learner has been introduced using K-folds with K = 5
and K = 10: 80\% of the data (4 folds) to train the nuisances and the remaining
fold to minimize the corresponding R-loss \cite{nie_quasioracle_2017}.
Yet, it has been implemented with K=5 in causallib
\cite{causalevaluations} or K=3 in econML \cite{econml}.
Likewise, for ATE estimation, \citet{chernozhukov_double_2018}
introduce doubly-robust machine learning,
recommending K=5 based on an empirical comparison K=2. However,
subsequent works use doubly robust ML with varying choices
of K: \citet{loiseau_external_2022} use K=3, \citet{gao_assessment_2021} use
K=2. In the econML implementation, K is set to 3 \cite{econml}.
\citet{naimi2021challenges} evaluate various machine-learning approaches
--including R-learners-- using K=5 and 10, drawing inspiration from the
TMLE literature which sets
K=5 in the TMLE package \cite{tmle_package_2012}.

Causal model selection has been much less discussed. The only study that
we are aware of, \citet{schuler_comparison_2018}, use a different data
split: a 2-folds train/test procedure,
training the nuisances on the first half of the data, and using the
second half to estimate the $R\text{-risk}$ and select the best treatment
effect model.

% \paragraph{Simulations: naive reweighting of the $R \text{-risk}$}

% Applying a naive reweighting, $w(x, a)=\frac{1}{e(x)\big(1-e(x) \big )}$ to the
% $R \text{-risk}$ to recover the $\tau \text{-risk}$ in the first part of
% \ref{theory:prop:r_risk_rewrite} makes the residuals explode in case of noise as
% shown in Figure \ref{apd:simu_noised}.

% \begin{figure}[htbp]
%   \centering
%   \caption{Caussim simulations (500 repetitions):
%   $R\text{-risk}_{IPW^*_{naive}}$ (in green) is exploding}\label
%   {apd:simu_noised}
%   \includegraphics[width=0.7\linewidth]{images/noised_0.001_simu_best_normalized_abs_bias_ate_vs_mmd.pdf}
% \end{figure}

\end{document}